\documentclass[10pt,letterpaper]{preprint}
\makeatletter\providecommand\LenToUnit[1]{#1\@gobble}\makeatother
\usepackage{calc}

\newcommand{\newref}[2]{~\cite{#1}}   %

\makeatletter\renewcommand\LenToUnit[1]{\strip@pt\dimexpr(#1)*\p@/\unitlength\relax}\makeatother

\usepackage{cuted} %

\usepackage{currfile} %

\usepackage{caption} %

\usepackage{amsmath,amssymb,amsthm,mathtools}
\usepackage{enumitem}
\usepackage{booktabs}
\usepackage{array}
\usepackage{tabularx}
\usepackage{colortbl}                    %
\definecolor{oursrow}{HTML}{D6E6F7}     %
\newcolumntype{Y}{>{\centering\arraybackslash}X}   %
\usepackage{graphicx}
\usepackage{xspace}
\newtheorem{proposition}{Proposition}

\newtheorem{lemma}{Lemma}
\theoremstyle{definition}

\theoremstyle{remark}

\providecommand{\vsDcPilotRows}{53,976}
\providecommand{\vsDcPilotImgs}{25,727}
\providecommand{\vsDcStates}{2,958}
\providecommand{\vsDcPhashMin}{10}
\providecommand{\vsDcCmds}{40}
\newcommand{\vfResMin}{$256^2$}
\newcommand{\vfResMax}{1\,MP}
\newcommand{\vfEdgeRes}{$384^2$}
\newcommand{\vfMixSize}{600k}
\newcommand{\vfMixSources}{12}
\newcommand{\vfMixPointFrac}{35\%}
\newcommand{\vfMixBoxFrac}{35\%}
\newcommand{\vfMixGuiFrac}{30\%}
\newcommand{\vfLeakImages}{8{,}987}
\newcommand{\vfLeakPB}{133}
\newcommand{\vfVllmVer}{0.25.1}
\newcommand{\vNsFinPt}{67.0}
\newcommand{\vNsFinPB}{56.5}
\newcommand{\vNsFinSS}{55.6}
\newcommand{\vNsFinWP}{51.0}
\newcommand{\vNsFinRR}{68.8}
\newcommand{\vNsFinRC}{89.8}
\newcommand{\vNsFinMw}{54.1}
\newcommand{\vNsFinMwY}{71.5}
\newcommand{\vNsFinMwYAct}{2.3}
\newcommand{\vNsSetsPt}{67.1}
\newcommand{\vNsSFTPt}{62.0}
\newcommand{\vNsSFTPB}{55.2}
\newcommand{\vNsSFTSS}{48.4}
\newcommand{\vNsSFTWP}{46.0}
\newcommand{\vNsSFTRR}{64.6}
\newcommand{\vNsSFTRC}{82.5}
\newcommand{\vNsSFTMw}{42.7}
\newcommand{\vNsSFTMwG}{62.1}
\newcommand{\vNmFinPt}{73.3}
\newcommand{\vNmFinPB}{60.1}
\newcommand{\vNmFinSS}{66.8}
\newcommand{\vNmFinWP}{61.0}
\newcommand{\vNmFinRR}{75.8}
\newcommand{\vNmFinRC}{90.6}
\newcommand{\vNmFinCov}{90.3}
\newcommand{\vNmFinReg}{1.36}
\newcommand{\vNmFinPBCov}{80.5}
\newcommand{\vNmFinSSCov}{91.2}
\newcommand{\vNmFinWPCov}{84.0}
\newcommand{\vNmFinRRCov}{91.0}
\newcommand{\vNmFinRCCov}{97.6}
\newcommand{\vNmFinMw}{70.8}
\newcommand{\vNmFinMwY}{73.2}
\newcommand{\vNmFinMwYAct}{29.3}
\newcommand{\vNmSetsPt}{72.3}
\newcommand{\vNmSFTPt}{68.7}
\newcommand{\vNmSFTPB}{60.1}
\newcommand{\vNmSFTSS}{59.0}
\newcommand{\vNmSFTWP}{50.0}
\newcommand{\vNmSFTRR}{73.0}
\newcommand{\vNmSFTRC}{82.8}
\newcommand{\vNmSFTMw}{60.1}
\newcommand{\vNmSFTMwG}{67.7}
\newcommand{\vNmFinBxAvg}{89.0}

\newcommand{\vNmFsPBTenCov}{90.8}
\newcommand{\vNmFsPBTenLow}{80.7}
\newcommand{\vNmFsWPTenCov}{90.8}
\newcommand{\vNmFsWPTenLow}{82.2}
\newcommand{\vNlFinPt}{79.4}
\newcommand{\vNlFinPB}{67.6}
\newcommand{\vNlFinSS}{74.3}
\newcommand{\vNlFinWP}{65.0}
\newcommand{\vNlFinRR}{81.7}
\newcommand{\vNlFinRC}{94.3}
\newcommand{\vNlFinMw}{75.2}
\newcommand{\vNlFinMwY}{72.1}
\newcommand{\vNlFinMwYAct}{29.3}
\newcommand{\vNlSetsPt}{78.1}
\newcommand{\vNlSFTPt}{75.7}
\newcommand{\vNlSFTPB}{65.3}
\newcommand{\vNlSFTSS}{67.8}
\newcommand{\vNlSFTWP}{67.0}
\newcommand{\vNlSFTRR}{81.3}
\newcommand{\vNlSFTRC}{85.6}
\newcommand{\vNlSFTMw}{59.9}
\newcommand{\vNlSFTMwG}{75.6}
\newcommand{\vNlFinBxAvg}{90.7}
\newcommand{\vNxFinPt}{81.9}
\newcommand{\vNxFinPB}{71.2}
\newcommand{\vNxFinSS}{82.6}
\newcommand{\vNxFinWP}{74.0}
\newcommand{\vNxFinRR}{81.8}
\newcommand{\vNxFinRC}{92.4}
\newcommand{\vNxFinMw}{77.9}
\newcommand{\vNxFinMwY}{80.3}
\newcommand{\vNxFinMwYAct}{53.7}
\newcommand{\vNxSetsPt}{81.3}
\newcommand{\vNxSFTPt}{78.9}
\newcommand{\vNxSFTPB}{69.1}
\newcommand{\vNxSFTSS}{76.5}
\newcommand{\vNxSFTWP}{71.0}
\newcommand{\vNxSFTRR}{82.2}
\newcommand{\vNxSFTRC}{85.7}
\newcommand{\vNxSFTMw}{67.3}
\newcommand{\vNxSFTMwG}{72.9}
\newcommand{\vNxSFTMwNat}{23.9}
\newcommand{\vNxFinBxAvg}{92.0}
\newcommand{\vNivlFinPt}{73.1}
\newcommand{\vNivlFinPB}{64.1}
\newcommand{\vNivlFinSS}{70.2}
\newcommand{\vNivlFinWP}{53.0}
\newcommand{\vNivlFinRR}{71.5}
\newcommand{\vNivlFinRC}{90.9}
\newcommand{\vNivlFinMw}{76.1}
\newcommand{\vNivlFinMwY}{67.1}
\newcommand{\vNivlFinMwYAct}{7.9}
\newcommand{\vNivlSFTPt}{69.0}
\newcommand{\vNivlSFTPB}{62.7}
\newcommand{\vNivlSFTSS}{65.1}
\newcommand{\vNivlSFTWP}{46.0}
\newcommand{\vNivlSFTRR}{68.2}
\newcommand{\vNivlSFTRC}{83.7}
\newcommand{\vNivlSFTMw}{70.4}
\newcommand{\vNivlSFTMwG}{68.3}
\newcommand{\vNgemFinPt}{76.5}
\newcommand{\vNgemFinPB}{63.5}
\newcommand{\vNgemFinSS}{78.6}
\newcommand{\vNgemFinWP}{63.0}
\newcommand{\vNgemFinRR}{75.3}
\newcommand{\vNgemFinRC}{90.3}
\newcommand{\vNgemFinMw}{82.1}
\newcommand{\vNgemFinMwY}{81.9}
\newcommand{\vNgemFinMwYAct}{49.6}
\newcommand{\vNgemSFTPt}{73.9}
\newcommand{\vNgemSFTPB}{63.2}
\newcommand{\vNgemSFTSS}{76.3}
\newcommand{\vNgemSFTWP}{60.0}
\newcommand{\vNgemSFTRR}{73.0}
\newcommand{\vNgemSFTRC}{84.3}
\newcommand{\vNgemSFTMw}{80.3}
\newcommand{\vNlfmFinPt}{78.1}
\newcommand{\vNlfmFinPB}{70.0}
\newcommand{\vNlfmFinSS}{72.6}
\newcommand{\vNlfmFinWP}{62.0}
\newcommand{\vNlfmFinRR}{79.3}
\newcommand{\vNlfmFinRC}{92.4}
\newcommand{\vNlfmFinMw}{73.1}
\newcommand{\vNlfmFinMwY}{75.1}
\newcommand{\vNlfmFinMwYAct}{57.2}
\newcommand{\vNlfmSFTPt}{73.8}
\newcommand{\vNlfmSFTPB}{65.6}
\newcommand{\vNlfmSFTSS}{68.9}
\newcommand{\vNlfmSFTWP}{61.0}
\newcommand{\vNlfmSFTRR}{75.4}
\newcommand{\vNlfmSFTRC}{86.1}
\newcommand{\vNlfmSFTMw}{65.2}
\newcommand{\vNlfmSFTMwG}{42.1}
\newcommand{\vNllvFinPt}{59.3}
\newcommand{\vNllvFinPB}{52.9}
\newcommand{\vNllvFinSS}{43.8}
\newcommand{\vNllvFinWP}{54.0}
\newcommand{\vNllvFinRR}{58.2}
\newcommand{\vNllvFinRC}{88.0}
\newcommand{\vNllvFinMw}{52.7}
\newcommand{\vNllvSFTPt}{52.0}
\newcommand{\vNllvSFTPB}{49.6}
\newcommand{\vNllvSFTSS}{28.6}
\newcommand{\vNllvSFTWP}{49.0}
\newcommand{\vNllvSFTRR}{52.6}
\newcommand{\vNllvSFTRC}{82.9}
\newcommand{\vNllvSFTMw}{32.5}
\newcommand{\vNllvSFTMwG}{43.7}
\newcommand{\vNivlAFinPt}{62.8}
\newcommand{\vNivlAFinPB}{55.7}
\newcommand{\vNivlAFinSS}{50.2}
\newcommand{\vNivlAFinWP}{40.0}
\newcommand{\vNivlAFinRR}{63.3}
\newcommand{\vNivlAFinRC}{87.2}
\newcommand{\vNivlAFinMw}{64.7}
\newcommand{\vNivlAFinMwY}{67.5}
\newcommand{\vNivlAFinMwYAct}{15.9}
\newcommand{\vNivlASFTPt}{58.5}
\newcommand{\vNivlASFTPB}{56.4}
\newcommand{\vNivlASFTSS}{41.5}
\newcommand{\vNivlASFTWP}{34.0}
\newcommand{\vNivlASFTRR}{60.8}
\newcommand{\vNivlASFTRC}{80.3}
\newcommand{\vNivlASFTMw}{61.9}
\newcommand{\vNivlDFinPt}{76.2}
\newcommand{\vNivlDFinPB}{67.6}
\newcommand{\vNivlDFinSS}{75.4}
\newcommand{\vNivlDFinWP}{68.0}
\newcommand{\vNivlDFinRR}{73.5}
\newcommand{\vNivlDFinRC}{92.1}
\newcommand{\vNivlDFinMw}{83.9}
\newcommand{\vNivlDFinMwY}{77.5}
\newcommand{\vNivlDFinMwYAct}{46.7}
\newcommand{\vNivlDSFTPt}{72.7}
\newcommand{\vNivlDSFTPB}{64.4}
\newcommand{\vNivlDSFTSS}{72.3}
\newcommand{\vNivlDSFTWP}{61.0}
\newcommand{\vNivlDSFTRR}{71.6}
\newcommand{\vNivlDSFTRC}{84.8}
\newcommand{\vNivlDSFTMw}{85.6}
\newcommand{\vNivlHFinPt}{76.3}
\newcommand{\vNivlHFinPB}{67.3}
\newcommand{\vNivlHFinSS}{75.2}
\newcommand{\vNivlHFinWP}{61.0}
\newcommand{\vNivlHFinRR}{74.1}
\newcommand{\vNivlHFinRC}{92.5}
\newcommand{\vNivlHFinMw}{85.7}
\newcommand{\vNivlHFinMwY}{80.9}
\newcommand{\vNivlHFinMwYAct}{50.5}
\newcommand{\vNivlHSFTPt}{72.3}
\newcommand{\vNivlHSFTPB}{65.2}
\newcommand{\vNivlHSFTSS}{71.5}
\newcommand{\vNivlHSFTWP}{63.0}
\newcommand{\vNivlHSFTRR}{70.8}
\newcommand{\vNivlHSFTRC}{84.3}
\newcommand{\vNivlHSFTMw}{74.0}
\newcommand{\vNivlHFinBxAvg}{87.2}
\newcommand{\vNlfvFinPt}{73.9}
\newcommand{\vNlfvFinPB}{66.4}
\newcommand{\vNlfvFinSS}{55.3}
\newcommand{\vNlfvFinWP}{69.0}
\newcommand{\vNlfvFinRR}{80.0}
\newcommand{\vNlfvFinRC}{92.9}
\newcommand{\vNlfvFinMw}{61.5}
\newcommand{\vNlfvFinMwY}{72.5}
\newcommand{\vNlfvFinMwYAct}{31.6}
\newcommand{\vNlfvSFTPt}{70.8}
\newcommand{\vNlfvSFTPB}{62.4}
\newcommand{\vNlfvSFTSS}{50.9}
\newcommand{\vNlfvSFTWP}{66.0}
\newcommand{\vNlfvSFTRR}{80.2}
\newcommand{\vNlfvSFTRC}{85.8}
\newcommand{\vNlfvSFTMw}{46.1}
\newcommand{\vNlfvSFTMwG}{14.3}
\newcommand{\vNowlMw}{74.4}
\newcommand{\vNowlMwG}{74.9}
\newcommand{\vNguidMw}{62.3}
\newcommand{\vNguidMwG}{64.9}
\newcommand{\vNgpiMw}{55.7}
\newcommand{\vNgpiMwG}{52.0}
\newcommand{\vNPixRatio}{6.8}
\newcommand{\vNsFinLo}{67.0}
\newcommand{\vNsFinHi}{79.0}
\newcommand{\vNsFinNoShare}{43.9}
\newcommand{\vNsFinNoAcc}{77.5}
\newcommand{\vNsFinNoPix}{3.99}
\newcommand{\vNsFinNoRn}{72.3}
\newcommand{\vNsFinNoTm}{74.6}
\newcommand{\vNsFinNoGain}{88}
\newcommand{\vNmFinLo}{73.3}
\newcommand{\vNmFinHi}{82.8}
\newcommand{\vNmFinNoShare}{36.9}
\newcommand{\vNmFinNoAcc}{80.8}
\newcommand{\vNmFinNoPix}{3.51}
\newcommand{\vNmFinNoRn}{76.8}
\newcommand{\vNmFinNoTm}{78.9}
\newcommand{\vNmFinNoGain}{79}
\newcommand{\vNlFinLo}{79.4}
\newcommand{\vNlFinHi}{85.5}
\newcommand{\vNlFinNoShare}{27.9}
\newcommand{\vNlFinNoAcc}{84.6}
\newcommand{\vNlFinNoPix}{2.90}
\newcommand{\vNlFinNoRn}{81.1}
\newcommand{\vNlFinNoTm}{82.6}
\newcommand{\vNlFinNoGain}{85}
\newcommand{\vNxFinLo}{81.9}
\newcommand{\vNxFinHi}{86.2}
\newcommand{\vNxFinNoShare}{22.1}
\newcommand{\vNxFinNoAcc}{85.2}
\newcommand{\vNxFinNoPix}{2.50}
\newcommand{\vNxFinNoRn}{82.8}
\newcommand{\vNxFinNoTm}{84.0}
\newcommand{\vNxFinNoGain}{78}
\newcommand{\vNCigemFinSFT}{+2.6}
\newcommand{\vNCigemFinSFTLo}{+1.6}
\newcommand{\vNCigemFinSFTHi}{+3.7}
\newcommand{\vNCiivlAFinSFT}{+4.3}
\newcommand{\vNCiivlAFinSFTLo}{+3.2}
\newcommand{\vNCiivlAFinSFTHi}{+5.4}
\newcommand{\vNCiivlDFinSFT}{+3.5}
\newcommand{\vNCiivlDFinSFTLo}{+2.6}
\newcommand{\vNCiivlDFinSFTHi}{+4.5}
\newcommand{\vNCiivlFinSFT}{+4.1}
\newcommand{\vNCiivlFinSFTLo}{+3.1}
\newcommand{\vNCiivlFinSFTHi}{+5.2}
\newcommand{\vNCiivlHFinSFT}{+4.0}
\newcommand{\vNCiivlHFinSFTLo}{+3.0}
\newcommand{\vNCiivlHFinSFTHi}{+5.0}
\newcommand{\vNCilFinSFT}{+3.7}
\newcommand{\vNCilFinSFTLo}{+2.8}
\newcommand{\vNCilFinSFTHi}{+4.7}
\newcommand{\vNCilFinSets}{+1.3}
\newcommand{\vSoFvSIvAcell}{$-$0.8}
\newcommand{\vNCilfmFinSFT}{+4.3}
\newcommand{\vNCilfmFinSFTLo}{+3.4}
\newcommand{\vNCilfmFinSFTHi}{+5.3}
\newcommand{\vNCilfvFinSFT}{+3.1}
\newcommand{\vNCilfvFinSFTLo}{+2.1}
\newcommand{\vNCilfvFinSFTHi}{+4.1}
\newcommand{\vNCillvFinSFT}{+7.3}
\newcommand{\vNCillvFinSFTLo}{+6.1}
\newcommand{\vNCillvFinSFTHi}{+8.5}
\newcommand{\vNCimFinSFT}{+4.5}
\newcommand{\vNCimFinSFTLo}{+3.5}
\newcommand{\vNCimFinSFTHi}{+5.6}
\newcommand{\vNCisFinSFT}{+5.0}
\newcommand{\vNCisFinSFTLo}{+3.9}
\newcommand{\vNCisFinSFTHi}{+6.1}
\newcommand{\vNCixFinSFT}{+3.0}
\newcommand{\vNCixFinSFTLo}{+2.1}
\newcommand{\vNCixFinSFTHi}{+3.9}
\newcommand{\vNnPB}{982}
\newcommand{\vNnSS}{1,272}
\newcommand{\vNnWP}{100}
\newcommand{\vNnRR}{2,000}
\newcommand{\vNnRC}{1,000}
\newcommand{\vNValN}{2,000}
\newcommand{\vNllvFinBxAvg}{76.0}
\newcommand{\vNmjevMw}{78.5}
\newcommand{\vQlPoolN}{305}
\newcommand{\vQlPoolOurs}{64.9}
\newcommand{\vQlPoolPts}{43.3}
\newcommand{\vQlPoolDiff}{+21.6}
\newcommand{\vQlPoolLo}{+15.4}
\newcommand{\vQlPoolHi}{+27.9}
\newcommand{\vAxCtopAuc}{0.806}
\newcommand{\vAxCtopAucFin}{0.810}
\newcommand{\vAxCtopNo}{23.2}
\newcommand{\vAxCtopAucDiffV}{\tm{}0.004}
\newcommand{\vAxCtopAucLoV}{\tm{}0.016}
\newcommand{\vAxCtopAucHiV}{+0.008}
\newcommand{\vPlFinPt}{73.3}
\newcommand{\vPlFinBx}{89.0}
\newcommand{\vPlEiPt}{72.9}
\newcommand{\vPlEiPtD}{\tm{}0.3}
\newcommand{\vPlEiPtLo}{\tm{}1.2}
\newcommand{\vPlEiPtHi}{+0.5}
\newcommand{\vPlEiBx}{88.2}
\newcommand{\vPlEiBxD}{\tm{}0.8}
\newcommand{\vPlTfPt}{72.0}
\newcommand{\vPlTfPtD}{\tm{}1.3}
\newcommand{\vPlTfPtLo}{\tm{}2.2}
\newcommand{\vPlTfPtHi}{\tm{}0.4}
\newcommand{\vPlTfBx}{88.9}
\newcommand{\vPlTfBxD}{\tm{}0.1}
\newcommand{\vPlTsPt}{71.7}
\newcommand{\vPlTsPtD}{\tm{}1.6}
\newcommand{\vPlTsPtLo}{\tm{}2.5}
\newcommand{\vPlTsPtHi}{\tm{}0.7}
\newcommand{\vPlTsBx}{88.8}
\newcommand{\vPlTsBxD}{\tm{}0.2}
\newcommand{\vPlCtPt}{73.1}
\newcommand{\vPlCtPtD}{\tm{}0.2}
\newcommand{\vPlCtPtLo}{\tm{}1.0}
\newcommand{\vPlCtPtHi}{+0.6}
\newcommand{\vPlCtBx}{88.7}
\newcommand{\vPlCtBxD}{\tm{}0.3}
\newcommand{\vPlEcPt}{73.1}
\newcommand{\vPlEcPtD}{\tm{}0.2}
\newcommand{\vPlEcPtLo}{\tm{}1.1}
\newcommand{\vPlEcPtHi}{+0.6}
\newcommand{\vPlEcBx}{88.1}
\newcommand{\vPlEcBxD}{\tm{}1.0}
\newcommand{\vLHivlAGen}{335}
\newcommand{\vLHivlGen}{344}
\newcommand{\vLHivlDGen}{491}
\newcommand{\vLHgemGen}{941}
\newcommand{\vLHgemFull}{490}
\newcommand{\vLHgemCell}{188}
\newcommand{\vLHgemRatioF}{1.9}
\newcommand{\vLHlfmGen}{284}
\newcommand{\vLHlfmFull}{127}
\newcommand{\vLHlfmCell}{127}
\newcommand{\vLHlfmRatioF}{2.2}
\newcommand{\vLHlfvGen}{368}
\newcommand{\vLHlfvFull}{136}
\newcommand{\vLHlfvCell}{136}
\newcommand{\vLHlfvRatioF}{2.7}
\newcommand{\vLHivlHGen}{441}
\newcommand{\vLHivlHCell}{121}
\newcommand{\vLHivlHRatioC}{3.7}
\newcommand{\vLHllvGen}{407}
\newcommand{\vLHllvCell}{103}
\newcommand{\vLHllvRatioC}{3.9}
\newcommand{\vRvCrossOursQm}{12.7}
\newcommand{\vRvCrossPtsQm}{13.8}
\newcommand{\vRvCrossAllOursQm}{3.4}
\newcommand{\vRvCrossAllPtsQm}{4.3}
\newcommand{\vRvCrossOursQl}{12.3}
\newcommand{\vRvCrossPtsQl}{16.2}
\newcommand{\vRvCrossAllOursQl}{2.5}
\newcommand{\vRvCrossAllPtsQl}{3.9}
\newcommand{\vRvWsPtsQx}{12.9}
\newcommand{\vRvWsPyesQx}{8.8}
\newcommand{\vRvWsPtsIvd}{18.0}
\newcommand{\vRvWsPyesIvd}{11.5}
\newcommand{\vRvWsPtsLlv}{37.3}
\newcommand{\vRvWsPyesLlv}{28.0}
\newcommand{\vRvAuYesGainMin}{0.07}
\newcommand{\vRvAuYesGainMax}{0.24}
\newcommand{\vWfsReadC}{25.0}
\newcommand{\vWfsTpl}{40.1}
\newcommand{\vWfsFree}{58.6}
\newcommand{\vWfsHiReadC}{45.9}
\newcommand{\vWfsHiTpl}{61.5}
\newcommand{\vWfsHiVl}{123.0}
\newcommand{\vWfmReadC}{32.6}
\newcommand{\vWfmTpl}{54.1}
\newcommand{\vWfmFree}{66.6}
\newcommand{\vWfmHiReadC}{76.6}
\newcommand{\vWfmHiTpl}{98.6}
\newcommand{\vWfmHiVl}{136.7}
\newcommand{\vWflReadC}{49.3}
\newcommand{\vWflTpl}{86.5}
\newcommand{\vWflFree}{115.6}
\newcommand{\vWflVl}{121.1}
\newcommand{\vWflHiReadC}{117.5}
\newcommand{\vWflHiTpl}{154.1}
\newcommand{\vWflHiVl}{180.4}
\newcommand{\vWfxReadC}{65.0}
\newcommand{\vWfxTpl}{116.2}
\newcommand{\vWfxVl}{180.3}
\newcommand{\vWfxVlRatioC}{2.8}
\newcommand{\vWfxHiReadC}{171.1}
\newcommand{\vWfxHiTpl}{220.0}
\newcommand{\vWfxHiVl}{282.5}
\newcommand{\vSpStopsRatioThou}{1.8}
\newcommand{\vSpStopmRatioThou}{1.8}
\newcommand{\vSpStoplRatioThou}{1.5}
\newcommand{\vPcAccOurs}{73.3}
\newcommand{\vPcWrYesOurs}{14.2}
\newcommand{\vPcAccPtsQ}{69.9}
\newcommand{\vPcWrYesPtsQ}{17.0}
\newcommand{\vPcDWrOursQ}{$-$2.8}
\newcommand{\vPcDWrOursQLo}{$-$4.0}
\newcommand{\vPcDWrOursQHi}{$-$1.8}
\newcommand{\vSoSetsLfv}{73.6}
\newcommand{\vSoSetsGem}{76.5}
\newcommand{\vSoSetsIvA}{63.6}
\newcommand{\vSoSetsIvD}{76.0}
\newcommand{\vSoSetsIvH}{76.3}
\newcommand{\vSoSetsQvTwo}{76.6}
\newcommand{\vSoSetsQvFour}{80.7}
\newcommand{\vSoSetsLfm}{77.4}
\newcommand{\vSoSetsLlv}{58.4}
\newcommand{\vSoSetsRex}{76.8}
\newcommand{\vPubJagZeroEight}{88.3}
\newcommand{\vPubFlorenceTwoL}{91.1}
\newcommand{\vLocusPasses}{5.6}
\newcommand{\vVsMjevThouDec}{238.8}
\newcommand{\vFdMjevThouDec}{277.2}
\newcommand{\vPubBestThree}{87.4}
\newcommand{\vPubBestNine}{89.7}
\newcommand{\vFtQmPts}{31.3}
\newcommand{\vFtQmOurs}{11.7}
\newcommand{\vFtQmAct}{63}
\newcommand{\vBcNAll}{5,354}
\newcommand{\vBcNPool}{305}
\newcommand{\vBcTwoAllPt}{73.3}
\newcommand{\vBcTwoAllBx}{66.6}
\newcommand{\vBcTwoAllD}{+6.7}
\newcommand{\vBcTwoAllDLo}{+5.6}
\newcommand{\vBcTwoAllDHi}{+7.7}
\newcommand{\vBcTwoPoolPt}{64.9}
\newcommand{\vBcTwoPoolBx}{31.1}
\newcommand{\vBcTwoPoolD}{+33.8}
\newcommand{\vBcTwoPoolDLo}{+27.5}
\newcommand{\vBcTwoPoolDHi}{+40.0}
\newcommand{\vBcFourAllPt}{79.4}
\newcommand{\vBcFourAllBx}{73.1}
\newcommand{\vBcFourPoolPt}{68.9}
\newcommand{\vBcFourPoolBx}{32.1}
\newcommand{\vWtRand}{10.9}
\newcommand{\vPcImgTok}{144}
\newcommand{\vPcAccQeAll}{82.6}
\newcommand{\vPcSameAll}{96.2}
\newcommand{\vPcNClick}{50}
\newcommand{\vPcNAll}{900}
\newcommand{\vPnbPoolN}{305}
\newcommand{\vPnbKeptN}{100,951}
\newcommand{\vPnbDroppedN}{187,049}
\newcommand{\vPnbNbAcc}{69.6}
\newcommand{\vPnbNbPool}{67.2}
\newcommand{\vPnbNbCen}{7.5}
\newcommand{\vPnbPtPool}{43.3}
\newcommand{\vPnbPtCen}{31.5}
\newcommand{\vPnbOuPool}{64.9}
\newcommand{\vPnbOuCen}{5.6}
\newcommand{\vPnbDOuNbAcc}{+3.7}
\newcommand{\vPnbDOuNbAccLo}{+2.6}
\newcommand{\vPnbDOuNbAccHi}{+4.8}
\newcommand{\vPnbDOuNbCen}{\tm{}2.0}
\newcommand{\vPnbDOuNbCenLo}{\tm{}4.9}
\newcommand{\vPnbDOuNbCenHi}{+0.7}
\newcommand{\vPnbDOuPtCen}{-25.9}
\newcommand{\vPnbDOuPtCenLo}{-31.1}
\newcommand{\vPnbDOuPtCenHi}{-20.7}
\newcommand{\vRnNgN}{4,082}
\newcommand{\vRnNgD}{+1.9}
\newcommand{\vRnNgDLo}{+0.7}
\newcommand{\vRnNgDHi}{+3.0}
\newcommand{\vRgQvlTwoAcc}{74.2}
\newcommand{\vRgQvlTwoW}{20.0}
\newcommand{\vRgQvlTwoAu}{0.67}
\newcommand{\vRgQvlFourAcc}{79.1}
\newcommand{\vRgQvlFourW}{16.5}
\newcommand{\vRgQvlFourAu}{0.64}
\newcommand{\vRgQvlFourDFour}{+0.3}
\newcommand{\vRgQvlFourDFourLo}{\tm0.7}
\newcommand{\vRgQvlFourDFourHi}{+1.4}
\newcommand{\vRgGpiAcc}{77.8}
\newcommand{\vRgGpiW}{19.6}
\newcommand{\vRgGpiAu}{0.59}
\newcommand{\vRgRexAu}{0.55}
\newcommand{\vRgMolmoAu}{0.58}
\newcommand{\vRtRexHf}{603}
\newcommand{\vRtMolmoHf}{1,444}
\newcommand{\vPubSsLow}{84.7}
\newcommand{\vPubSsHigh}{89.7}
\newcommand{\vIsSetsPt}{73.3}
\newcommand{\vSvsPts}{26.4}
\newcommand{\vSvsOurs}{19.2}
\newcommand{\vAusPts}{0.75}
\newcommand{\vAusOurs}{0.82}
\newcommand{\vSvmPts}{20.1}
\newcommand{\vSvmOurs}{14.2}
\newcommand{\vAumPts}{0.75}
\newcommand{\vAumOurs}{0.81}
\newcommand{\vSvlPts}{14.7}
\newcommand{\vSvlOurs}{9.3}
\newcommand{\vAulPts}{0.75}
\newcommand{\vAulOurs}{0.82}
\newcommand{\vSvivlAPts}{29.1}
\newcommand{\vSvivlAOurs}{23.0}
\newcommand{\vAuivlAPts}{0.77}
\newcommand{\vAuivlAOurs}{0.81}
\newcommand{\vSvivlPts}{20.2}
\newcommand{\vSvivlOurs}{15.0}
\newcommand{\vAuivlPts}{0.75}
\newcommand{\vAuivlOurs}{0.80}
\newcommand{\vSvivlDPts}{17.7}
\newcommand{\vSvivlDOurs}{12.3}
\newcommand{\vAuivlDPts}{0.74}
\newcommand{\vAuivlDOurs}{0.81}
\newcommand{\vSvgemPts}{21.3}
\newcommand{\vSvgemOurs}{12.6}
\newcommand{\vAugemPts}{0.63}
\newcommand{\vAugemOurs}{0.79}
\newcommand{\vSvlfmPts}{16.9}
\newcommand{\vSvlfmOurs}{10.8}
\newcommand{\vAulfmPts}{0.74}
\newcommand{\vAulfmOurs}{0.80}
\newcommand{\vSvlfvPts}{18.2}
\newcommand{\vSvlfvOurs}{13.5}
\newcommand{\vAulfvPts}{0.76}
\newcommand{\vAulfvOurs}{0.81}
\newcommand{\vDNGraphTwoSame}{60}
\newcommand{\vDNGraphNineSame}{56}
\newcommand{\vDNGraphNineN}{60}
\newcommand{\vfAgentN}{9,090}
\newcommand{\vfTypedMixN}{354,825}
\newcommand{\vMwSecNsFinClick}{52.5}
\newcommand{\vMwSecNsSFTClick}{38.4}
\newcommand{\vMwSecNmFinClick}{66.3}
\newcommand{\vMwSecNmSFTClick}{57.3}
\newcommand{\vMwSecNlFinClick}{74.7}
\newcommand{\vMwSecNlSFTClick}{58.9}
\newcommand{\vMwLostLAll}{144}
\newcommand{\vMwWonLAll}{90}
\newcommand{\vDrEpis}{80}
\newcommand{\vDrReachQJGpu}{20}
\newcommand{\vDrReachQJPi}{11}
\newcommand{\vDrReachQDGpu}{20}
\newcommand{\vDrMsQDGpu}{40}
\newcommand{\vDrReachQDPi}{22}
\newcommand{\vDrReachCJGpu}{26}
\newcommand{\vDrReachCJPi}{5}
\newcommand{\vDrReachCDGpu}{29}
\newcommand{\vDrMsCDGpu}{77}
\newcommand{\vDrReachCDPi}{11}
\newcommand{\vDrReachQJPiS}{4.0}
\newcommand{\vDrRangeQJPiS}{4--5}
\newcommand{\vDrReachQDPiS}{11.0}
\newcommand{\vDrRangeQDPiS}{9--13}
\newcommand{\vDrReachCJPiS}{2.3}
\newcommand{\vDrReachCDPiS}{3.0}
\newcommand{\vDrNearQDGpu}{21}
\newcommand{\vDrNearQDPi}{33}
\newcommand{\vDrNearQDPiS}{22.7}
\newcommand{\vDrNearCDGpu}{30}
\newcommand{\vDrNearCDPi}{13}
\newcommand{\vDrNearCDPiS}{8.3}
\newcommand{\vDrMovEpis}{40}
\newcommand{\vDrScAppQDPiS}{2.7}
\newcommand{\vDrScSeaQDPiS}{2.3}
\newcommand{\vDrScTraQDPiS}{4.0}
\newcommand{\vDrScSetQDPiS}{2.0}
\newcommand{\vDrScAppCDPiS}{2.3}
\newcommand{\vDrScSeaCDPiS}{0.7}
\newcommand{\vDrScTraCDPiS}{0.0}
\newcommand{\vDrScSetCDPiS}{0.0}
\newcommand{\vDrScEpis}{20}
\newcommand{\vPiImgQBase}{32}
\newcommand{\vPiImgCBase}{72}
\newcommand{\vPiInQJNat}{8.4}
\newcommand{\vPiInQDNat}{6.5}
\newcommand{\vPiImgQNat}{220}
\newcommand{\vPiInCJNat}{28.2}
\newcommand{\vPiInCDNat}{23.0}
\newcommand{\vPiImgCNat}{220}
\newcommand{\vPiSpreadMax}{2}
\newcommand{\vPiMeanQJ}{4.63}
\newcommand{\vPiSdQJ}{0.02}
\newcommand{\vPiMeanQD}{2.14}
\newcommand{\vPiSdQD}{0.01}
\newcommand{\vPiMeanCJ}{16.94}
\newcommand{\vPiSdCJ}{0.18}
\newcommand{\vPiMeanCD}{9.04}
\newcommand{\vPiSdCD}{0.12}
\newcommand{\vAuStopS}{65.4}
\newcommand{\vAuStopM}{83.3}
\newcommand{\vAuStopMTok}{10.5}
\newcommand{\vAuStopL}{120.2}
\newcommand{\vAuStopLTok}{12.2}
\newcommand{\vPqClickS}{4.0}
\newcommand{\vPqHundS}{4.7}
\newcommand{\vPqThouS}{11.5}
\newcommand{\vPqClickWarmS}{1.07}
\newcommand{\vOssCell}{25.0}
\newcommand{\vOssHund}{25.8}
\newcommand{\vOssThou}{36.1}
\newcommand{\vOsmHund}{33.3}
\newcommand{\vOsmThou}{45.6}
\newcommand{\vOslHund}{50.5}
\newcommand{\vOslThou}{79.0}
\newcommand{\vOsxThou}{101.2}
\newcommand{\vOssHiThou}{58.2}
\newcommand{\vOssHiVlRatioThou}{2.1}
\newcommand{\vOsmHiThou}{91.6}
\newcommand{\vOsmHiVlRatioThou}{1.5}
\newcommand{\vOslHiThou}{150.4}
\newcommand{\vOslHiVlRatioThou}{1.2}
\newcommand{\vOsxVlRatioThou}{1.8}
\newcommand{\vOsxHiThou}{208.0}
\newcommand{\vOsxHiVlRatioThou}{1.4}
\newcommand{\vOsHivlACell}{99}
\newcommand{\vOsHivlAThou}{99}
\newcommand{\vOsHivlARatioThou}{3.4}
\newcommand{\vOsHivlCell}{95}
\newcommand{\vOsHivlThou}{96}
\newcommand{\vOsHivlRatioThou}{3.6}
\newcommand{\vOsHivlDCell}{121}
\newcommand{\vOsHivlDThou}{121}
\newcommand{\vOsHivlDRatioThou}{4.0}
\newcommand{\vOsHivlHThou}{123}
\newcommand{\vOsHivlHRatioThou}{3.6}
\newcommand{\vOsHllvThou}{112}
\newcommand{\vOsHllvRatioThou}{3.6}
\newcommand{\vOsFpMaxDp}{$3{\times}10^{-5}$}
\newcommand{\vLyLayPtS}{+1.7}
\newcommand{\vLyLayPtSLo}{+0.7}
\newcommand{\vLyLayPtSHi}{+2.7}
\newcommand{\vLyLabPtS}{+3.4}
\newcommand{\vLyLabPtSLo}{+2.3}
\newcommand{\vLyLabPtSHi}{+4.4}
\newcommand{\vLyLayPtM}{+1.6}
\newcommand{\vLyLayPtMLo}{+0.7}
\newcommand{\vLyLayPtMHi}{+2.5}
\newcommand{\vLyLabPtM}{+1.9}
\newcommand{\vLyLabPtMLo}{+0.9}
\newcommand{\vLyLabPtMHi}{+3.0}
\newcommand{\vLyLayPtL}{+0.4}
\newcommand{\vLyLayPtLLo}{\tm{}0.4}
\newcommand{\vLyLayPtLHi}{+1.2}
\newcommand{\vLyLabPtL}{+2.0}
\newcommand{\vLyLabPtLLo}{+1.0}
\newcommand{\vLyLabPtLHi}{+2.9}
\newcommand{\vWbMwTaskN}{1{,}256}
\newcommand{\vWbMwSiteN}{973}
\newcommand{\vWbMwDomN}{3{,}827}
\newcommand{\vWbMwBaseZeroEightAcc}{6.0}
\newcommand{\vWbMwPtsZeroEightAcc}{73.5}
\newcommand{\vWbMwOursZeroEightAcc}{73.6}
\newcommand{\vWbMwDiffZeroEight}{$+$0.1}
\newcommand{\vWbMwDiffZeroEightLo}{$-$0.8}
\newcommand{\vWbMwDiffZeroEightHi}{$+$1.0}
\newcommand{\vWbMwOursZeroEightWrong}{11.7}
\newcommand{\vWbMwPtsZeroEightWrong}{14.0}
\newcommand{\vWbMwWrongDiffZeroEightLo}{$-$3.4}
\newcommand{\vWbMwWrongDiffZeroEightHi}{$-$1.3}
\newcommand{\vWbMwOursZeroEightAuc}{0.86}
\newcommand{\vWbMwPtsZeroEightAuc}{0.80}
\newcommand{\vWbMwAucDiffZeroEightLo}{$+$0.04}
\newcommand{\vWbMwAucDiffZeroEightHi}{$+$0.07}
\newcommand{\vWbMwBaseTwoAcc}{52.1}
\newcommand{\vWbMwPtsTwoAcc}{77.4}
\newcommand{\vWbMwOursTwoAcc}{78.8}
\newcommand{\vWbMwDiffTwo}{$+$1.4}
\newcommand{\vWbMwDiffTwoLo}{$+$0.6}
\newcommand{\vWbMwDiffTwoHi}{$+$2.3}
\newcommand{\vWbMwOursTwoWrong}{5.8}
\newcommand{\vWbMwPtsTwoWrong}{9.8}
\newcommand{\vWbMwWrongDiffTwoLo}{$-$4.9}
\newcommand{\vWbMwWrongDiffTwoHi}{$-$3.1}
\newcommand{\vWbMwOursTwoAuc}{0.89}
\newcommand{\vWbMwPtsTwoAuc}{0.83}
\newcommand{\vWbMwAucDiffTwoLo}{$+$0.05}
\newcommand{\vWbMwAucDiffTwoHi}{$+$0.07}
\newcommand{\vWbMwBaseFourAcc}{78.6}
\newcommand{\vWbMwPtsFourAcc}{79.0}
\newcommand{\vWbMwOursFourAcc}{79.7}
\newcommand{\vWbMwDiffFour}{$+$0.8}
\newcommand{\vWbMwDiffFourLo}{0.0}
\newcommand{\vWbMwDiffFourHi}{$+$1.6}
\newcommand{\vWbMwOursFourWrong}{5.8}
\newcommand{\vWbMwPtsFourWrong}{9.4}
\newcommand{\vWbMwWrongDiffFourLo}{$-$4.3}
\newcommand{\vWbMwWrongDiffFourHi}{$-$2.6}
\newcommand{\vWbMwOursFourAuc}{0.89}
\newcommand{\vWbMwPtsFourAuc}{0.83}
\newcommand{\vWbMwAucDiffFourLo}{$+$0.05}
\newcommand{\vWbMwAucDiffFourHi}{$+$0.08}
\newcommand{\vQvTwoOursPt}{76.2}
\newcommand{\vQvTwoPtsPt}{72.8}
\newcommand{\vQvTwoBasePt}{74.2}
\newcommand{\vQvTwoOursCell}{84.1}
\newcommand{\vQvTwoPtsCell}{82.2}
\newcommand{\vQvTwoOursPool}{65.6}
\newcommand{\vQvTwoPtsPool}{44.6}
\newcommand{\vQvTwoOursAtc}{4.9}
\newcommand{\vQvTwoPtsAtc}{29.5}
\newcommand{\vQvTwoD}{+3.4}
\newcommand{\vQvTwoDLo}{+2.4}
\newcommand{\vQvTwoDHi}{+4.3}
\newcommand{\vQvTwoDCell}{+2.0}
\newcommand{\vQvTwoDCellLo}{+1.2}
\newcommand{\vQvTwoDCellHi}{+2.8}
\newcommand{\vQvTwoDPool}{+21.0}
\newcommand{\vQvTwoDPoolLo}{+14.8}
\newcommand{\vQvTwoDPoolHi}{+27.2}
\newcommand{\vQvTwoOursWrong}{10.8}
\newcommand{\vQvTwoPtsWrong}{17.0}
\newcommand{\vQvTwoBaseWrong}{20.0}
\newcommand{\vQvTwoOursAu}{0.82}
\newcommand{\vQvTwoPtsAu}{0.75}
\newcommand{\vQvTwoBaseAu}{0.67}
\newcommand{\vQvTwoDPtsBase}{\tm{}1.3}
\newcommand{\vQvTwoDPtsBaseLo}{\tm{}2.4}
\newcommand{\vQvTwoDPtsBaseHi}{\tm{}0.3}
\newcommand{\vQvFourOursPt}{80.3}
\newcommand{\vQvFourPtsPt}{78.4}
\newcommand{\vQvFourBasePt}{79.1}
\newcommand{\vQvFourOursCell}{87.3}
\newcommand{\vQvFourPtsCell}{85.7}
\newcommand{\vQvFourOursPool}{69.2}
\newcommand{\vQvFourPtsPool}{47.9}
\newcommand{\vQvFourOursAtc}{7.2}
\newcommand{\vQvFourPtsAtc}{30.2}
\newcommand{\vQvFourD}{+1.9}
\newcommand{\vQvFourDLo}{+1.1}
\newcommand{\vQvFourDHi}{+2.8}
\newcommand{\vQvFourDCell}{+1.6}
\newcommand{\vQvFourDCellLo}{+0.9}
\newcommand{\vQvFourDCellHi}{+2.2}
\newcommand{\vQvFourDPool}{+21.3}
\newcommand{\vQvFourDPoolLo}{+15.1}
\newcommand{\vQvFourDPoolHi}{+27.5}
\newcommand{\vQvFourOursWrong}{8.9}
\newcommand{\vQvFourPtsWrong}{13.8}
\newcommand{\vQvFourBaseWrong}{16.5}
\newcommand{\vQvFourOursAu}{0.81}
\newcommand{\vQvFourPtsAu}{0.72}
\newcommand{\vQvFourBaseAu}{0.64}
\newcommand{\vQvFourDPtsBase}{\tm{}0.7}
\newcommand{\vQvFourDPtsBaseLo}{\tm{}1.6}
\newcommand{\vQvFourDPtsBaseHi}{+0.2}
\newcommand{\vElGainMin}{1.9}
\newcommand{\vElGainMax}{4.5}
\newcommand{\vMxSFTPair}{62.7}
\newcommand{\vMxFinPair}{62.5}
\newcommand{\vMivlASFTPair}{44.5}
\newcommand{\vMivlAFinPair}{45.3}
\newcommand{\vMivlSFTPair}{50.8}
\newcommand{\vMivlFinPair}{50.2}
\newcommand{\vMivlDSFTPair}{55.2}
\newcommand{\vMivlDFinPair}{54.9}
\newcommand{\vMivlHSFTPair}{56.7}
\newcommand{\vMivlHFinPair}{56.8}
\newcommand{\vMgemSFTPair}{57.1}
\newcommand{\vMgemFinPair}{54.2}
\newcommand{\vMlfmSFTPair}{55.2}
\newcommand{\vMlfmFinPair}{53.6}
\newcommand{\vMllvSFTPair}{40.4}
\newcommand{\vMllvFinPair}{36.9}
\newcommand{\vMlfvSFTPair}{56.1}
\newcommand{\vMlfvFinPair}{55.8}
\newcommand{\vMsSFTPair}{45.5}
\newcommand{\vMsFinPair}{43.5}
\newcommand{\vMmSFTPair}{54.5}
\newcommand{\vMmFinPair}{52.8}
\newcommand{\vMmNPair}{20}
\newcommand{\vMlSFTPair}{58.7}
\newcommand{\vMlFinPair}{59.4}
\newcommand{\vSsSS}{86.0}
\newcommand{\vSsWP}{57.0}
\newcommand{\vSsRRbox}{78.1}
\newcommand{\vSsSSLo}{55.7}
\newcommand{\vSsMWft}{44.7}
\newcommand{\vSmSS}{90.4}
\newcommand{\vSmPBSpatial}{74.9}
\newcommand{\vSmPBAvg}{53.7}
\newcommand{\vSmWP}{63.1}
\newcommand{\vSmRRbox}{84.6}
\newcommand{\vSmSSLo}{67.0}
\newcommand{\vSmMWft}{61.7}
\newcommand{\vSlSS}{92.1}
\newcommand{\vSlPBSpatial}{75.4}
\newcommand{\vSlPBAvg}{57.4}
\newcommand{\vSlWP}{69.0}
\newcommand{\vSlRRbox}{85.3}
\newcommand{\vSlSSLo}{74.4}
\newcommand{\vSlMWft}{66.4}
\newcommand{\vSivlSS}{80.2}
\newcommand{\vSivlWP}{51.1}
\newcommand{\vSivlRRbox}{71.7}
\newcommand{\vSivlSSLo}{70.4}
\newcommand{\vSivlMWft}{67.4}
\newcommand{\vSbsSS}{14.3}
\newcommand{\vSbsSSLo}{7.5}
\newcommand{\vSbmSS}{55.0}
\newcommand{\vSbmWP}{17.0}
\newcommand{\vSbmRRbox}{63.8}
\newcommand{\vSbmSSLo}{26.6}
\newcommand{\vSbmMWft}{15.4}
\newcommand{\vSblSS}{86.6}
\newcommand{\vSblWP}{49.1}
\newcommand{\vSblRRbox}{84.7}
\newcommand{\vSblSSLo}{56.0}
\newcommand{\vSblMWft}{33.7}
\newcommand{\vHdXPtDelta}{\tm{}1.4}
\newcommand{\vHdRoReg}{1.36}
\newcommand{\vHdReg}{1.60}
\newcommand{\vPoolRef}{111}
\newcommand{\vPoolPb}{185}
\newcommand{\vPoolWp}{9}
\newcommand{\vAtCentreTol}{3}
\newcommand{\vSnAccSOursOrig}{67.0}
\newcommand{\vSnWrSOursOrig}{19.2}
\newcommand{\vSnAuSOursOrig}{0.82}
\newcommand{\vSnAccSOursSeedA}{67.4}
\newcommand{\vSnAccSOursNa}{66.7}
\newcommand{\vSnWrSOursNa}{19.1}
\newcommand{\vSnAuSOursNa}{0.81}
\newcommand{\vSnAccSPtsOrig}{62.0}
\newcommand{\vSnWrSPtsOrig}{26.4}
\newcommand{\vSnAuSPtsOrig}{0.75}
\newcommand{\vSnAccSPtsSeedA}{63.5}
\newcommand{\vSnAccSPtsNa}{62.7}
\newcommand{\vSnWrSPtsNa}{25.2}
\newcommand{\vSnAuSPtsNa}{0.77}
\newcommand{\vSnAccMOursOrig}{73.3}
\newcommand{\vSnWrMOursOrig}{14.2}
\newcommand{\vSnAuMOursOrig}{0.81}
\newcommand{\vSnAccMOursSeedA}{73.7}
\newcommand{\vSnAccMOursSeedB}{73.1}
\newcommand{\vSnAccMOursNa}{73.5}
\newcommand{\vSnWrMOursNa}{13.6}
\newcommand{\vSnAuMOursNa}{0.81}
\newcommand{\vSnAccMPtsOrig}{68.7}
\newcommand{\vSnWrMPtsOrig}{20.1}
\newcommand{\vSnAuMPtsOrig}{0.75}
\newcommand{\vSnAccMPtsSeedA}{69.6}
\newcommand{\vSnAccMPtsSeedB}{69.9}
\newcommand{\vSnAccMPtsNa}{69.0}
\newcommand{\vSnWrMPtsNa}{20.0}
\newcommand{\vSnAuMPtsNa}{0.75}
\newcommand{\vSnAccLOursOrig}{79.4}
\newcommand{\vSnWrLOursOrig}{9.3}
\newcommand{\vSnAuLOursOrig}{0.82}
\newcommand{\vSnAccLOursSeedA}{79.1}
\newcommand{\vSnAccLOursNa}{79.2}
\newcommand{\vSnWrLOursNa}{8.6}
\newcommand{\vSnAuLOursNa}{0.83}
\newcommand{\vSnAccLPtsOrig}{75.7}
\newcommand{\vSnWrLPtsOrig}{14.7}
\newcommand{\vSnAuLPtsOrig}{0.75}
\newcommand{\vSnAccLPtsSeedA}{76.4}
\newcommand{\vSnAccLPtsNa}{76.1}
\newcommand{\vSnWrLPtsNa}{14.1}
\newcommand{\vSnAuLPtsNa}{0.76}
\newcommand{\vSnSeedSpreadM}{0.6}
\newcommand{\vSnNaAccMin}{\tm{}0.3}
\newcommand{\vSnNaAccMax}{+0.6}
\newcommand{\vSnNaSOursAcc}{\tm{}0.3}
\newcommand{\vSnNaSOursAccLo}{\tm{}1.3}
\newcommand{\vSnNaSOursAccHi}{+0.7}
\newcommand{\vSnNaSPtsAcc}{+0.6}
\newcommand{\vSnNaSPtsAccLo}{\tm{}0.4}
\newcommand{\vSnNaSPtsAccHi}{+1.7}
\newcommand{\vSnNaMPtsAcc}{+0.3}
\newcommand{\vSnNaMPtsAccLo}{\tm{}0.6}
\newcommand{\vSnNaMPtsAccHi}{+1.2}
\newcommand{\vSnNaLPtsAcc}{+0.4}
\newcommand{\vSnNaLPtsAccLo}{\tm{}0.4}
\newcommand{\vSnNaLPtsAccHi}{+1.2}
\newcommand{\vSnNaLOursAcc}{\tm{}0.1}
\newcommand{\vSnNaLOursAccLo}{\tm{}0.9}
\newcommand{\vSnNaLOursAccHi}{+0.6}
\newcommand{\vSnNaMOursAcc}{+0.3}
\newcommand{\vSnNaMOursAccLo}{\tm{}0.7}
\newcommand{\vSnNaMOursAccHi}{+1.2}
\providecommand{\vSetDcSteps}{3{,}200}
\providecommand{\vSetDcBatch}{96}
\providecommand{\vSetDcPaperMammoth}{90}
\providecommand{\vSetEdgeTol}{10}
\providecommand{\vSetEdgeMin}{10}
\providecommand{\vSetBoot}{10{,}000}
\providecommand{\vSetPiFrame}{256}
\providecommand{\vSetReachLim}{2}
\providecommand{\vSetTrainSteps}{6{,}000}
\providecommand{\vSetMaxTok}{64}
\providecommand{\vSetPiSeqs}{128}
\providecommand{\vSetVllmBlock}{544}
\providecommand{\vSetPiFramesN}{50}
\providecommand{\vSetPcPerBench}{300}
\providecommand{\vSetCountCap}{19.4}
\providecommand{\vSetOneMP}{1{,}003{,}520}
\providecommand{\vSetRcValN}{10{,}834}
\newcommand{\vSetMwTasks}{15}
\newcommand{\vVgQvlTwoFasterVl}{1.8}
\newcommand{\vVgQvlFourFasterVl}{1.6}
\newcommand{\vRwAbsMax}{1.0}
\newcommand{\vUtAxmDiffPlain}{+5.0}
\newcommand{\vUtAxlDiffPlain}{+18.7}
\newcommand{\vHdXPtLo}{\tm{}2.4}
\newcommand{\vHdXPtHi}{\tm{}0.4}
\providecommand{\vSdQDPiS}{2.0}
\providecommand{\vSdQJPiS}{0.6}
\providecommand{\vSdCJPiS}{1.0}
\providecommand{\vSdCDPiS}{1.0}
\providecommand{\vQvTwoOverOts}{+2.0}
\newcommand{\vRzOursSCellAcc}{79.5}
\newcommand{\vRzOursSHundAcc}{69.4}
\newcommand{\vRzOursSPtAcc}{67.0}
\newcommand{\vRzOursMCellAcc}{83.9}
\newcommand{\vRzOursMHundAcc}{76.5}
\newcommand{\vRzOursMPtAcc}{73.3}
\newcommand{\vRzOursLCellAcc}{86.8}
\newcommand{\vRzOursLHundAcc}{81.2}
\newcommand{\vRzOursLPtAcc}{79.4}
\newcommand{\vRzPtsSPtAcc}{62.0}
\newcommand{\vRzPtsMPtAcc}{68.7}
\newcommand{\vRzPtsLPtAcc}{75.7}
\newcommand{\vRzGuidCellAcc}{75.9}
\newcommand{\vRzGuidHundAcc}{62.4}
\newcommand{\vRzGuidPtAcc}{58.0}
\newcommand{\vRzGuidPtFP}{34.2}
\newcommand{\vRzGuidPtAu}{0.65}
\newcommand{\vRzMjevCellAcc}{83.4}
\newcommand{\vRzMjevHundAcc}{78.4}
\newcommand{\vRzMjevPtAcc}{76.2}
\newcommand{\vRzMjevPtFP}{17.5}
\newcommand{\vRzMjevPtAu}{0.68}
\newcommand{\vRzDlCellAcc}{30.3}
\newcommand{\vRzDlHundAcc}{15.2}
\newcommand{\vRzDlPtAcc}{13.4}
\newcommand{\vRzDlPtFP}{86.6}
\newcommand{\vRzDlPtAu}{0.50}
\newcommand{\vRzDnCellAcc}{37.4}
\newcommand{\vRzDnHundAcc}{17.2}
\newcommand{\vRzDnPtAcc}{14.5}
\newcommand{\vRzDnPtFP}{85.6}
\newcommand{\vRzDnPtAu}{0.46}
\newcommand{\vRzIntSCellAcc}{21.1}
\newcommand{\vRzIntSHundAcc}{9.1}
\newcommand{\vRzIntSPtAcc}{7.4}
\newcommand{\vRzIntSPtFP}{93.4}
\newcommand{\vRzIntSPtAu}{0.40}
\newcommand{\vRzIntMCellAcc}{21.0}
\newcommand{\vRzIntMHundAcc}{8.4}
\newcommand{\vRzIntMPtAcc}{6.9}
\newcommand{\vRzIntMPtFP}{92.6}
\newcommand{\vRzIntMPtAu}{0.49}
\newcommand{\vRzIntLCellAcc}{22.2}
\newcommand{\vRzIntLHundAcc}{7.7}
\newcommand{\vRzIntLPtAcc}{6.4}
\newcommand{\vRzIntLPtFP}{94.4}
\newcommand{\vRzIntLPtAu}{0.40}
\newcommand{\vRzPpCellAcc}{40.6}
\newcommand{\vRzPpHundAcc}{15.6}
\newcommand{\vRzPpPtAcc}{12.6}
\newcommand{\vRzPpPtFP}{90.2}
\newcommand{\vRzPpPtAu}{0.36}
\newcommand{\vRzOjfCellAcc}{12.3}
\newcommand{\vRzOjfHundAcc}{4.3}
\newcommand{\vRzOjfPtAcc}{3.2}
\newcommand{\vRzOjfPtFP}{97.6}
\newcommand{\vRzOjfPtAu}{0.35}
\newcommand{\vRzOjCellAcc}{18.0}
\newcommand{\vRzOjHundAcc}{6.6}
\newcommand{\vRzOjPtAcc}{5.1}
\newcommand{\vRzOjPtFP}{95.5}
\newcommand{\vRzOjPtAu}{0.45}
\newcommand{\vRzOjPtMs}{2002}
\newcommand{\vRzJpCellAcc}{17.4}
\newcommand{\vRzJpHundAcc}{5.9}
\newcommand{\vRzJpPtAcc}{4.2}
\newcommand{\vRzJpPtFP}{96.3}
\newcommand{\vRzJpPtAu}{0.42}
\newcommand{\vRzJpPtMs}{2037}
\newcommand{\vRzTaCellAcc}{24.1}
\newcommand{\vRzTaHundAcc}{7.9}
\newcommand{\vRzTaPtAcc}{6.1}
\newcommand{\vRzTaPtFP}{95.4}
\newcommand{\vRzTaPtAu}{0.33}
\newcommand{\vRzTaPtMs}{2243}
\newcommand{\vRzImCellAcc}{16.1}
\newcommand{\vRzImHundAcc}{5.8}
\newcommand{\vRzImPtAcc}{4.6}
\newcommand{\vRzImPtFP}{96.2}
\newcommand{\vRzImPtAu}{0.41}
\newcommand{\vRzImPtMs}{2228}
\newcommand{\vRzKkCellAcc}{12.0}
\newcommand{\vRzKkHundAcc}{4.6}
\newcommand{\vRzKkPtAcc}{3.6}
\newcommand{\vRzKkPtFP}{96.7}
\newcommand{\vRzKkPtAu}{0.44}
\newcommand{\vRzKkPtMs}{521}
\newcommand{\vRzNatSCellAcc}{68.9}
\newcommand{\vRzNatSHundAcc}{52.9}
\newcommand{\vRzNatMCellAcc}{75.0}
\newcommand{\vRzNatMHundAcc}{61.2}
\newcommand{\vRzNatMPtFP}{34.2}
\newcommand{\vRzNatMPtAu}{0.69}
\newcommand{\vRzNatLCellAcc}{83.1}
\newcommand{\vRzNatLHundAcc}{74.6}
\newcommand{\vRzNatLPtFP}{19.7}
\newcommand{\vRzNatLPtAu}{0.69}
\newcommand{\vRzNatReadSPtFP}{41.5}
\newcommand{\vRzNatReadSPtAu}{0.71}
\newcommand{\vRzNatReadMPtFP}{33.8}
\newcommand{\vRzNatReadMPtAu}{0.70}
\newcommand{\vRzNatReadLPtFP}{19.7}
\newcommand{\vRzNatReadLPtAu}{0.70}
\newcommand{\vRzQvlTwoOursCellAcc}{84.1}
\newcommand{\vRzQvlTwoOursHundAcc}{77.7}
\newcommand{\vRzQvlTwoOursPtAcc}{76.2}
\newcommand{\vRzQvlFourOursCellAcc}{87.3}
\newcommand{\vRzQvlFourOursHundAcc}{81.7}
\newcommand{\vRzQvlFourOursPtAcc}{80.3}
\newcommand{\vRzQvlTwoPtsPtAcc}{72.8}
\newcommand{\vRzQvlFourPtsPtAcc}{78.4}
\newcommand{\vRzOptCellMax}{40.6}
\newcommand{\vRzOptPtMax}{14.5}
\newcommand{\vRzDOursMGuidHund}{+14.1}
\newcommand{\vRzDOursMGuidHundLo}{+12.8}
\newcommand{\vRzDOursMGuidHundHi}{+15.4}
\newcommand{\vRzDOursMGuidPt}{+15.3}
\newcommand{\vRzDOursMGuidPtLo}{+13.9}
\newcommand{\vRzDOursMGuidPtHi}{+16.6}
\newcommand{\vRzDOursMMjevPt}{\tm3.0}
\newcommand{\vRzDOursMMjevPtLo}{\tm4.2}
\newcommand{\vRzDOursMMjevPtHi}{\tm1.7}
\newcommand{\vRzDOursMQvlTwoHund}{+0.4}
\newcommand{\vRzDOursMQvlTwoHundLo}{\tm0.8}
\newcommand{\vRzDOursMQvlTwoHundHi}{+1.5}
\newcommand{\vRwQSRead}{48.8}
\newcommand{\vRwQSWrite}{49.9}
\newcommand{\vRwQMRead}{57.9}
\newcommand{\vRwQMWrite}{58.2}
\newcommand{\vRwQLRead}{71.9}
\newcommand{\vRwQLWrite}{72.4}
\newcommand{\vRwQvlTwoRead}{73.8}
\newcommand{\vRwQvlTwoWrite}{74.2}
\newcommand{\vRwQvlFourRead}{78.6}
\newcommand{\vRwQvlFourWrite}{79.1}
\newcommand{\vRwGpiRead}{77.8}
\newcommand{\vRwGpiWrite}{77.8}
\newcommand{\vRwRexRead}{76.7}
\newcommand{\vRwRexWrite}{76.7}
\newcommand{\vRwMolmoRead}{75.3}
\newcommand{\vRwMolmoWrite}{76.0}
\newcommand{\vWbRelGuidAcc}{69.2}
\newcommand{\vWbRelGuidWrong}{22.6}
\newcommand{\vWbRelGpiAcc}{78.5}
\newcommand{\vWbRelGpiWrong}{10.1}
\newcommand{\vWbRelMjevAcc}{65.4}
\newcommand{\vWbRelMjevWrong}{27.2}
\newcommand{\vUtQvlTwoPtsVl}{69}
\newcommand{\vUtQvlTwoPtsVlTok}{9.2}
\newcommand{\vUtQvlFourPtsVl}{102}
\newcommand{\vUtQvlFourPtsVlTok}{9.0}
\newcommand{\vWsStopmTen}{72.4}
\newcommand{\vWsStopmHund}{75.8}
\newcommand{\vWsTplmTen}{41.6}
\newcommand{\vWsTplmHund}{48.2}
\newcommand{\vWsStoplTen}{96.6}
\newcommand{\vWsStoplHund}{104.0}
\newcommand{\vWsTpllTen}{66.0}
\newcommand{\vWsTpllHund}{76.3}
\newcommand{\vFtQvlTwoVl}{115}
\newcommand{\vFtQvlTwoTok}{20.7}
\newcommand{\vFtQvlFourVl}{167}
\newcommand{\vFtQvlFourTok}{18.5}
\newcommand{\vFtGpiVl}{143}
\newcommand{\vFtGpiTok}{13.6}
\newcommand{\vWtSafePtsZeroEight}{62.8}
\newcommand{\vWtSafeOursZeroEight}{68.4}
\newcommand{\vWtSafePtsTwo}{70.7}
\newcommand{\vWtSafeOursTwo}{81.2}
\newcommand{\vWtSafePtsFour}{73.0}
\newcommand{\vWtSafeOursFour}{82.5}
\newcommand{\vWtSafeGuid}{45.8}
\newcommand{\vWtSafeGpi}{70.5}
\newcommand{\vWtSafeMjev}{37.9}
\newcommand{\vVgQvlTwoCell}{31.2}
\newcommand{\vVgQvlTwoHund}{32.4}
\newcommand{\vVgQvlTwoThou}{37.7}
\newcommand{\vVgQvlTwoStMaxDp}{0.0097}
\newcommand{\vVgQvlFourCell}{47.2}
\newcommand{\vVgQvlFourHund}{51.2}
\newcommand{\vVgQvlFourThou}{62.8}
\newcommand{\vVgQvlFourStMaxDp}{0.0108}
\newcommand{\vFdGuidCell}{24.1}
\newcommand{\vFdGuidHund}{24.1}
\newcommand{\vFdGuidThou}{24.1}
\newcommand{\vFdDnCell}{62}
\newcommand{\vFdDnCellOld}{160}
\newcommand{\vFdDnHund}{124}
\newcommand{\vFdDnThou}{187}
\newcommand{\vFdDnAgree}{1230/1230}
\newcommand{\vFdDnBatchAgree}{1169/1230}
\newcommand{\vFdPpCell}{80}
\newcommand{\vFdPpCellOld}{188}
\newcommand{\vFdPpHund}{160}
\newcommand{\vFdPpThou}{241}
\newcommand{\vFdPpBatchAgree}{1158/1230}
\newcommand{\vFdDlCell}{46}
\newcommand{\vFdDlHund}{92}
\newcommand{\vFdDlThou}{139}
\newcommand{\vFdIntSCell}{89}
\newcommand{\vFdIntSHund}{179}
\newcommand{\vFdIntSThou}{270}
\newcommand{\vFdIntMCell}{108}
\newcommand{\vFdIntMHund}{216}
\newcommand{\vFdIntMThou}{325}
\newcommand{\vFdIntLCell}{165}
\newcommand{\vFdIntLHund}{333}
\newcommand{\vFdIntLThou}{503}
\newcommand{\vFdOjfCell}{134}
\newcommand{\vFdOjfHund}{267}
\newcommand{\vFdOjfThou}{400}
\newcommand{\vFdMjevThou}{285}
\newcommand{\vFdQvlTwoTplTen}{42.8}
\newcommand{\vFdQvlTwoTplHund}{48.3}
\newcommand{\vFdQvlTwoTplThou}{54.5}
\newcommand{\vFdQvlFourTplTen}{64.3}
\newcommand{\vFdQvlFourTplHund}{74.0}
\newcommand{\vFdQvlFourTplThou}{84.3}
\newcommand{\vNsQJGpuApp}{5}
\newcommand{\vNsQJGpuSea}{5}
\newcommand{\vNsQJGpuTra}{5}
\newcommand{\vNsQJGpuSet}{5}
\newcommand{\vNsQJGpuReach}{20}
\newcommand{\vNsQJGpuNear}{23}
\newcommand{\vNsQJGpuStep}{208}
\newcommand{\vNsQJPiApp}{4}
\newcommand{\vNsQJPiSea}{2}
\newcommand{\vNsQJPiTra}{4}
\newcommand{\vNsQJPiSet}{3}
\newcommand{\vNsQJPiReach}{13}
\newcommand{\vNsQJPiNear}{23}
\newcommand{\vNsQJPiRuns}{--}
\newcommand{\vNsQJPiSApp}{1.3}
\newcommand{\vNsQJPiSSea}{1.3}
\newcommand{\vNsQJPiSTra}{1.0}
\newcommand{\vNsQJPiSSet}{1.0}
\newcommand{\vNsQJPiSReach}{4.7}
\newcommand{\vNsQJPiSNear}{7.0}
\newcommand{\vNsQJPiSRuns}{5, 5, 4}
\newcommand{\vNsCJGpuApp}{7}
\newcommand{\vNsCJGpuSea}{7}
\newcommand{\vNsCJGpuTra}{8}
\newcommand{\vNsCJGpuSet}{8}
\newcommand{\vNsCJGpuReach}{30}
\newcommand{\vNsCJGpuNear}{31}
\newcommand{\vNsCJGpuStep}{158}
\newcommand{\vNsCJPiApp}{4}
\newcommand{\vNsCJPiSea}{3}
\newcommand{\vNsCJPiTra}{0}
\newcommand{\vNsCJPiSet}{0}
\newcommand{\vNsCJPiReach}{7}
\newcommand{\vNsCJPiNear}{11}
\newcommand{\vNsCJPiRuns}{--}
\newcommand{\vNsCJPiSApp}{1.0}
\newcommand{\vNsCJPiSSea}{3.0}
\newcommand{\vNsCJPiSTra}{0.0}
\newcommand{\vNsCJPiSSet}{0.0}
\newcommand{\vNsCJPiSReach}{4.0}
\newcommand{\vNsCJPiSNear}{8.3}
\newcommand{\vNsCJPiSRuns}{5, 3, 4}
\newcommand{\vNtQJGpu}{20}
\newcommand{\vNtQJPiS}{4.7}
\newcommand{\vNtQJAppPiS}{1.3}
\newcommand{\vNtQJSeaPiS}{1.3}
\newcommand{\vNtQJTraPiS}{1.0}
\newcommand{\vNtQJSetPiS}{1.0}
\newcommand{\vNtCJPiS}{4.0}
\newcommand{\vNtCJAppPiS}{1.0}
\newcommand{\vNtCJSeaPiS}{3.0}
\newcommand{\vNtCJTraPiS}{0.0}
\newcommand{\vNtCJSetPiS}{0.0}
\newcommand{\vWcOursMLo}{92.3}
\newcommand{\vWcOthMLo}{84.6}
\newcommand{\vAsPtsPtMean}{69.4}
\newcommand{\vAsPtsPtSd}{0.59}
\newcommand{\vAsPtsWrongMean}{20.1}
\newcommand{\vAsPtsWrongSd}{0.15}
\newcommand{\vAsPtsAuMean}{0.74}
\newcommand{\vAsPtsAuSd}{0.005}
\newcommand{\vAsParPtMean}{70.1}
\newcommand{\vAsParPtSd}{0.28}
\newcommand{\vAsParWrongMean}{18.6}
\newcommand{\vAsParWrongSd}{0.26}
\newcommand{\vAsParAuMean}{0.76}
\newcommand{\vAsParAuSd}{0.005}
\newcommand{\vAsSetPtMean}{72.5}
\newcommand{\vAsSetPtSd}{0.20}
\newcommand{\vAsSetNWord}{three}
\newcommand{\vAsSetWrongMean}{17.7}
\newcommand{\vAsSetWrongSd}{1.08}
\newcommand{\vAsSetAuMean}{0.73}
\newcommand{\vAsSetAuSd}{0.017}
\newcommand{\vAsSetBPtMean}{73.2}
\newcommand{\vAsSetBPtSd}{0.41}
\newcommand{\vAsSetBWrongMean}{17.2}
\newcommand{\vAsSetBWrongSd}{0.59}
\newcommand{\vAsSetBAuMean}{0.73}
\newcommand{\vAsSetBAuSd}{0.017}
\newcommand{\vAsSetQPtMean}{72.7}
\newcommand{\vAsSetQPtSd}{0.28}
\newcommand{\vAsSetQWrongMean}{15.2}
\newcommand{\vAsSetQWrongSd}{0.44}
\newcommand{\vAsSetQAuMean}{0.79}
\newcommand{\vAsSetQAuSd}{0.005}
\newcommand{\vAsOursPtMean}{73.4}
\newcommand{\vAsOursPtSd}{0.30}
\newcommand{\vAsOursWrongMean}{14.1}
\newcommand{\vAsOursWrongSd}{0.37}
\newcommand{\vAsOursAuMean}{0.81}
\newcommand{\vAsOursAuSd}{0.005}
\newcommand{\vSmSPts}{62.8}
\newcommand{\vSmSOurs}{67.2}
\newcommand{\vSmSPtsWr}{25.1}
\newcommand{\vSmSOursWr}{19.0}
\newcommand{\vSmSPtsAu}{0.76}
\newcommand{\vSmSOursAu}{0.82}
\newcommand{\vSmMPts}{69.4}
\newcommand{\vSmMOurs}{73.4}
\newcommand{\vSmMPtsWr}{20.1}
\newcommand{\vSmMOursWr}{14.1}
\newcommand{\vSmMPtsAu}{0.74}
\newcommand{\vSmMOursAu}{0.81}
\newcommand{\vSmLPts}{76.0}
\newcommand{\vSmLOurs}{79.2}
\newcommand{\vSmLPtsWr}{14.6}
\newcommand{\vSmLOursWr}{9.1}
\newcommand{\vSmLPtsAu}{0.75}
\newcommand{\vSmLOursAu}{0.82}
\newcommand{\vHzDigitsCell}{83.8}
\newcommand{\vHzDigitsHund}{75.6}
\newcommand{\vHzDigitsPt}{72.3}
\newcommand{\vHzTenCell}{83.1}
\newcommand{\vHzTenD}{\tm{}0.8}
\newcommand{\vHzHundHund}{74.2}
\newcommand{\vHzHundD}{\tm{}1.4}
\newcommand{\vCgDnTwo}{80.2}
\newcommand{\vCgDnThree}{72.5}
\newcommand{\vCgDnTen}{37.4}
\newcommand{\vCgPpTwo}{86.5}
\newcommand{\vCgPpThree}{83.5}
\newcommand{\vCgPpTen}{40.6}
\newcommand{\vCgOursTwo}{93.0}
\newcommand{\vCgOursThree}{90.7}
\newcommand{\vCgOursTen}{83.9}
\newcommand{\vCgChanceTwo}{45.8}
\newcommand{\vCgChanceThree}{28.8}
\newcommand{\vCgChanceTen}{10.9}
\newcommand{\vCgConstTwo}{49.0}
\newcommand{\vCgConstThree}{54.9}
\newcommand{\vCgConstTen}{23.4}
\newcommand{\vNdOursPt}{76.4}
\newcommand{\vNdWrPt}{73.5}
\newcommand{\vNdDfPt}{+2.9}
\newcommand{\vNdDfPtLo}{+2.0}
\newcommand{\vNdDfPtHi}{+3.9}
\newcommand{\vNdOursWrong}{10.9}
\newcommand{\vNdWrWrong}{17.5}
\newcommand{\vNdOursAuc}{0.83}
\newcommand{\vNdWrAuc}{0.73}
\newcommand{\vNdOursPeakWrong}{18.6}
\newcommand{\vNdOursPeakAuc}{0.65}
\newcommand{\vOpTokShared}{567}
\newcommand{\vOpTokOwn}{1{,}240}
\newcommand{\vOpOwlAcc}{83.5}
\newcommand{\vOpOwlWrong}{8.5}
\newcommand{\vOpOwlSafe}{74.9}
\newcommand{\vOpOwlGain}{3.6}
\newcommand{\vOpOursAcc}{79.9}
\newcommand{\vOpOursWrong}{5.3}
\newcommand{\vOpOursSafe}{84.1}
\newcommand{\vOpOursGain}{1.1}
\newcommand{\vOpOwlOverOurs}{3.6}
\newcommand{\vOpPtsAcc}{78.5}
\newcommand{\vOpPtsWrong}{9.9}
\newcommand{\vOpPtsSafe}{71.1}
\newcommand{\vGoAcc}{79.9}
\newcommand{\vGoWrong}{10.9}
\newcommand{\vGoSafe}{68.7}
\newcommand{\vGoMw}{76.4}
\newcommand{\vGoOverOurs}{1.1}
\newcommand{\vLrQJGpuMean}{20.0}
\newcommand{\vLrQJGpuSd}{2.0}
\newcommand{\vLrQJGpuRuns}{20, 22, 18}
\newcommand{\vLrQJAppGpu}{5.0}
\newcommand{\vLrQJSeaGpu}{5.0}
\newcommand{\vLrQJTraGpu}{5.0}
\newcommand{\vLrQJSetGpu}{5.0}
\newcommand{\vLrQJSecGpu}{0.20}
\newcommand{\vLrQDGpuMean}{19.0}
\newcommand{\vLrQDGpuSd}{1.0}
\newcommand{\vLrQDGpuRuns}{20, 19, 18}
\newcommand{\vLrQDAppGpu}{5.7}
\newcommand{\vLrQDSeaGpu}{5.0}
\newcommand{\vLrQDTraGpu}{5.0}
\newcommand{\vLrQDSetGpu}{3.3}
\newcommand{\vLrQDSecGpu}{0.04}
\newcommand{\vLrCJGpuMean}{30.0}
\newcommand{\vLrCJGpuSd}{0.0}
\newcommand{\vLrCJGpuRuns}{30, 30, 30}
\newcommand{\vLrCJAppGpu}{6.3}
\newcommand{\vLrCJSeaGpu}{6.7}
\newcommand{\vLrCJTraGpu}{8.3}
\newcommand{\vLrCJSetGpu}{8.7}
\newcommand{\vLrCJSecGpu}{0.16}
\newcommand{\vLrCDGpuMean}{28.7}
\newcommand{\vLrCDGpuSd}{0.6}
\newcommand{\vLrCDGpuRuns}{29, 28, 29}
\newcommand{\vLrCDAppGpu}{7.3}
\newcommand{\vLrCDSeaGpu}{7.0}
\newcommand{\vLrCDTraGpu}{9.7}
\newcommand{\vLrCDSetGpu}{4.7}
\newcommand{\vLrCDSecGpu}{0.07}
\newcommand{\vDvAMCell}{32.6}
\newcommand{\vDvAMHund}{33.3}
\newcommand{\vDvAMFull}{45.6}
\newcommand{\vDvAMTpl}{54.1}
\newcommand{\vDvAMFree}{66.6}
\newcommand{\vDvASCell}{25.0}
\newcommand{\vDvASHund}{25.8}
\newcommand{\vDvASFull}{36.1}
\newcommand{\vDvASTpl}{40.1}
\newcommand{\vDvASFree}{58.6}
\newcommand{\vDvLsMCell}{26.4}
\newcommand{\vDvLsMHund}{26.9}
\newcommand{\vDvLsMFull}{45.5}
\newcommand{\vDvLsMTpl}{53.0}
\newcommand{\vDvLsMFree}{66.3}
\newcommand{\vDvLsSCell}{17.2}
\newcommand{\vDvLsSHund}{17.9}
\newcommand{\vDvLsSFull}{34.6}
\newcommand{\vDvLsSTpl}{32.5}
\newcommand{\vDvLsSFree}{48.0}
\newcommand{\vDvTplOverHundMin}{1.6}
\newcommand{\vDvTplOverHundMax}{2.0}
\newcommand{\vDvTplOverFullMin}{0.94}
\newcommand{\vDvTplOverFullMax}{1.2}
\newcommand{\vDvFreeOverFullMin}{1.4}
\newcommand{\vDvFreeOverFullMax}{1.6}
\newcommand{\vLHrexGen}{424}
\newcommand{\vLHrexFull}{249}
\newcommand{\vLHrexTok}{15.7}
\newcommand{\vLHrexGenFmt}{338}
\newcommand{\vLHrexRatioFFmt}{1.4}
\newcommand{\vLHrexGenMin}{168}
\newcommand{\vRxIvlAWrite}{42.6}
\newcommand{\vRxIvlARead}{40.1}
\newcommand{\vRxIvlAAu}{0.73}
\newcommand{\vRxIvlWrite}{41.6}
\newcommand{\vRxIvlRead}{37.0}
\newcommand{\vRxIvlAu}{0.73}
\newcommand{\vRxIvlDWrite}{52.1}
\newcommand{\vRxIvlDRead}{46.3}
\newcommand{\vRxIvlDAu}{0.76}
\newcommand{\vRxGemWrite}{62.9}
\newcommand{\vRxGemRead}{62.5}
\newcommand{\vRxGemAu}{0.71}
\newcommand{\vRxLfmWrite}{73.9}
\newcommand{\vRxLfmRead}{73.9}
\newcommand{\vRxLfmAu}{0.69}
\newcommand{\vRxLfvWrite}{67.6}
\newcommand{\vRxLfvRead}{67.6}
\newcommand{\vRxLfvAu}{0.73}
\newcommand{\vRxBoxGapMin}{2.6}
\newcommand{\vRxBoxGapMax}{5.8}
\newcommand{\vRxIvlAPlain}{18.6}
\newcommand{\vRxIvlDPlain}{38.2}
\newcommand{\vNwGraphs}{109}
\newcommand{\vNwGraphm}{148}
\newcommand{\vNwGraphl}{246}
\newcommand{\vNwBestOverReadMin}{2.3}
\newcommand{\vNwBestOverReadMax}{2.8}
\newcommand{\vNwFreeOverMapMin}{1.5}
\newcommand{\vNwFreeOverMapMax}{1.6}
\newcommand{\vGcPtsSCell}{75.0}
\newcommand{\vGcPtsSHund}{64.5}
\newcommand{\vGcPtsMCell}{79.9}
\newcommand{\vGcPtsMHund}{71.1}
\newcommand{\vGcPtsLCell}{83.1}
\newcommand{\vGcPtsLHund}{77.1}
\newcommand{\vGcQvlTwoPtsCell}{80.7}
\newcommand{\vGcQvlTwoPtsHund}{74.2}
\newcommand{\vGcQvlFourPtsCell}{84.9}
\newcommand{\vGcQvlFourPtsHund}{79.6}
\newcommand{\vGcQvlTwoCell}{83.2}
\newcommand{\vGcQvlTwoHund}{76.2}
\newcommand{\vGcQvlFourCell}{86.2}
\newcommand{\vGcQvlFourHund}{80.6}
\newcommand{\vGcNatMCell}{75.0}
\newcommand{\vGcNatMHund}{61.0}
\newcommand{\vGcNatLCell}{83.1}
\newcommand{\vGcNatLHund}{74.7}
\newcommand{\vGcGpiCell}{84.6}
\newcommand{\vGcGpiHund}{79.4}
\newcommand{\vQnTwoCellAcc}{83.4}
\newcommand{\vQnTwoHundAcc}{76.2}
\newcommand{\vQnTwoPtAcc}{73.8}
\newcommand{\vQnTwoWrong}{20.0}
\newcommand{\vQnTwoAu}{0.68}
\newcommand{\vQnTwoMs}{40}
\newcommand{\vQnFourCellAcc}{86.3}
\newcommand{\vQnFourHundAcc}{80.3}
\newcommand{\vQnFourPtAcc}{78.6}
\newcommand{\vQnFourWrong}{16.5}
\newcommand{\vQnFourAu}{0.65}
\newcommand{\vQnFourMs}{61}
\newcommand{\vHtHeadPt}{70.1}
\newcommand{\vHtDPt}{\tm{}2.2}
\newcommand{\vHtTopHundShareX}{13.4}
\newcommand{\vHtTopHundShareY}{14.8}
\newcommand{\vHtStrRareMinusCommonLo}{\tm{}3.7}
\newcommand{\vHtStrRareMinusCommonHi}{+0.3}
\newcommand{\vRcQvlTwoPPeakAu}{0.75}
\newcommand{\vRcQvlTwoPPaceAu}{0.79}
\newcommand{\vRcQvlTwoPEntAu}{0.78}
\newcommand{\vRcQvlTwoPWinAu}{0.81}
\newcommand{\vRcQvlTwoPRegionAu}{--}
\newcommand{\vRcQvlTwoOursPeakAu}{0.68}
\newcommand{\vRcQvlTwoOursPaceAu}{0.79}
\newcommand{\vRcQvlTwoOursEntAu}{0.76}
\newcommand{\vRcQvlTwoOursWinAu}{0.80}
\newcommand{\vRcQvlTwoOursRegionAu}{--}
\newcommand{\vRcQvlTwoOursYesAu}{0.82}
\newcommand{\vRcQvlTwoPBestW}{14.3}
\newcommand{\vRcQvlTwoOursYesW}{10.8}
\newcommand{\vRcQvlFourPPeakAu}{0.72}
\newcommand{\vRcQvlFourPPaceAu}{0.77}
\newcommand{\vRcQvlFourPEntAu}{0.74}
\newcommand{\vRcQvlFourPWinAu}{0.79}
\newcommand{\vRcQvlFourPRegionAu}{--}
\newcommand{\vRcQvlFourOursPeakAu}{0.63}
\newcommand{\vRcQvlFourOursPaceAu}{0.75}
\newcommand{\vRcQvlFourOursEntAu}{0.70}
\newcommand{\vRcQvlFourOursWinAu}{0.76}
\newcommand{\vRcQvlFourOursRegionAu}{--}
\newcommand{\vRcQvlFourOursYesAu}{0.81}
\newcommand{\vRcQvlFourPBestW}{11.4}
\newcommand{\vRcQvlFourOursYesW}{8.9}
\newcommand{\vRcRexPPeakAu}{0.73}
\newcommand{\vRcRexPPaceAu}{0.77}
\newcommand{\vRcRexPEntAu}{0.75}
\newcommand{\vRcRexPWinAu}{0.79}
\newcommand{\vRcRexPRegionAu}{--}
\newcommand{\vRcRexOursPeakAu}{0.65}
\newcommand{\vRcRexOursPaceAu}{0.78}
\newcommand{\vRcRexOursEntAu}{0.74}
\newcommand{\vRcRexOursWinAu}{0.79}
\newcommand{\vRcRexOursRegionAu}{--}
\newcommand{\vRcRexOursYesAu}{0.83}
\newcommand{\vRcRexPBestW}{14.7}
\newcommand{\vRcRexOursYesW}{10.9}
\newcommand{\vRcElRelBestMin}{18}
\newcommand{\vRcElRelBestMax}{32}
\newcommand{\vQlOnlyOurs}{87}
\newcommand{\vQlOnlyCentre}{21}
\newcommand{\vUtFastsThou}{42}
\newcommand{\vUtVls}{97}
\newcommand{\vUtAxsAcc}{45.7}
\newcommand{\vUtAxsTwo}{48.8}
\newcommand{\vUtAxsDiff}{\tm3.1}
\newcommand{\vUtAxsLo}{\tm3.9}
\newcommand{\vUtAxsHi}{\tm2.3}
\newcommand{\vUtFastmThou}{50}
\newcommand{\vUtVlm}{140}
\newcommand{\vUtVlmTok}{21.2}
\newcommand{\vUtAxmAcc}{52.9}
\newcommand{\vUtAxmTwo}{57.9}
\newcommand{\vUtAxmDiff}{\tm5.0}
\newcommand{\vUtAxmLo}{\tm5.8}
\newcommand{\vUtAxmHi}{\tm4.1}
\newcommand{\vUtFastlThou}{88}
\newcommand{\vUtFastlGrDp}{0.0434}
\newcommand{\vUtVll}{213}
\newcommand{\vUtVllTok}{19.0}
\newcommand{\vUtAxlAcc}{53.2}
\newcommand{\vUtAxlTwo}{71.9}
\newcommand{\vUtAxlDiff}{\tm18.7}
\newcommand{\vUtAxlLo}{\tm19.8}
\newcommand{\vUtAxlHi}{\tm17.5}
\newcommand{\vRhiGdThouAcc}{68.5}
\newcommand{\vRhiGdThouWr}{25.5}
\newcommand{\vRhiQvsLoThouAcc}{74.2}
\newcommand{\vRhiQvsLoThouWr}{20.0}
\newcommand{\vRhiQvlThouAcc}{85.5}
\newcommand{\vRhiQvlThouWr}{10.8}
\newcommand{\vRhiOursMThouAcc}{82.8}
\newcommand{\vRhiOursMThouWr}{7.7}
\newcommand{\vRhiOursMLoThouAcc}{73.3}
\newcommand{\vRhiOursMLoThouWr}{14.2}
\newcommand{\vRhiOursLThouAcc}{85.5}
\newcommand{\vRhiOursLThouWr}{6.1}
\newcommand{\vLHivlAGenNew}{333}
\newcommand{\vOsHivlACellNew}{95.6}
\newcommand{\vOsHivlAThouNew}{95.0}
\newcommand{\vOsHivlARatioThouNew}{3.5}
\newcommand{\vLHivlGenNew}{348}
\newcommand{\vOsHivlCellNew}{94.6}
\newcommand{\vOsHivlThouNew}{95.5}
\newcommand{\vOsHivlRatioThouNew}{3.6}
\newcommand{\vLHivlDGenNew}{488}
\newcommand{\vOsHivlDCellNew}{114}
\newcommand{\vOsHivlDThouNew}{115}
\newcommand{\vOsHivlDRatioThouNew}{4.3}
\newcommand{\vLHgemGenNew}{937}
\newcommand{\vLHgemFullNew}{489}
\newcommand{\vLHgemCellNew}{189}
\newcommand{\vLHgemRatioFNew}{1.9}
\newcommand{\vLHlfmGenNew}{292}
\newcommand{\vLHlfmFullNew}{130}
\newcommand{\vLHlfmCellNew}{130}
\newcommand{\vLHlfmRatioFNew}{2.2}
\newcommand{\vLHlfvGenNew}{359}
\newcommand{\vLHlfvFullNew}{133}
\newcommand{\vLHlfvCellNew}{133}
\newcommand{\vLHlfvRatioFNew}{2.7}
\renewcommand{\vLHivlAGen}{\vLHivlAGenNew}
\renewcommand{\vOsHivlACell}{\vOsHivlACellNew}
\renewcommand{\vOsHivlAThou}{\vOsHivlAThouNew}
\renewcommand{\vOsHivlARatioThou}{\vOsHivlARatioThouNew}
\renewcommand{\vLHivlGen}{\vLHivlGenNew}
\renewcommand{\vOsHivlCell}{\vOsHivlCellNew}
\renewcommand{\vOsHivlThou}{\vOsHivlThouNew}
\renewcommand{\vOsHivlRatioThou}{\vOsHivlRatioThouNew}
\renewcommand{\vLHivlDGen}{\vLHivlDGenNew}
\renewcommand{\vOsHivlDCell}{\vOsHivlDCellNew}
\renewcommand{\vOsHivlDThou}{\vOsHivlDThouNew}
\renewcommand{\vOsHivlDRatioThou}{\vOsHivlDRatioThouNew}
\renewcommand{\vLHgemGen}{\vLHgemGenNew}
\renewcommand{\vLHgemFull}{\vLHgemFullNew}
\renewcommand{\vLHgemCell}{\vLHgemCellNew}
\renewcommand{\vLHgemRatioF}{\vLHgemRatioFNew}
\renewcommand{\vLHlfmGen}{\vLHlfmGenNew}
\renewcommand{\vLHlfmFull}{\vLHlfmFullNew}
\renewcommand{\vLHlfmCell}{\vLHlfmCellNew}
\renewcommand{\vLHlfmRatioF}{\vLHlfmRatioFNew}
\renewcommand{\vLHlfvGen}{\vLHlfvGenNew}
\renewcommand{\vLHlfvFull}{\vLHlfvFullNew}
\renewcommand{\vLHlfvCell}{\vLHlfvCellNew}
\renewcommand{\vLHlfvRatioF}{\vLHlfvRatioFNew}
\newcommand{\vAfArithDp}{$4{\times}10^{-8}$}
\newcommand{\vAfFpImages}{seven}
\newcommand{\vAfBfMaxDp}{0.011}
\newcommand{\vAfBfAxes}{410}
\newcommand{\vAfBfImages}{205}
\newcommand{\vAfBfCellSame}{401}
\newcommand{\vAfBfValueSame}{245}
\newcommand{\vAfCropMinSide}{40}
\newcommand{\vAfPointRecords}{288,000}
\newcommand{\vAfStepExamples}{96}
\newcommand{\vAfTrainSteps}{6,000}
\newcommand{\vAfRexWrPB}{63.5}
\newcommand{\vAfRexOursPB}{66.0}
\newcommand{\vAfRexWrSS}{70.8}
\newcommand{\vAfRexOursSS}{73.3}
\newcommand{\vAfRexWrWP}{65.0}
\newcommand{\vAfRexOursWP}{67.0}
\newcommand{\vAfRexWrRR}{75.7}
\newcommand{\vAfRexOursRR}{75.6}
\newcommand{\vAfRexWrRC}{83.3}
\newcommand{\vAfRexOursRC}{93.3}
\newcommand{\vAfQvlTwoWrPB}{62.7}
\newcommand{\vAfQvlTwoOursPB}{62.4}
\newcommand{\vAfQvlTwoWrSS}{66.8}
\newcommand{\vAfQvlTwoOursSS}{74.1}
\newcommand{\vAfQvlTwoWrWP}{58.0}
\newcommand{\vAfQvlTwoOursWP}{64.0}
\newcommand{\vAfQvlTwoWrRR}{76.5}
\newcommand{\vAfQvlTwoOursRR}{77.5}
\newcommand{\vAfQvlTwoWrRC}{84.6}
\newcommand{\vAfQvlTwoOursRC}{91.3}
\newcommand{\vAfQvlFourWrPB}{70.0}
\newcommand{\vAfQvlFourOursPB}{69.2}
\newcommand{\vAfQvlFourWrSS}{77.3}
\newcommand{\vAfQvlFourOursSS}{79.3}
\newcommand{\vAfQvlFourWrWP}{66.0}
\newcommand{\vAfQvlFourOursWP}{72.0}
\newcommand{\vAfQvlFourWrRR}{81.3}
\newcommand{\vAfQvlFourOursRR}{81.1}
\newcommand{\vAfQvlFourWrRC}{83.5}
\newcommand{\vAfQvlFourOursRC}{91.6}
\newcommand{\vAfQvlTwoMcWr}{51.8}
\newcommand{\vAfQvlTwoMcOurs}{51.4}
\newcommand{\vAfQvlFourMcWr}{57.6}
\newcommand{\vAfQvlFourMcOurs}{57.2}
\newcommand{\vAfRexMcWr}{46.3}
\newcommand{\vAfRexMcOurs}{46.9}
\newcommand{\vAfCtopWrong}{14.9}
\newcommand{\vEmJRPt}{\stale{71.8}}
\newcommand{\vEmJDPt}{\stale{72.1}}
\renewcommand{\vOssThou}{30.8}
\renewcommand{\vOsmThou}{40.3}
\renewcommand{\vOslThou}{63.3}
\renewcommand{\vSpStopsRatioThou}{2.1}
\renewcommand{\vSpStopmRatioThou}{2.1}
\renewcommand{\vSpStoplRatioThou}{1.9}
\renewcommand{\vNwFreeOverMapMin}{1.7}
\renewcommand{\vNwFreeOverMapMax}{1.9}
\renewcommand{\vDvASFull}{30.8}
\renewcommand{\vDvAMFull}{40.3}
\renewcommand{\vDvLsSFull}{21.9}
\renewcommand{\vDvLsMFull}{32.6}
\renewcommand{\vDvTplOverFullMin}{1.3}
\renewcommand{\vDvTplOverFullMax}{1.6}
\renewcommand{\vDvFreeOverFullMin}{1.7}
\renewcommand{\vDvFreeOverFullMax}{2.2}
\newcommand{\vFzAccS}{67.52}
\newcommand{\vFzAccM}{73.48}
\newcommand{\vFzAccL}{79.68}
\newcommand{\vFzRepS}{67.02}
\newcommand{\vFzRepM}{73.27}
\newcommand{\vFzRepL}{79.40}
\newcommand{\vFzMaxShift}{0.06}
\newcommand{\vDvLfSCell}{34.0}
\newcommand{\vDvLfSHund}{35.7}
\newcommand{\vDvLfSFull}{46.3}
\newcommand{\vDvLfSTpl}{63.8}
\newcommand{\vDvLfSFree}{95.4}
\newcommand{\vDvLfMCell}{60.8}
\newcommand{\vDvLfMHund}{61.8}
\newcommand{\vDvLfMFull}{79.1}
\newcommand{\vDvLfMTpl}{124.5}
\newcommand{\vDvLfMFree}{155.6}
\newcommand{\vPfHundWarmRef}{1.73}
\newcommand{\vPfHundWarmNew}{1.40}
\newcommand{\vPfThouWarmRef}{8.49}
\newcommand{\vPfThouWarmNew}{5.09}
\renewcommand{\vUtFastsThou}{41}
\renewcommand{\vUtFastmThou}{49}
\renewcommand{\vUtFastlThou}{84}
\renewcommand{\vNwBestOverReadMin}{2.4}
\renewcommand{\vNwBestOverReadMax}{2.8}
\renewcommand{\vOsxThou}{85.3}
\renewcommand{\vOsxVlRatioThou}{2.1}
\renewcommand{\vOssHiThou}{52.5}
\renewcommand{\vOsmHiThou}{86.8}
\renewcommand{\vOslHiThou}{132.5}
\renewcommand{\vOsxHiThou}{193.4}
\renewcommand{\vOssHiVlRatioThou}{2.3}
\renewcommand{\vOsmHiVlRatioThou}{1.6}
\renewcommand{\vOslHiVlRatioThou}{1.4}
\renewcommand{\vOsxHiVlRatioThou}{1.5}
\newcommand{\vFzHiMaxShift}{0.04}
\renewcommand{\vPqClickS}{4.0}
\renewcommand{\vPqClickWarmS}{1.09}
\renewcommand{\vPqHundS}{4.3}
\renewcommand{\vPqThouS}{8.1}
\renewcommand{\vPfHundWarmRef}{1.73}
\renewcommand{\vPfHundWarmNew}{1.40}
\renewcommand{\vPfThouWarmRef}{8.50}
\renewcommand{\vPfThouWarmNew}{5.13}
\renewcommand{\vPqClickS}{3.8}
\renewcommand{\vPqClickWarmS}{1.07}
\renewcommand{\vPqHundS}{4.1}
\renewcommand{\vPqThouS}{7.2}
\renewcommand{\vPfHundWarmRef}{1.73}
\renewcommand{\vPfHundWarmNew}{1.38}
\renewcommand{\vPfThouWarmRef}{8.48}
\renewcommand{\vPfThouWarmNew}{4.51}
\newcommand{\vPfPeakGb}{2.9}
\renewcommand{\vElGainMax}{4.4}
\renewcommand{\vSvivlDOurs}{12.4}
\renewcommand{\vRwAbsMax}{1.1}
\renewcommand{\vUtFastsThou}{40.6}
\renewcommand{\vUtFastmThou}{49.2}
\renewcommand{\vUtFastlThou}{84.0}
\renewcommand{\vPiSpreadMax}{2.4}
\renewcommand{\vSbmSSLo}{26.7}
\renewcommand{\vSblSSLo}{56.1}
\renewcommand{\vSivlWP}{51.2}
\renewcommand{\vOpTokShared}{566}
\newcommand{\vTrHoursS}{3.4}
\newcommand{\vTrHoursM}{6.4}
\newcommand{\vTrHoursL}{11.3}
\newcommand{\vTrGpus}{16}
\newcommand{\vOtGuidEx}{7{,}000}
\newcommand{\vOtGuidMin}{40}
\newcommand{\vOtInternEx}{17{,}880}
\newcommand{\vOtMjevQa}{12.3M}
\newcommand{\vTrExamples}{576{,}000}
\newcommand{\vOtMolmoSteps}{22{,}000}
\newcommand{\vOtMolmoBatch}{160}
\newcommand{\vVsDnCell}{31.5}
\newcommand{\vVsDnHund}{54.5}
\newcommand{\vVsDnThou}{77.9}
\newcommand{\vVsDnRound}{23}
\newcommand{\vVsDnOverOurs}{1.9}
\newcommand{\vVsDnReqMs}{46.2}
\newcommand{\vVsDnImgMs}{29.2}
\newcommand{\vVsPpCell}{51.3}
\newcommand{\vVsPpHund}{94.2}
\newcommand{\vVsPpThou}{137.7}
\newcommand{\vVsPpRound}{43}
\newcommand{\vVsPpOverOurs}{2.2}
\newcommand{\vVsDlCell}{22.4}
\newcommand{\vVsDlHund}{39.1}
\newcommand{\vVsDlThou}{56.1}
\newcommand{\vVsIntSCell}{66.3}
\newcommand{\vVsIntSHund}{127.6}
\newcommand{\vVsIntSThou}{190.0}
\newcommand{\vVsIntMCell}{78.2}
\newcommand{\vVsIntMHund}{148.1}
\newcommand{\vVsIntMThou}{219.2}
\newcommand{\vVsIntLCell}{136.6}
\newcommand{\vVsIntLHund}{265.0}
\newcommand{\vVsIntLThou}{395.4}
\newcommand{\vVsOjfCell}{22.7}
\newcommand{\vVsOjfHund}{39.6}
\newcommand{\vVsOjfThou}{56.9}
\newcommand{\vVsMjevThou}{252.0}
\newcommand{\vVsMjevAgree}{615/615}
\newcommand{\vVsAgree}{1230/1230}
\newcommand{\vOhMaxOpt}{255}
\newcommand{\vOhDnTenCell}{37.4}
\newcommand{\vOhDnTenHund}{17.2}
\newcommand{\vOhDnTenPt}{14.5}
\newcommand{\vOhDnTenMs}{125.5}
\newcommand{\vOhDnOneCell}{24.7}
\newcommand{\vOhDnOneHund}{10.8}
\newcommand{\vOhDnOnePt}{9.7}
\newcommand{\vOhDnOneMs}{105.5}
\newcommand{\vOhDnDCell}{-12.7}
\newcommand{\vOhDnDCellLo}{-14.1}
\newcommand{\vOhDnDCellHi}{-11.4}
\newcommand{\vOhPpTenCell}{40.6}
\newcommand{\vOhPpTenHund}{15.6}
\newcommand{\vOhPpTenPt}{12.6}
\newcommand{\vOhPpTenMs}{164.2}
\newcommand{\vOhPpOneCell}{46.2}
\newcommand{\vOhPpOneHund}{18.9}
\newcommand{\vOhPpOnePt}{16.8}
\newcommand{\vOhPpOneMs}{185.7}
\newcommand{\vOhPpDCell}{+5.6}
\newcommand{\vOhPpDCellLo}{+4.2}
\newcommand{\vOhPpDCellHi}{+7.0}
\newcommand{\vPbNPB}{982}
\newcommand{\vPbNSS}{1{,}272}
\newcommand{\vPbNWP}{100}
\newcommand{\vPbNRR}{2{,}000}
\newcommand{\vPbNRC}{1{,}000}
\newcommand{\vPbQwSPBWrWrong}{35.9}
\newcommand{\vPbQwSPBOursWrong}{35.9}
\newcommand{\vPbQwSSSWrWrong}{37.6}
\newcommand{\vPbQwSSSOursWrong}{28.4}
\newcommand{\vPbQwSWPWrWrong}{52.9}
\newcommand{\vPbQwSWPOursWrong}{42.9}
\newcommand{\vPbQwSRRWrWrong}{24.9}
\newcommand{\vPbQwSRROursWrong}{18.3}
\newcommand{\vPbQwSRCWrWrong}{12.0}
\newcommand{\vPbQwSRCOursWrong}{5.1}
\newcommand{\vPbQwMPBWrWrong}{32.1}
\newcommand{\vPbQwMPBOursWrong}{33.9}
\newcommand{\vPbQwMSSWrWrong}{28.3}
\newcommand{\vPbQwMSSOursWrong}{15.9}
\newcommand{\vPbQwMWPWrWrong}{50.0}
\newcommand{\vPbQwMWPOursWrong}{34.3}
\newcommand{\vPbQwMRRWrWrong}{18.1}
\newcommand{\vPbQwMRROursWrong}{11.5}
\newcommand{\vPbQwMRCWrWrong}{10.6}
\newcommand{\vPbQwMRCOursWrong}{4.6}
\newcommand{\vPbQvlTwoPBWrWrong}{30.8}
\newcommand{\vPbQvlTwoPBOursWrong}{28.5}
\newcommand{\vPbQvlTwoSSWrWrong}{16.9}
\newcommand{\vPbQvlTwoSSOursWrong}{9.5}
\newcommand{\vPbQvlTwoWPWrWrong}{41.4}
\newcommand{\vPbQvlTwoWPOursWrong}{37.1}
\newcommand{\vPbQvlTwoRRWrWrong}{14.9}
\newcommand{\vPbQvlTwoRROursWrong}{10.4}
\newcommand{\vPbQvlTwoRCWrWrong}{8.6}
\newcommand{\vPbQvlTwoRCOursWrong}{3.7}
\newcommand{\vPbIvlAPBWrWrong}{34.4}
\newcommand{\vPbIvlAPBOursWrong}{34.7}
\newcommand{\vPbIvlASSWrWrong}{47.4}
\newcommand{\vPbIvlASSOursWrong}{35.4}
\newcommand{\vPbIvlAWPWrWrong}{67.1}
\newcommand{\vPbIvlAWPOursWrong}{54.3}
\newcommand{\vPbIvlARRWrWrong}{27.3}
\newcommand{\vPbIvlARROursWrong}{23.6}
\newcommand{\vPbIvlARCWrWrong}{9.9}
\newcommand{\vPbIvlARCOursWrong}{4.3}
\newcommand{\vPbIvlBPBWrWrong}{28.5}
\newcommand{\vPbIvlBPBOursWrong}{28.3}
\newcommand{\vPbIvlBSSWrWrong}{21.1}
\newcommand{\vPbIvlBSSOursWrong}{13.2}
\newcommand{\vPbIvlBWPWrWrong}{55.7}
\newcommand{\vPbIvlBWPOursWrong}{37.1}
\newcommand{\vPbIvlBRRWrWrong}{22.9}
\newcommand{\vPbIvlBRROursWrong}{17.1}
\newcommand{\vPbIvlBRCWrWrong}{10.1}
\newcommand{\vPbIvlBRCOursWrong}{2.6}
\newcommand{\vPbLfvPBWrWrong}{29.5}
\newcommand{\vPbLfvPBOursWrong}{26.5}
\newcommand{\vPbLfvSSWrWrong}{36.9}
\newcommand{\vPbLfvSSOursWrong}{29.5}
\newcommand{\vPbLfvWPWrWrong}{27.1}
\newcommand{\vPbLfvWPOursWrong}{25.7}
\newcommand{\vPbLfvRRWrWrong}{11.6}
\newcommand{\vPbLfvRROursWrong}{10.1}
\newcommand{\vPbLfvRCWrWrong}{9.1}
\newcommand{\vPbLfvRCOursWrong}{2.4}
\newcommand{\vPbQwLPBWrWrong}{26.9}
\newcommand{\vPbQwLPBOursWrong}{26.7}
\newcommand{\vPbQwLSSWrWrong}{19.1}
\newcommand{\vPbQwLSSOursWrong}{9.5}
\newcommand{\vPbQwLWPWrWrong}{27.1}
\newcommand{\vPbQwLWPOursWrong}{25.7}
\newcommand{\vPbQwLRRWrWrong}{12.1}
\newcommand{\vPbQwLRROursWrong}{9.1}
\newcommand{\vPbQwLRCWrWrong}{8.7}
\newcommand{\vPbQwLRCOursWrong}{2.0}
\newcommand{\vPbQvlFourPBWrWrong}{24.7}
\newcommand{\vPbQvlFourPBOursWrong}{21.7}
\newcommand{\vPbQvlFourSSWrWrong}{10.4}
\newcommand{\vPbQvlFourSSOursWrong}{5.7}
\newcommand{\vPbQvlFourWPWrWrong}{34.3}
\newcommand{\vPbQvlFourWPOursWrong}{21.4}
\newcommand{\vPbQvlFourRRWrWrong}{11.6}
\newcommand{\vPbQvlFourRROursWrong}{7.4}
\newcommand{\vPbQvlFourRCWrWrong}{8.9}
\newcommand{\vPbQvlFourRCOursWrong}{2.6}
\newcommand{\vPbIvlDPBWrWrong}{27.2}
\newcommand{\vPbIvlDPBOursWrong}{24.1}
\newcommand{\vPbIvlDSSWrWrong}{14.9}
\newcommand{\vPbIvlDSSOursWrong}{9.2}
\newcommand{\vPbIvlDWPWrWrong}{41.4}
\newcommand{\vPbIvlDWPOursWrong}{25.7}
\newcommand{\vPbIvlDRRWrWrong}{20.6}
\newcommand{\vPbIvlDRROursWrong}{16.0}
\newcommand{\vPbIvlDRCWrWrong}{10.4}
\newcommand{\vPbIvlDRCOursWrong}{2.1}
\newcommand{\vPbGemPBWrWrong}{34.0}
\newcommand{\vPbGemPBOursWrong}{27.3}
\newcommand{\vPbGemSSWrWrong}{19.1}
\newcommand{\vPbGemSSOursWrong}{8.9}
\newcommand{\vPbGemWPWrWrong}{41.4}
\newcommand{\vPbGemWPOursWrong}{27.1}
\newcommand{\vPbGemRRWrWrong}{22.3}
\newcommand{\vPbGemRROursWrong}{14.2}
\newcommand{\vPbGemRCWrWrong}{12.7}
\newcommand{\vPbGemRCOursWrong}{4.1}
\newcommand{\vPbLfmPBWrWrong}{25.4}
\newcommand{\vPbLfmPBOursWrong}{23.4}
\newcommand{\vPbLfmSSWrWrong}{18.4}
\newcommand{\vPbLfmSSOursWrong}{12.0}
\newcommand{\vPbLfmWPWrWrong}{38.6}
\newcommand{\vPbLfmWPOursWrong}{34.3}
\newcommand{\vPbLfmRRWrWrong}{17.1}
\newcommand{\vPbLfmRROursWrong}{10.0}
\newcommand{\vPbLfmRCWrWrong}{10.6}
\newcommand{\vPbLfmRCOursWrong}{2.0}
\newcommand{\vPbRexPBWrWrong}{31.1}
\newcommand{\vPbRexPBOursWrong}{26.7}
\newcommand{\vPbRexSSWrWrong}{15.6}
\newcommand{\vPbRexSSOursWrong}{9.8}
\newcommand{\vPbRexWPWrWrong}{31.4}
\newcommand{\vPbRexWPOursWrong}{22.9}
\newcommand{\vPbRexRRWrWrong}{17.5}
\newcommand{\vPbRexRROursWrong}{12.0}
\newcommand{\vPbRexRCWrWrong}{9.7}
\newcommand{\vPbRexRCOursWrong}{1.3}
\newcommand{\vPbSumPBLower}{nine}
\newcommand{\vPbSumPBSig}{two}
\newcommand{\vPbSumPBHigher}{two}
\newcommand{\vPbSumPBTie}{one}
\newcommand{\vPbSumWPSig}{five}
\newcommand{\vCcQwSTenWrong}{26.4}
\newcommand{\vCcQwSTenAu}{0.75}
\newcommand{\vCcQwSHundWrong}{27.8}
\newcommand{\vCcQwSHundAu}{0.73}
\newcommand{\vCcQwSOursWrong}{19.2}
\newcommand{\vCcQwSOursAu}{0.82}
\newcommand{\vCcQwMTenWrong}{20.1}
\newcommand{\vCcQwMTenAu}{0.75}
\newcommand{\vCcQwMHundWrong}{22.4}
\newcommand{\vCcQwMHundAu}{0.71}
\newcommand{\vCcQwMOursWrong}{14.2}
\newcommand{\vCcQwMOursAu}{0.81}
\newcommand{\vCcQwLTenWrong}{14.7}
\newcommand{\vCcQwLTenAu}{0.75}
\newcommand{\vCcQwLHundWrong}{16.8}
\newcommand{\vCcQwLHundAu}{0.70}
\newcommand{\vCcQwLOursWrong}{9.3}
\newcommand{\vCcQwLOursAu}{0.82}
\newcommand{\vCcQvlTwoTenWrong}{15.6}
\newcommand{\vCcQvlTwoTenAu}{0.78}
\newcommand{\vCcQvlTwoHundWrong}{17.0}
\newcommand{\vCcQvlTwoHundAu}{0.75}
\newcommand{\vCcQvlTwoOursWrong}{10.8}
\newcommand{\vCcQvlTwoOursAu}{0.82}
\newcommand{\vCcQvlFourTenWrong}{12.3}
\newcommand{\vCcQvlFourTenAu}{0.76}
\newcommand{\vCcQvlFourHundWrong}{13.8}
\newcommand{\vCcQvlFourHundAu}{0.72}
\newcommand{\vCcQvlFourOursWrong}{8.9}
\newcommand{\vCcQvlFourOursAu}{0.81}
\newcommand{\vCcRexTenWrong}{16.2}
\newcommand{\vCcRexTenAu}{0.76}
\newcommand{\vCcRexHundWrong}{17.5}
\newcommand{\vCcRexHundAu}{0.73}
\newcommand{\vCcRexOursWrong}{10.9}
\newcommand{\vCcRexOursAu}{0.83}
\newcommand{\vCcGapMin}{1.4}
\newcommand{\vCcGapMax}{2.3}
\newcommand{\vDbDPiS}{9.0}
\newcommand{\vDbDPisAll}{3}
\newcommand{\vDbDPisSearch}{0.7}
\newcommand{\vDbDStDecl}{28}
\newcommand{\vDbDStDist}{11.8}
\newcommand{\vDbDStView}{26}
\newcommand{\vDbWPiS}{16.9}
\newcommand{\vDbWPisAll}{4}
\newcommand{\vDbWPisSearch}{3}
\newcommand{\vDbWStDecl}{136}
\newcommand{\vDbWStDist}{6.6}
\newcommand{\vDbWStView}{50}
\newcommand{\vDbVidOursOnly}{12}
\newcommand{\vDbVidBoth}{10}
\newcommand{\vDbVidWrOnly}{3}
\newcommand{\vPbPubRlAvg}{56.8}
\newcommand{\vPbPubRlSpatial}{63.1}
\newcommand{\vPbPubMoAvg}{67.7}
\newcommand{\vPbPubMoSpatial}{71.8}
\newcommand{\vPubWpRoboPoint}{46.8}
\newcommand{\vPubWpEmbR}{69.5}
\newcommand{\vPubWpRoboRefer}{71.0}
\newcommand{\vPubRrEmbR}{85.6}
\newcommand{\vPubMwSeeClick}{81.1}
\newcommand{\vPubMwPixAct}{64.7}
\newcommand{\vAfSpreadD}{\tm3.4}
\newcommand{\vAfSpreadLo}{\tm4.4}
\newcommand{\vAfSpreadHi}{\tm2.4}
\newcommand{\vAfNoPtCeD}{\tm2.7}
\newcommand{\vAfNoPtCeLo}{\tm3.6}
\newcommand{\vAfNoPtCeHi}{\tm1.8}
\newcommand{\vApPplxNatCell}{56.2}
\newcommand{\vApPplxNatCellH}{18.3}
\newcommand{\vApPplxNatPt}{16.2}
\newcommand{\vApPplxNatAuc}{0.35}
\newcommand{\vApPplxNatWr}{88.2}
\newcommand{\vApLunaNatCell}{55.5}
\newcommand{\vApLunaNatCellH}{28.8}
\newcommand{\vApLunaNatPt}{26.5}
\newcommand{\vApLunaNatAuc}{0.60}
\newcommand{\vApLunaNatWr}{69.5}
\newcommand{\vApLunaNatRefuse}{562}
\newcommand{\vApLunaNatMs}{661}
\newcommand{\vApLunaNatTwoMs}{1,384}
\newcommand{\vApClefNatCell}{69.2}
\newcommand{\vApClefNatCellH}{34.0}
\newcommand{\vApClefNatPt}{30.6}
\newcommand{\vApClefNatAuc}{0.50}
\newcommand{\vApClefNatWr}{68.4}
\newcommand{\vApClefNatMs}{2,093}
\newcommand{\vApClefNatTwoMs}{4,215}
\newcommand{\vApQwMClefNatDPt}{+42.7}
\newcommand{\vApQwMClefNatDPtLo}{+41.2}
\newcommand{\vApQwMClefNatDPtHi}{+44.3}
\newcommand{\vApQwMClefNatDWrLo}{\tm{}55.9}
\newcommand{\vApQwMClefNatDWrHi}{\tm{}52.4}
\newcommand{\vApPplxDepCell}{52.3}
\newcommand{\vApPplxDepCellH}{17.0}
\newcommand{\vApPplxDepPt}{14.9}
\newcommand{\vApPplxDepAuc}{0.39}
\newcommand{\vApPplxDepWr}{87.4}
\newcommand{\vApLunaDepCell}{51.0}
\newcommand{\vApLunaDepCellH}{25.7}
\newcommand{\vApLunaDepPt}{23.7}
\newcommand{\vApLunaDepAuc}{0.64}
\newcommand{\vApLunaDepWr}{71.7}
\newcommand{\vApLunaDepAucPP}{0.67}
\newcommand{\vApLunaDepRefuse}{883}
\newcommand{\vApClefDepCell}{66.1}
\newcommand{\vApClefDepCellH}{32.6}
\newcommand{\vApClefDepPt}{29.5}
\newcommand{\vApClefDepAuc}{0.54}
\newcommand{\vApClefDepWr}{67.8}
\newcommand{\vApQwMAuc}{0.81}
\newcommand{\vApQwMWr}{14.2}
\newcommand{\vApQwLAuc}{0.82}
\newcommand{\vApQwLWr}{9.3}
\newcommand{\vMcDiffLo}{\tm{}3.5}
\newcommand{\vMcDiffRows}{15}
\newcommand{\vMcDiffWithinN}{ten}
\newcommand{\vWbMwWrongDiffZeroEight}{$-$2.3}
\newcommand{\vWbMwAucDiffZeroEight}{$+$0.05}
\newcommand{\vWbMwWrongDiffTwo}{$-$4.0}
\newcommand{\vWbMwAucDiffTwo}{$+$0.06}
\newcommand{\vWbMwWrongDiffFour}{$-$3.6}
\newcommand{\vWbMwAucDiffFour}{$+$0.06}
\newcommand{\vPbNcOursM}{62.9}
\newcommand{\vPbNcRl}{57.3}
\newcommand{\vPbNcOursL}{67.5}
\newcommand{\vPbNcMo}{66.8}
\newcommand{\vVidGpuSec}{0.2}
\newcommand{\vVidGpuReach}{34}
\newcommand{\vVidPiSec}{4.6}
\newcommand{\vVidOursSec}{2.2}
\newcommand{\vVidOursReach}{39}
\newcommand{\vVidGpuClose}{1.0}
\newcommand{\vVidGpuDecForty}{66}
\newcommand{\vVidPiClose}{0.4}
\newcommand{\vVidPiDecForty}{6}
\newcommand{\vVidPiLostT}{47}
\newcommand{\vVidPiLostD}{16}
\newcommand{\vVidOursClose}{0.9}
\newcommand{\vVidOursDecForty}{13}
\providecommand{\vsDcPilotRows}{53,976}
\providecommand{\vsDcPilotImgs}{25,727}
\providecommand{\vsDcStates}{2,958}
\providecommand{\vsDcPhashMin}{10}
\providecommand{\vsDcCmds}{40}
\providecommand{\Ys}{\mathcal{Y}}
\providecommand{\xp}{x^{+}}

\providecommand{\ps}{p^{\star}}

\providecommand{\KL}{\mathrm{KL}}
\providecommand{\CE}{\mathrm{CE}}

\providecommand{\E}{\mathbb{E}}

\definecolor{linkblue}{rgb}{0.21,0.49,0.74}
\hypersetup{breaklinks,colorlinks,allcolors=linkblue}
\crefname{figure}{Fig.}{Figs.}\Crefname{figure}{Figure}{Figures}\crefname{table}{Tab.}{Tabs.}\Crefname{table}{Table}{Tables}
\crefname{section}{Sec.}{Secs.}\Crefname{section}{Section}{Sections}\crefname{appendix}{Appendix}{Appendices}\crefname{equation}{Eq.}{Eqs.}

\crefname{proposition}{Prop.}{Props.}
\Crefname{proposition}{Proposition}{Propositions}
\crefname{remark}{Remark}{Remarks}
\crefname{result}{Result}{Results}
\crefname{assumption}{Assumption}{Assumptions}
\crefname{theorem}{Theorem}{Theorems}
\crefname{lemma}{Lemma}{Lemmas}
\crefname{definition}{Definition}{Definitions}

\usepackage{placeins}   %
\newcommand{\stale}[1]{\textcolor{red}{#1}}

\DeclareRobustCommand{\tm}{\ifmmode-\else\textminus\fi}   %

\makeatletter\newcommand{\nosign}[1]{\expandafter\@gobble#1}\makeatother %

\renewcommand{\topfraction}{0.85}\renewcommand{\textfraction}{0.1}   %
\usepackage{capt-of}
\definecolor{oursbg}{RGB}{238,242,248}   %
\newcommand{\oursrow}{\rowcolor{oursbg}}
\providecommand{\vfNrecs}{1{,}300}  %
\usepackage{multirow}

\newcommand{\papertitle}{Small Vision-Language Models Are Fast Self-Checking Pointing Decision Models}  %
\title{\papertitle}  %

\newcommand{\codeurl}{https://github.com/naver-ai/ReadPoint}
\newcommand{\projecturl}{https://naver-ai.github.io/ReadPoint/}
\author{Geewook Kim}
\affiliation{NAVER Cloud AI\\[2pt]{\small\texttt{gwkim.rsrch@gmail.com}}}
\metadata[Code]{\href{\codeurl}{\texttt{github.com/naver-ai/ReadPoint}}}
\metadata[Project page]{\href{\projecturl}{\texttt{naver-ai.github.io/ReadPoint}}}
\abstract{Robots, drones and web agents act by pointing, yet vision-language models (VLMs) point by generating coordinate tokens one at a time: each point costs a chain of dependent forward passes, and its token probabilities are only a rough guide to whether it is right. A decision model instead scores every answer in one forward pass and acts on the best one, but today's decision models for grounding choose among coarse options or need a round of questions for each finer level. We show that a VLM that writes coordinates as digits is already an accurate pointing decision model. Each axis has only 111 digit prefixes, all known before anything is generated. Read together as short continuations of the same image and instruction, their next-digit probabilities give each axis a distribution over all its values, and so a probability map on a $1000{\times}1000$ grid or any coarser one, with no new head. Off the shelf, this read is as accurate as the generated point and, for Qwen3.5, \vNwBestOverReadMin{} to \vNwBestOverReadMax{} times faster. Fine-tuning then makes the read one pass, by asking each axis as its own question, and more accurate, by training each box as a set that contains the answer rather than as its centre. The same pass also reads a self-check, the model's yes or no on whether it can point precisely. Across twelve VLMs of up to 4B parameters, the resulting model is \vElGainMin{} to \vElGainMax\,pp more accurate than the same VLM fine-tuned to write its point, and makes fewer confident errors than that model ranked by its own probabilities. To test at on-device speed, we hold each decision to its measured Raspberry Pi~5 time in a drone simulator, where our 0.8B model reaches the most targets; on web tasks, its self-check leaves fewer confident wrong clicks.%
}
\input{layout}

\begin{document}
\makeatletter\@secpenalty=0\makeatother   %

\maketitle
\begin{center}
\includegraphics[width=\textwidth]{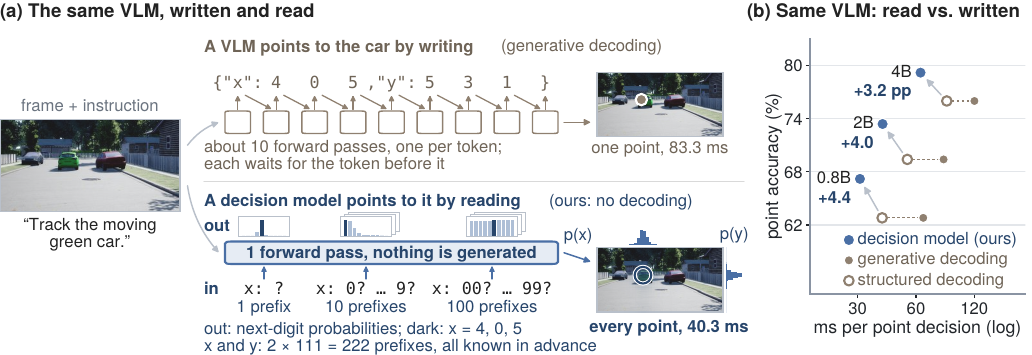}
\vspace{-1.2em}
\captionof{figure}{\textbf{A VLM writes its point token by token, and a decision model reads it in one forward pass.} (a) Qwen3.5-2B on a DroneCATS frame. Writing costs one forward pass per token, each waiting for the one before (\vAuStopM\,ms in vLLM, \vWfmTpl\,ms with structured decoding, which writes only the six digits). Reading generates nothing: the 222 digit prefixes of $x$ and $y$ are known in advance, so they all continue the same image and instruction in one forward pass, which gives every point its probability (\vOsmThou\,ms). (b) Point accuracy against time for Qwen3.5 at three sizes: our decision model (blue) against the same VLM fine-tuned to write its point, with the gain in pp (seed mean).}
\label{fig:teaser}
\end{center}

\section{Introduction}
\label{sec:intro}

\looseness=-1 A \emph{decision model}~\cite{almeida2026jev,interndecision,jevspatial,d3nano,pplxdecider,kananadecide} reads the probability of every possible answer from the model's forward pass and acts on the most probable one, instead of generating its answer token by token. It can be seen as multiple-choice answering: the probabilities of the option letters are read and the highest is taken, with no decoding and with a probability that says how sure the model is. Because nothing is decoded, the input is processed once and many questions about it are answered in parallel, which makes such models fast and cheap to serve, down to a small vision-language model (VLM) on an edge device. For agents and robots the decision that matters is often \emph{where} to act: a GUI agent clicks an element~\cite{screenspot}, a drone flies toward its target~\cite{dronecats}, and a robot grasps at a point~\cite{robopoint,molmo}.

\begin{figure*}[t]
\centering
\includegraphics[width=\textwidth]{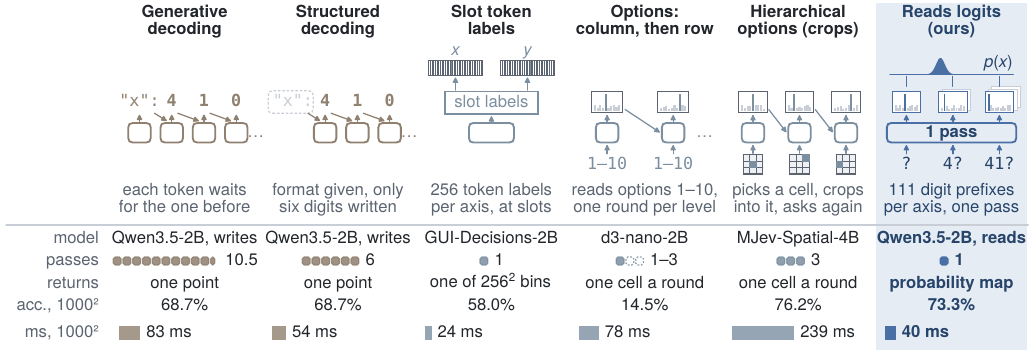}
\vspace{-1.4em}
\caption{\textbf{Six ways to decide where: the passes each needs and what it returns.} A box is a forward pass, with its input below and its output above, and an arrow into a box means that it waits for that output. Option models read the probabilities of ten numbered choices for a column and then a row, and a finer grid needs another round. Ours lists no options: its choices are the digits the VLM already writes, and all digit prefixes enter one pass. Numbers are those of \cref{tab:dec}, with Qwen3.5-2B fine-tuned, at the $1000{\times}1000$ grid (the point). Times are on one A100 GPU, with the written rows in vLLM and every decision model on our CUDA-graph engine with the image processed once; the released runtimes of the option models are slower (\cref{tab:optonce}).}
\label{fig:kinds}
\vspace{-0.6em}
\end{figure*}

\looseness=-1 We find that today's decision models ground poorly or slowly. Asked for the column and the row as two ten-option questions, models built to choose among options pick the right $10{\times}10$ cell on at most \vRzOptCellMax\% of images, and each finer grid costs another round over the image (\cref{tab:dec}, \cref{fig:kinds}). Pointing ones read 256 token labels per axis, which is right on \vRzGuidPtAcc\% of points~\cite{guidecisions}, or take three rounds of crops (\vVsMjevThouDec\,ms)~\cite{jevspatial}.

\looseness=-1 Our idea is simple: a VLM that points well is already a pointing decision model. It generates (writes) a coordinate digit by digit and gives at each digit a distribution over the next one (\cref{sec:readout}). To write $x{=}143$ it is asked for the first digit after an empty prefix, for the second after ``1'' and for the third after ``14''. Every prefix that can occur is known in advance, one empty, 10 of one digit and 100 of two digits per axis. They are thus many questions about one input, and we read them all in one forward pass under a tree mask, with no new head (\cref{fig:teaser}a). The digit probabilities multiply into a probability for every value of each axis, hence a map on any grid up to $1000{\times}1000$ and a probability $\mu(A)$ for any region $A$ (\cref{tab:families}). The read already works off the shelf: in the eight VLMs we read, the most probable point is as accurate as the generated one, within \vRwAbsMax{} percentage points (pp), and Qwen3.5 reads it \vNwBestOverReadMin{} to \vNwBestOverReadMax{} times faster than it generates it, in two passes because it writes $y$ after $x$ (\cref{tab:readwrite}, times in \cref{app:engine}). The mass of its most probable cell is also a confidence, which the generated point lacks.

\looseness=-1 Fine-tuning makes it a better one, and our second idea is how to label it. Standard fine-tuning teaches the model to write one point per label, the centre of a box, with a cross-entropy that counts every other point as wrong. Yet a point next to it on the object is just as right, and the centre of a box need not lie on the object at all. We treat each label as a set $A$ that contains the right answer, the whole box for a box, and maximize the probability $\mu(A)$ that the read puts on the set (\cref{sec:training}). The model is then free to place its mass on the object inside the box, and $\mu(A)$ estimates the probability that the target lies in $A$, which training on one point per label does not give (\cref{prop:sets}).

\looseness=-1 The same probability also tells the model when it is unsure. We add a yes/no self-check to the training, labelled yes when at least half of the model's own mass lies on the target, and it ranks the model's errors better than any score read from the map alone (\cref{sec:closer}). This matters on an edge device, where the cost is the pass over the image: the model can decide on a small image and look closer only where the self-check says no, and those are the images that a higher resolution helps.

\looseness=-1 We test this on twelve VLMs of up to 4B. Against the same VLM fine-tuned to write its point on the same data, the decision model is more accurate for all twelve, by \vElGainMin{} to \vElGainMax{} pp, and makes fewer confident errors, wrong points among its most confident 70\% of images (\cref{tab:decb}). For Qwen3.5 the full map is also \vSpStoplRatioThou--\vSpStopmRatioThou{} times faster than generating the point in vLLM (\cref{fig:teaser}b). Our 0.8B model also reaches the most targets in a drone simulator~\cite{dronecats} at a Raspberry Pi~5's speed, and keeps more web tasks free of confident wrong clicks (\cref{sec:closed}).

\looseness=-1 \textbf{Contributions.} (i) A finding: a VLM that writes coordinates is already a pointing decision model, and we read it out exactly with no training and no new head (\cref{sec:readout}). (ii) A training recipe: every label is a set that contains the right answer, and a self-check is labelled by the model's own mass. We show what this estimates (\cref{prop:sets}). It makes the read one pass, raises point accuracy for all twelve VLMs by \vElGainMin{} to \vElGainMax{} pp over fine-tuning the same VLM to write its point, and cuts confident errors (\cref{sec:training,sec:main,sec:closer}). (iii) Agent tasks: in a drone simulator at a Raspberry Pi~5's speed a 0.8B decision model reaches the most targets, and on web tasks it keeps more tasks free of confident errors (\cref{sec:closed}). We will open-source our code and models.

\begin{table*}[t]
\centering\footnotesize
\setlength{\tabcolsep}{4pt}
\providecommand{\relh}[2]{\begin{tabular}[b]{@{}c@{}}#1\\#2\end{tabular}}   %
\begin{tabular}{@{}ll@{\hspace{6pt}}c@{\hspace{7pt}}ccc@{\hspace{7pt}}ccc@{\hspace{7pt}}cc@{}}
\toprule
& & & \multicolumn{3}{c}{accuracy (\%)\,$\uparrow$} & \multicolumn{3}{c}{ms per decision\,$\downarrow$} & \multicolumn{2}{c@{}}{at the point} \\
\cmidrule(lr){4-6}\cmidrule(lr){7-9}\cmidrule(l){10-11}
Model & decides by & passes & $10^2$ & $100^2$ & $1000^2$ & $10^2$ & $100^2$ & $1000^2$ & \relh{wrong}{@70\%\,$\downarrow$} & AUROC\,$\uparrow$ \\
\midrule
\multicolumn{11}{@{}l}{\textit{Released models, 2B (base VLM in brackets)}} \\
\quad d3-nano~\cite{d3nano} (Qwen3.5-2B) & options, column then row & 1/2/3 & \vRzDnCellAcc & \vRzDnHundAcc & \vRzDnPtAcc & \vVsDnCell & \vVsDnHund & \vVsDnThou & \vRzDnPtFP & \vRzDnPtAu \\
\quad GUI-Decisions$^\ddagger$~\cite{guidecisions} (Qwen3-VL-2B) & 256 token labels & 1 & \vRzGuidCellAcc & \vRzGuidHundAcc & \vRzGuidPtAcc & \vFdGuidCell & \vFdGuidHund & \vFdGuidThou & \vRzGuidPtFP & \vRzGuidPtAu \\
\cmidrule[0.3pt](l{0pt}r{0pt}){1-11}
\multicolumn{11}{@{}l}{\textit{Qwen3.5-2B, one base VLM used five ways (our runs)}} \\
\quad off-the-shelf & generative decoding & \vUtVlmTok & \vGcNatMCell & \vGcNatMHund & \vRwQMWrite & \vUtVlm & \vUtVlm & \vUtVlm & \vRzNatMPtFP & \vRzNatMPtAu \\
\quad off-the-shelf, our read & reads logits & 2$^\dagger$ & \vRzNatMCellAcc & \vRzNatMHundAcc & \vRwQMRead & \vUtFastmThou & \vUtFastmThou & \vUtFastmThou & \vRzNatReadMPtFP & \vRzNatReadMPtAu \\
\quad fine-tuned to write its point & generative decoding & \vAuStopMTok & \vGcPtsMCell & \vGcPtsMHund & \vRzPtsMPtAcc & \vWsStopmTen & \vWsStopmHund & \vAuStopM & \vSvmPts & \vAumPts \\
\quad the same model$^\ast$ & structured decoding & 6 & \vGcPtsMCell & \vGcPtsMHund & \vRzPtsMPtAcc & \vWsTplmTen & \vWsTplmHund & \vWfmTpl & \vSvmPts & \vAumPts \\
\oursrow\quad\textbf{decision model (ours)} & \textbf{reads logits} & \textbf{1} & \textbf{\vRzOursMCellAcc} & \textbf{\vRzOursMHundAcc} & \textbf{\vRzOursMPtAcc} & \textbf{\vWfmReadC} & \textbf{\vOsmHund} & \textbf{\vOsmThou} & \textbf{\vSvmOurs} & \textbf{\vAumOurs} \\
\cmidrule[0.3pt](l{0pt}r{0pt}){1-11}
\multicolumn{11}{@{}l}{\textit{Qwen3-VL-2B, one base VLM used five ways (our runs)}} \\
\quad off-the-shelf~\cite{qwen3vl} & generative decoding & \vFtQvlTwoTok & \vGcQvlTwoCell & \vGcQvlTwoHund & \vRgQvlTwoAcc & \vFtQvlTwoVl & \vFtQvlTwoVl & \vFtQvlTwoVl & \vRgQvlTwoW & \vRgQvlTwoAu \\
\quad off-the-shelf, our read & reads logits & 2$^\dagger$ & \vQnTwoCellAcc & \vQnTwoHundAcc & \vQnTwoPtAcc & \vQnTwoMs & \vQnTwoMs & \vQnTwoMs & \vQnTwoWrong & \vQnTwoAu \\
\quad fine-tuned to write its point & generative decoding & \vUtQvlTwoPtsVlTok & \vGcQvlTwoPtsCell & \vGcQvlTwoPtsHund & \vRzQvlTwoPtsPtAcc & \vUtQvlTwoPtsVl & \vUtQvlTwoPtsVl & \vUtQvlTwoPtsVl & \vQvTwoPtsWrong & \vQvTwoPtsAu \\
\quad the same model$^\ast$ & structured decoding & 6 & \vGcQvlTwoPtsCell & \vGcQvlTwoPtsHund & \vRzQvlTwoPtsPtAcc & \vFdQvlTwoTplTen & \vFdQvlTwoTplHund & \vFdQvlTwoTplThou & \vQvTwoPtsWrong & \vQvTwoPtsAu \\
\oursrow\quad\textbf{decision model (ours)} & \textbf{reads logits} & \textbf{1} & \textbf{\vRzQvlTwoOursCellAcc} & \textbf{\vRzQvlTwoOursHundAcc} & \textbf{\vRzQvlTwoOursPtAcc} & \textbf{\vVgQvlTwoCell} & \textbf{\vVgQvlTwoHund} & \textbf{\vVgQvlTwoThou} & \textbf{\vQvTwoOursWrong} & \textbf{\vQvTwoOursAu} \\
\midrule\midrule
\multicolumn{11}{@{}l}{\textit{Released models, 4B (3.7B to 4.8B; base VLM in brackets)}} \\
\quad pplx-decider~\cite{pplxdecider} (Qwen3.5-4B) & options, column then row & 1/2/3 & \vRzPpCellAcc & \vRzPpHundAcc & \vRzPpPtAcc & \vVsPpCell & \vVsPpHund & \vVsPpThou & \vRzPpPtFP & \vRzPpPtAu \\
\quad MJev-Spatial~\cite{jevspatial} (Qwen3.5-4B) & options on crops & 3 & \vRzMjevCellAcc & \vRzMjevHundAcc & \vRzMjevPtAcc & \vVsMjevThouDec & \vVsMjevThouDec & \vVsMjevThouDec & \vRzMjevPtFP & \vRzMjevPtAu \\
\quad GroundingPI~\cite{yu2026groundingpi} (Qwen3-4B, MoonViT) & generative decoding & \vFtGpiTok & \vGcGpiCell & \vGcGpiHund & \vRgGpiAcc & \vFtGpiVl & \vFtGpiVl & \vFtGpiVl & \vRgGpiW & \vRgGpiAu \\
\cmidrule[0.3pt](l{0pt}r{0pt}){1-11}
\multicolumn{11}{@{}l}{\textit{Qwen3.5-4B, one base VLM used five ways (our runs)}} \\
\quad off-the-shelf & generative decoding & \vUtVllTok & \vGcNatLCell & \vGcNatLHund & \vRwQLWrite & \vUtVll & \vUtVll & \vUtVll & \vRzNatLPtFP & \vRzNatLPtAu \\
\quad off-the-shelf, our read & reads logits & 2$^\dagger$ & \vRzNatLCellAcc & \vRzNatLHundAcc & \vRwQLRead & \vUtFastlThou & \vUtFastlThou & \vUtFastlThou & \vRzNatReadLPtFP & \vRzNatReadLPtAu \\
\quad fine-tuned to write its point & generative decoding & \vAuStopLTok & \vGcPtsLCell & \vGcPtsLHund & \vRzPtsLPtAcc & \vWsStoplTen & \vWsStoplHund & \vAuStopL & \vSvlPts & \vAulPts \\
\quad the same model$^\ast$ & structured decoding & 6 & \vGcPtsLCell & \vGcPtsLHund & \vRzPtsLPtAcc & \vWsTpllTen & \vWsTpllHund & \vWflTpl & \vSvlPts & \vAulPts \\
\oursrow\quad\textbf{decision model (ours)} & \textbf{reads logits} & \textbf{1} & \textbf{\vRzOursLCellAcc} & \textbf{\vRzOursLHundAcc} & \textbf{\vRzOursLPtAcc} & \textbf{\vWflReadC} & \textbf{\vOslHund} & \textbf{\vOslThou} & \textbf{\vSvlOurs} & \textbf{\vAulOurs} \\
\cmidrule[0.3pt](l{0pt}r{0pt}){1-11}
\multicolumn{11}{@{}l}{\textit{Qwen3-VL-4B, one base VLM used five ways (our runs)}} \\
\quad off-the-shelf~\cite{qwen3vl} & generative decoding & \vFtQvlFourTok & \vGcQvlFourCell & \vGcQvlFourHund & \vRgQvlFourAcc & \vFtQvlFourVl & \vFtQvlFourVl & \vFtQvlFourVl & \vRgQvlFourW & \vRgQvlFourAu \\
\quad off-the-shelf, our read & reads logits & 2$^\dagger$ & \vQnFourCellAcc & \vQnFourHundAcc & \vQnFourPtAcc & \vQnFourMs & \vQnFourMs & \vQnFourMs & \vQnFourWrong & \vQnFourAu \\
\quad fine-tuned to write its point & generative decoding & \vUtQvlFourPtsVlTok & \vGcQvlFourPtsCell & \vGcQvlFourPtsHund & \vRzQvlFourPtsPtAcc & \vUtQvlFourPtsVl & \vUtQvlFourPtsVl & \vUtQvlFourPtsVl & \vQvFourPtsWrong & \vQvFourPtsAu \\
\quad the same model$^\ast$ & structured decoding & 6 & \vGcQvlFourPtsCell & \vGcQvlFourPtsHund & \vRzQvlFourPtsPtAcc & \vFdQvlFourTplTen & \vFdQvlFourTplHund & \vFdQvlFourTplThou & \vQvFourPtsWrong & \vQvFourPtsAu \\
\oursrow\quad\textbf{decision model (ours)} & \textbf{reads logits} & \textbf{1} & \textbf{\vRzQvlFourOursCellAcc} & \textbf{\vRzQvlFourOursHundAcc} & \textbf{\vRzQvlFourOursPtAcc} & \textbf{\vVgQvlFourCell} & \textbf{\vVgQvlFourHund} & \textbf{\vVgQvlFourThou} & \textbf{\vQvFourOursWrong} & \textbf{\vQvFourOursAu} \\
\bottomrule
\end{tabular}

\caption{\textbf{Ways to decide where, on five benchmarks (\vBcNAll{} test images).} Passes count the forward passes after the image and instruction, which every way processes once: one per token for generative decoding (the mean is shown), one per digit for structured decoding, one for slot token labels, one round per grid level for option models, and one for our read, or two off the shelf$^\dagger$. Times are on one A100 GPU, with generative decoding in vLLM and all other rows on our CUDA-graph engine (\cref{app:protocol,app:engine}), where the option models, like our rows, resize and encode each image once, and MJev-Spatial, which reads a new crop in every round, resizes each crop on the GPU. Their released runtimes are slower, as they encode the image again for every question: d3-nano takes \vFdDnCell, \vFdDnHund{} and \vFdDnThou\,ms, pplx-decider \vFdPpCell, \vFdPpHund{} and \vFdPpThou\,ms and MJev-Spatial \vFdMjevThouDec\,ms (\cref{tab:optonce}). On a $10{\times}10$, $100{\times}100$ or $1000{\times}1000$ grid (the point), a cell is right when it meets the target. Wrong@70\%, the wrong points among a model's most confident 70\% of images, and AUROC rank ours by its self-check, released models by their own confidence, and all other rows by the top cell mass of their map. Trained rows are one seed, with seed means in \cref{tab:decb}. $^\ast$The same fine-tuned model writes the same point, so only its time differs. The fine-tuned Qwen3.5 writer stops after the digits each grid needs; the other generating rows are timed for the whole point on every grid. $^\dagger$Read as $x$, then $y$ given $x$. $^\ddagger$Built for GUI agents.}
\label{tab:dec}\label{tab:released}\label{tab:where10}\label{tab:kanana}
\end{table*}

\begin{table}[t]
\centering\footnotesize
\setlength{\tabcolsep}{2.2pt}
\begin{tabular}{@{}l@{\hspace{2pt}}l@{\hspace{3pt}}c>{\columncolor{oursbg}}c@{\hspace{5pt}}c>{\columncolor{oursbg}}c@{\hspace{5pt}}c>{\columncolor{oursbg}}c@{\hspace{4pt}}c@{}}
\toprule
& & \multicolumn{2}{c}{point acc. $\uparrow$} & \multicolumn{2}{c}{wrong@70\% $\downarrow$} & \multicolumn{2}{c}{AUROC $\uparrow$} & \\
\cmidrule(lr){3-4}\cmidrule(lr){5-6}\cmidrule(lr){7-8}
Base VLM & format & writes & ours & writes & ours & writes & ours & faster \\
\midrule
Qwen3.5-0.8B$^\S$ & digits & \vSmSPts & \textbf{\vSmSOurs} & \vSmSPtsWr & \textbf{\vSmSOursWr} & \vSmSPtsAu & \textbf{\vSmSOursAu} & \vSpStopsRatioThou$\times$ \\
Qwen3.5-2B$^\S$ & digits & \vSmMPts & \textbf{\vSmMOurs} & \vSmMPtsWr & \textbf{\vSmMOursWr} & \vSmMPtsAu & \textbf{\vSmMOursAu} & \vSpStopmRatioThou$\times$ \\
Qwen3-VL-2B & digits & \vQvTwoPtsPt & \textbf{\vQvTwoOursPt} & \vQvTwoPtsWrong & \textbf{\vQvTwoOursWrong} & \vQvTwoPtsAu & \textbf{\vQvTwoOursAu} & \vVgQvlTwoFasterVl$\times$ \\
InternVL3.5-1B & digits & \vNivlASFTPt & \textbf{\vNivlAFinPt} & \vSvivlAPts & \textbf{\vSvivlAOurs} & \vAuivlAPts & \textbf{\vAuivlAOurs} & \vOsHivlARatioThou$\times$ \\
InternVL3.5-2B & digits & \vNivlSFTPt & \textbf{\vNivlFinPt} & \vSvivlPts & \textbf{\vSvivlOurs} & \vAuivlPts & \textbf{\vAuivlOurs} & \vOsHivlRatioThou$\times$ \\
LFM3-VL-2B & 1 tok. & \vNlfvSFTPt & \textbf{\vNlfvFinPt} & \vSvlfvPts & \textbf{\vSvlfvOurs} & \vAulfvPts & \textbf{\vAulfvOurs} & \vLHlfvRatioF$\times$ \\
\midrule
Qwen3.5-4B$^\S$ & digits & \vSmLPts & \textbf{\vSmLOurs} & \vSmLPtsWr & \textbf{\vSmLOursWr} & \vSmLPtsAu & \textbf{\vSmLOursAu} & \vSpStoplRatioThou$\times$ \\
Qwen3-VL-4B & digits & \vQvFourPtsPt & \textbf{\vQvFourOursPt} & \vQvFourPtsWrong & \textbf{\vQvFourOursWrong} & \vQvFourPtsAu & \textbf{\vQvFourOursAu} & \vVgQvlFourFasterVl$\times$ \\
InternVL3.5-4B & digits & \vNivlDSFTPt & \textbf{\vNivlDFinPt} & \vSvivlDPts & \textbf{\vSvivlDOurs} & \vAuivlDPts & \textbf{\vAuivlDOurs} & \vOsHivlDRatioThou$\times$ \\
Gemma-4-E4B$^\ddagger$ & digits & \vNgemSFTPt & \textbf{\vNgemFinPt} & \vSvgemPts & \textbf{\vSvgemOurs} & \vAugemPts & \textbf{\vAugemOurs} & \vLHgemRatioF$\times$ \\
LFM2.5-VL-3B & 1 tok. & \vNlfmSFTPt & \textbf{\vNlfmFinPt} & \vSvlfmPts & \textbf{\vSvlfmOurs} & \vAulfmPts & \textbf{\vAulfmOurs} & \vLHlfmRatioF$\times$ \\
Rex-Omni-3B$^\P$ & loc. & \vNdWrPt & \textbf{\vNdOursPt} & \vNdWrWrong & \textbf{\vNdOursWrong} & \vNdWrAuc & \textbf{\vNdOursAuc} & \vLHrexRatioFFmt$\times$ \\
\bottomrule
\end{tabular}

\caption{\textbf{The same VLM, data and steps, fine-tuned to write its point (writes) or as our decision model (ours).} Format is how the model writes a coordinate: three digits, one token per value (1 tok.) or one location token (loc.). Wrong@70\% is the share of wrong points among a model's most confident 70\% of images, by our self-check and by the point-writing model's top $10{\times}10$ cell mass ($100{\times}100$ for Qwen3-VL and Rex-Omni-3B, \cref{tab:confchoice}), and AUROC is of the same confidence. \emph{Faster} is the time to generate the point over the time of our full map, with engines in \cref{tab:speed}. Ours is in bold. $^\S$Means over seeds. $^\ddagger$Read with one pass per digit. $^\P$Timed for a complete answer of eight tokens (\cref{app:readout}).}
\label{tab:decb}
\end{table}
\section{Related Work}
\label{sec:related}

\begin{table*}[!t]
\centering\footnotesize
\setlength{\tabcolsep}{3pt}\renewcommand{\arraystretch}{0.9}
\newcommand{\ph}[2]{\begin{tabular}[b]{@{}c@{}}#1\\#2\end{tabular}}
\newcommand{\pl}[1]{\begin{tabular}[t]{@{}l@{}}#1\end{tabular}}
\begin{tabular}{@{}lllllcc@{}}
\toprule
Work & Venue & Decides by & Confidence from & Looks closer & \ph{probability}{map} & \ph{no added}{head or tokens} \\
\midrule
Intern-Decision~\cite{interndecision} & arXiv 2026 & options & the option's probability & -- & -- & $\sim$ \\
MJev-Spatial~\cite{jevspatial} & arXiv 2026 & options on crops, three rounds & option scores & always & $\sim$ & \checkmark \\
GUI-Decisions~\cite{guidecisions} & model card 2026 & 256 token labels per axis, at slots & its bin probabilities & -- & \checkmark & \checkmark \\
\addlinespace[3pt]
GUI-Actor~\cite{guiactor} & NeurIPS 2025 & attention head over image patches & a separate verifier & -- & \checkmark & -- \\
MolmoPoint~\cite{clark2026molmopoint} & ECCV 2026 & grounding tokens on patches & -- & -- & \checkmark & -- \\
UI-Zoomer~\cite{uizoomer} & arXiv 2026 & generative decoding & the spread of its points & if unsure & -- & \checkmark \\
\addlinespace[3pt]
SafeGround~\cite{safeground} & arXiv 2026 & generative decoding & sampled clicks, calibrated & -- & $\sim$ & \checkmark \\
PACE~\cite{pace} & EMNLP 2026 & generative decoding & the entropy of its digits & -- & -- & \checkmark \\
Argus~\cite{kumar2026argus} & NeurIPS 2026 & generative decoding & conformal regions & -- & -- & \checkmark \\
\midrule
\oursrow\textbf{Ours} & this paper & reads logits of its digits, one pass & a trained self-check & if unsure & \checkmark & \checkmark \\
\bottomrule
\end{tabular}

\caption{\textbf{Prior work on deciding where to act.} How each work decides where, where its confidence comes from, when it looks closer, and whether it returns a probability map (a probability for every point) and adds no head or tokens ($\checkmark$: yes, $\sim$: in part, --: none).}
\label{tab:position}
\end{table*}

\looseness=-1 \paragraph{Decision models.} VLMs write a point token by token, as location tokens~\cite{pix2seq,kosmos2} or text~\cite{shikra,qwen25vl,molmo}. Intern-Decision~\cite{interndecision} reads option symbols, MJev-Spatial~\cite{jevspatial} zooms through three rounds of $3{\times}3$ crops, and GUI-Decisions~\cite{guidecisions}, closest to ours, points by reading 256 single-token labels per axis at fixed slots of its answer (\cref{fig:kinds}). Our read needs neither crops nor bin labels learned in training: it gives the exact probability of every coordinate value in one pass, from the digits the VLM already writes (\cref{tab:position}). Hi-Token~\cite{hitoken} writes coordinates with digit tokens of its own, one per place, and decodes them. Other work makes the written answer faster: LocateAnything~\cite{locateanything} decodes a box in one step, Jag~\cite{zhu2026jag} predicts it with a new head, and LOCUS~\cite{talha2026locus} drafts it with such a head and checks it against the coordinate tokens in a few passes. Each returns one answer, while our read returns a probability for every point in one pass.

\looseness=-1 \paragraph{Training on partial labels.} A \emph{partial label} names a set that contains the right answer, and maximizing the set's probability is a consistent loss for it~\cite{cour2011partial,feng2020provably,wang2022pico}. GUI-Actor~\cite{guiactor} trains its attention head toward an even spread over the patches a box covers, and GUI-G\textsuperscript{2}~\cite{guig2} rewards clicks by a Gaussian around the element; both fix the distribution inside the box, which we leave to the model (\cref{app:recipe}). GUI-Decisions~\cite{guidecisions} trains toward soft targets around one bin, and PhysMoE~\cite{liu2026physmoe} supervises points with the signed distance field of a mask, which needs a mask for every label. We add what training on one written point converges to and where it must fail (\cref{prop:sets}), and apply the set loss to the coordinate read-out. Training toward the model's own guess~\cite{lv2020proden,rpopsd} did not help (\cref{app:selfteacher}).

\looseness=-1 \paragraph{Knowing when to act.} Grounding models attach a confidence to their point or box by sampling clicks~\cite{safeground}, asking the model~\cite{zhang2025hyperclick}, pooling digit entropies~\cite{pace}, reading attention~\cite{shalam2026propose} or drawing a calibrated disk~\cite{kumar2026argus}, for which conformal prediction~\cite{angelopoulos2023conformal} gives a coverage guarantee (\cref{app:conformal}). Acting only when confident is selective prediction~\cite{geifman2017selective,kadavath2022know}. GUI models zoom in when unsure, by rule~\cite{uizoomer} or learned zooming~\cite{innerzoom}. Our self-check is trained on the model's own map and says when not to act.

\section{A Decision Model from Coordinate Tokens}
\label{sec:regions}
\label{sec:readout}

We explain how a pointing decision is read from the coordinate tokens of a VLM. Given an image $I$ and a referring phrase $\ell$, VLMs write a coordinate as 143, 14.3, 0.143 or a location token, and \cref{tab:families} shows that the read applies to each. We use three zero-padded digits, thousandths of the image width or height, so $x{=}143$ is 14.3\% of the way across. Writing it takes three steps: each gives a distribution over the next digit and feeds the decoded digit back.

\looseness=-1 \paragraph{Reading every value in one pass.} The value 143 has the probability of 1 as the first digit, times that of 4 after ``1'', times that of 3 after ``14''. Every value has such a product. A writer sees only the distributions along the one value it decodes, but the prefixes a digit can follow are all known in advance: the empty one, 10 of one digit and 100 of two digits. All 111 prefixes of an axis, 222 for a point, follow the same image and instruction. The model processes this \emph{shared prefix} once, as it must before writing, and all digit prefixes then continue from it in one forward pass (\cref{fig:teaser}a), in which a tree attention mask~\cite{miao2024specinfer} lets each see only the shared prefix and its own digits. Nothing is decoded, and nothing is sampled as in~\cite{guirc,safeground}. The products are the exact probability of every value from 000 to 999 (\cref{app:readout}).\label{prop:exact}

\looseness=-1 \paragraph{From values to a decision.} Each axis $a$ is asked as its own short question, a \emph{branch} $s_a$ that continues the shared prefix and ends in \texttt{Answer:} (\cref{app:readout}). Its read gives
\begin{equation}
p_a(v)=\textstyle\prod_{k=1}^{3}q(v_k\mid I,\ell,s_a,v_{<k}),\qquad a\in\{x,y\},
\label{eq:mu}
\end{equation}
over the 1000 values $v=\overline{v_1v_2v_3}$ of the axis, with $q$ the next-token distribution. The \emph{map} is the product of the two axes, which treats them as independent (\cref{app:sets}): a point $(v,v')$ has probability $p_x(v)\,p_y(v')$, and a region $A$ the mass $\mu(A)$ of its points. A grid cell is a pair of digit prefixes $b\times b'$ with mass $\mu(b\times b')=p_x(b)\,p_y(b')$, the products of their digit probabilities: one digit each for the $10{\times}10$ grid, all three for the $1000{\times}1000$ points. The \emph{decision} is the most probable cell on the grid that the task needs. An off-the-shelf VLM writes $y$ after $x$ and loses accuracy when asked for $y$ alone (\cref{sec:main}), so we read it in two passes, $x$ and then $y$ given the $x$ it would write. Training on separate questions makes the read one pass.

\paragraph{Other questions and the self-check.} An agent's other questions are further branches of the same pass: a box takes four branches, one per edge, and a multiple-choice question is a branch whose option letters give its distribution, as in Intern-Decision~\cite{interndecision}. One more branch, the self-check, asks the model whether it can point to the target precisely at this image resolution (label in \cref{sec:training}). We read $P(\text{yes})=q(\texttt{Yes})/(q(\texttt{Yes})+q(\texttt{No}))$, and the agent acts when $P(\text{yes})$ reaches $\frac12$ (\cref{app:falsepos}). Otherwise it holds the action or reads the image again at 1\,MP (\cref{sec:conformal}).

\label{sec:selective}

\phantomsection\label{fig:sets_readout}
\section{Training with Label Sets and a Self-Check}
\label{sec:training}
\label{sec:lspd}
\label{sec:theory}

Off the shelf, the read takes two passes and its confidence is weak (\cref{sec:main}). Like other decision models~\cite{d3nano,pplxdecider,jevspatial,guidecisions}, ours is therefore fine-tuned from a base VLM. We train one LoRA adapter~\cite{hu2022lora} per model for \vSetTrainSteps{} steps of \vAfStepExamples{} examples, with the vision encoder frozen, on public pointing, box, GUI and multiple-choice annotations, after removing every test image (\cref{app:data,tab:hparams}).

\paragraph{Labels as sets.} Every point on the object is right for $(I,\ell)$, and a box contains them without saying which they are. We therefore treat each label as a set $A$ that contains the right answer, the \emph{label set}, and maximize the probability the model puts on it, as in partial-label learning~\cite{jin2002multiple,feng2020provably}:
\begin{equation}
\mathcal L_{\mathrm{lab}}(\theta)=-\,\mathbb E\,\log\mu_\theta(A\mid I,\ell),
\label{eq:lab}
\end{equation}
with $\mu_\theta$ the read-out of \cref{sec:regions}. For the point read, $A=A_x\times A_y$ is a rectangle and the loss is $-\log p_x(A_x)-\log p_y(A_y)$. The usual cross-entropy toward one point of $A$ equals \cref{eq:lab} plus a term that pulls the mass inside $A$ onto that point (\cref{lem:th:dec}). For a box, $A$ is the box and we drop the term, since its centre need not be right. For an annotated point, which is right, $A$ is a window of a tenth of each axis around it and we also keep the point's cross-entropy, which keeps the most probable value inside the window. Box edges and multiple-choice options are trained the same way (\cref{app:data}).

\paragraph{What each label estimates.} Let the right answer $Y$ be some point on the object, and let its label set $A$ contain it. We assume that the same set is drawn whichever point of the object is meant, a condition known as \emph{coarsening at random}~\cite{heitjan1991ignorability,gill1997car}. A box, drawn around the whole object, meets it approximately. A window moves with its annotated point, so it fails the condition and serves only as a tolerance.

\begin{proposition}[What each label estimates]\label{prop:sets}
Over all conditional distributions $\mu(\cdot\mid I,\ell)$, every minimizer of $\mathbb E[-\log\mu(A\mid I,\ell)]$ gives each label set the probability that the right answer lies in it, and the distribution of $Y$ is one of them. Training on one point $g(A)$ per label, such as the centre of a box, has the distribution of $g(A)$ as its unique minimizer, whose most probable answer can be one that is never right, whatever the capacity of the model.
\end{proposition}
In words, set labels teach the model how likely the target is to lie in each set, and where the most probable point falls inside a set is left to the point labels and to pretraining (\cref{sec:sets}). One chosen point per label teaches where that point lies, which need not be on the object. The set part is the consistency of partial-label learning~\cite{feng2020provably}, and the one-point part is what we add (proof in \cref{app:sets}).

\paragraph{Self-check label.} The self-check is labelled by the model itself: yes when its probability on the label set, $\mu_\theta(A\mid I,\ell)$, reaches one half, and no otherwise. Each training image is shown at a random pixel count between \vfResMin{} and \vfResMax{} (log-uniform), so that the label depends on the size at which the image is shown (\cref{fig:selfcheck}a). The branch is trained on this label with cross-entropy, and no gradient flows through $\mu_\theta$.

\section{Experiments}
\label{sec:exp}

\subsection{Setup}
\label{sec:setup}

\looseness=-1 \paragraph{Models.} We train twelve models of up to 4B from seven families: Qwen3.5~\cite{qwen35}, Qwen3-VL~\cite{qwen3vl}, InternVL3.5~\cite{internvl35}, Gemma-4-E4B~\cite{gemma4}, LFM2.5-VL~\cite{lfm25vl}, LFM3-VL~\cite{lfm3vl} and Rex-Omni~\cite{rexomni} (\cref{app:families}; 7--9B in \cref{app:large}). Each is trained twice on the same data and steps (\cref{app:data,tab:hparams}): as the \emph{point-writing} baseline, fine-tuned the standard way to write the annotated point (the box centre for box records), and as ours, the \emph{decision model} (\cref{sec:training}). Training takes \vTrHoursS{} hours on \vTrGpus{} A100 GPUs at 0.8B (\cref{app:data}), on more data than Intern-Decision's \vOtInternEx{} samples and far less than MJev-Spatial's \vOtMjevQa{} QA pairs.

\looseness=-1 \paragraph{Benchmarks and metrics.} We pool \vBcNAll{} test images from five benchmarks: Point-Bench~\cite{pointarena}, ScreenSpot-v2~\cite{screenspot}, Where2Place~\cite{robopoint}, RoboRefIt~\cite{lu2023vlgrasp} and RefCOCO~\cite{yu2016modeling} (\cref{app:protocol}, \cref{tab:sets}). Every model sees images at \vfEdgeRes, which a small device can afford. A point is right when it lies inside the target, and a $10{\times}10$ or $100{\times}100$ decision is right when its cell meets the target. Confident errors are scored by \emph{wrong@70\%}, the wrong points among a model's most confident 70\% of images. Differences are in pp with paired bootstrap 95\% intervals over images.

\phantomsection\label{tab:speedbody}
\subsection{Reading the decision: accuracy and time}
\label{sec:cost}
\label{sec:main}

\looseness=-1 \paragraph{Off the shelf.} Read our way, off-the-shelf VLMs are already decision models. For eight VLMs that write coordinates as digits, percents or location tokens, the most probable point is as accurate, within \vRwAbsMax{} pp, as the point the model generates with the same prompt (\cref{tab:dec,tab:readwrite}). The mass of the top cell adds a confidence, which is weak before training (AUROC \vRgRexAu{} to \vRzNatReadSPtAu) and useful after it (\cref{sec:closer}). For Qwen3.5, the read takes two passes, since the model writes $y$ after $x$, and is still \vNwBestOverReadMin{} to \vNwBestOverReadMax{} times faster than generating the point (\cref{tab:dec}, \cref{app:engine}). Asked for $x$ and $y$ apart in one pass, Qwen3.5 loses \nosign{\vUtAxmDiffPlain} pp at 2B and \nosign{\vUtAxlDiffPlain} pp at 4B (\cref{app:readout}), so the one-pass read comes from training.

\looseness=-1 \paragraph{Released decision models.} \Cref{tab:dec} scores each decision on three grids. Option models work only on coarse grids (\cref{tab:coarsegrid}): at best they find \vRzOptCellMax\% of $10{\times}10$ cells (\vOhPpOneCell\% when pplx-decider is asked in one round of 100 options, \cref{app:opttime}), against \vRzNatMCellAcc\% for the off-the-shelf Qwen3.5-2B read our way, and \vRzOptPtMax\% of points, in three rounds (\cref{fig:kinds}, \cref{tab:releasedauroc}). Reading 256 token labels per axis is the fastest but falls behind at the point, and crops are accurate but several times slower than ours, which is also more accurate at that size.

\paragraph{The same VLM fine-tuned to write its point.} Ours is more accurate for all twelve models, by \vElGainMin{} to \vElGainMax{} pp, with every 95\% interval above zero (\cref{tab:decb}). Ours is also \nosign{\vQvTwoOverOts} pp above the off-the-shelf Qwen3-VL-2B (\cref{tab:qwen3vl}), three times faster and about half as often wrong on its most confident decisions (\cref{tab:dec}). The gain does not depend on the coordinate format: the LFM models write each value as one token and Rex-Omni-3B as one of 1000 location tokens, and both gain like the digit models (Rex-Omni by \vNdDfPt{} pp; \cref{app:readout}).

\looseness=-1 \paragraph{Time.} On an A100, the Qwen3.5 decision models up to 4B read the full map \vSpStoplRatioThou--\vSpStopmRatioThou{} times faster than the point-writing models write the point in vLLM (\cref{tab:dec}), and \vNwFreeOverMapMin{} to \vNwFreeOverMapMax{} times faster than these write the whole answer on our engine (\cref{app:engine}). The fastest way to write is structured decoding on our engine, where the format of the answer is given and only its six digits are written, and the full map is still read \vDvTplOverFullMin{} to \vDvTplOverFullMax{} times faster, on an A100, an L40S and an L4 (\cref{app:devices}). Structured decoding changes only the time of the point-writing model, not its accuracy or confidence: it follows one path through the tree of digit prefixes and returns one point, while our read covers the whole tree and also answers an agent's yes/no questions (\cref{sec:closed}). On a Raspberry Pi~5, our 0.8B reads a click decision in \vPqClickS\,s and the $100{\times}100$ grid in \vPqHundS\,s (\cref{app:edge}).

\arxivwidenext{1}\begin{figure}[!t]
\centering
\includegraphics[width=\linewidth]{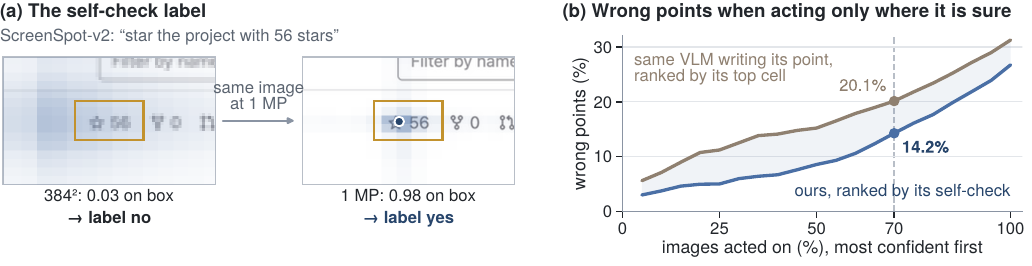}
\vspace{-1.4em}
\caption{\textbf{The self-check, and acting only where it says yes.} Qwen3.5-2B. (a) The training label on one ScreenSpot-v2 image, with the model's map in blue and the target's box in gold. The label is no at $384^2$, where under half of the mass lies on the box, and yes at 1\,MP. (b) Wrong points among the images acted on, most confident first, on the \vBcNAll{} test images, with ours ranked by its self-check and the point-writing model by its top $10{\times}10$ cell mass. One training run each; \cref{tab:decb} gives the means over seeds.}
\label{fig:selfcheck}\label{fig:decide}
\vspace{-1em}
\end{figure}

\begin{figure*}[!t]
\centering
\includegraphics[width=\textwidth]{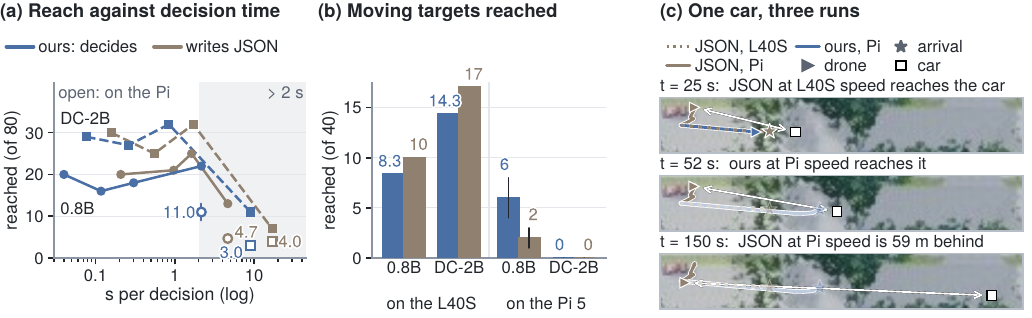}
\caption{\textbf{Drone episodes against the time per decision.} \vDrEpis{} DroneCATS episodes. (a) Episodes reached on the simulator's frames, with each decision delayed. Each line is one model (0.8B solid, DC-2B dashed), from its L40S time on the left to its own Pi time, and the shaded band is beyond \vSetReachLim\,s per decision. Open markers add the Pi's \vSetPiFrame-pixel frame. (b) The \vDrMovEpis{} moving targets. (c) One tracking episode.}
\label{fig:dronelat}
\vspace{-1.4em}
\end{figure*}

\begin{table}[!t]
\centering\footnotesize
\setlength{\tabcolsep}{1.0pt}
\newcommand{\drh}[2]{\begin{tabular}[b]{@{}c@{}}#1\\#2\end{tabular}}
\begin{tabular}{@{}l@{\hspace{2pt}}c@{\hspace{4pt}}cccc@{\hspace{4pt}}c@{}}
\toprule
& & \multicolumn{4}{c}{reached, of \vDrScEpis{} each\,$\uparrow$} & \\
\cmidrule(lr){3-6}
& \drh{s per}{decision\,$\downarrow$} & \drh{in}{view} & \drh{}{search} & \drh{}{track} & \drh{search}{+ track} & \drh{all, of}{\vDrEpis\,$\uparrow$} \\
\midrule
\multicolumn{7}{@{}l}{\textbf{L40S (GPU)}} \\
0.8B, writes JSON & \vLrQJSecGpu & \vLrQJAppGpu & \vLrQJSeaGpu & \vLrQJTraGpu & \vLrQJSetGpu & \vLrQJGpuMean{\scriptsize\,$\pm$\vLrQJGpuSd} \\
\oursrow 0.8B, ours: decides & \textbf{\vLrQDSecGpu} & \vLrQDAppGpu & \vLrQDSeaGpu & \vLrQDTraGpu & \vLrQDSetGpu & \vLrQDGpuMean{\scriptsize\,$\pm$\vLrQDGpuSd} \\
DC-2B, writes JSON & \vLrCJSecGpu & \vLrCJAppGpu & \vLrCJSeaGpu & \vLrCJTraGpu & \textbf{\vLrCJSetGpu} & \textbf{\vLrCJGpuMean}{\scriptsize\,$\pm$\vLrCJGpuSd} \\
\oursrow DC-2B, ours: decides & \vLrCDSecGpu & \textbf{\vLrCDAppGpu} & \textbf{\vLrCDSeaGpu} & \textbf{\vLrCDTraGpu} & \vLrCDSetGpu & \vLrCDGpuMean{\scriptsize\,$\pm$\vLrCDGpuSd} \\
\midrule
\multicolumn{7}{@{}l}{\textbf{At the Raspberry Pi 5's decision time and frame}} \\
0.8B, writes JSON & \vPiMeanQJ & \vNtQJAppPiS & \vNtQJSeaPiS & \vNtQJTraPiS & \vNtQJSetPiS & \vNtQJPiS{\scriptsize\,$\pm$\vSdQJPiS} \\
\oursrow 0.8B, ours: decides & \textbf{\vPiMeanQD} & \textbf{\vDrScAppQDPiS} & \vDrScSeaQDPiS & \textbf{\vDrScTraQDPiS} & \textbf{\vDrScSetQDPiS} & \textbf{\vDrReachQDPiS}{\scriptsize\,$\pm$\vSdQDPiS} \\
DC-2B, writes JSON & \vPiMeanCJ & \vNtCJAppPiS & \textbf{\vNtCJSeaPiS} & \vNtCJTraPiS & \vNtCJSetPiS & \vNtCJPiS{\scriptsize\,$\pm$\vSdCJPiS} \\
\oursrow DC-2B, ours: decides & \vPiMeanCD & \vDrScAppCDPiS & \vDrScSeaCDPiS & \vDrScTraCDPiS & \vDrScSetCDPiS & \vDrReachCDPiS{\scriptsize\,$\pm$\vSdCDPiS} \\
\bottomrule
\end{tabular}

\caption{\textbf{Drone episodes on a GPU and on the edge.} Episodes that reach the target, \vDrScEpis{} per scenario. A decision takes the model's own time on the L40S, and on the Pi~5 its median time measured there (\cref{tab:piinputs}), with the Pi's \vSetPiFrame-pixel frame. Cells are means over three runs, with the s.d.\ of the total, and bold is the best of each block. Every model acts with the same three actions. 0.8B is Qwen3.5-0.8B trained with the DroneCATS recipe and DC-2B the released DroneCATS-2B, and the shaded rows train them further with our recipe to decide (\cref{app:data}).}
\label{tab:dronebody}
\vspace{-1.4em}
\end{table}

\subsection{When to act}
\label{sec:conformal}
\label{sec:closer}

\looseness=-1 The self-check is the model's yes/no answer to whether it can point precisely, and we rank the test images by it, most confident first. \Cref{fig:selfcheck}b shows the share of wrong points among the images acted on: every decision model acts on fewer wrong points than the point-writing model ranked by its top cell mass (\cref{tab:decb}). The self-check is not the whole gain: trained with it, a point-writing 2B has a wrong@70\% of \vPcWrYesPtsQ\%, against \vSvmOurs\% for ours (\cref{app:ptsq}). Even against the point-writing model ranked by its best training-free score, ours is wrong \vRcElRelBestMin--\vRcElRelBestMax\% less often in relative terms, and the self-check ranks errors better than every such score of our own map, digit entropy~\cite{pace} included (\cref{tab:freeconf}).

The same self-check can ask for a closer look. At 2B it says no on \vNmFinNoShare\% of images, so most decisions keep the cost of the small image, and reading the others again at 1\,MP beats re-reading as many at random (\cref{fig:reread}, \cref{tab:closer}).

\subsection{Decisions in a closed loop}
\label{sec:closed}
\label{sec:drone}

\looseness=-1 \paragraph{Drone setting.} DroneCATS~\cite{dronecats} has \vDrEpis{} episodes, \vDrScEpis{} in each of four scenarios: a target in view, one to search for, and a moving car to track or to search for and track. We take two backbones that write the benchmark's JSON action, a Qwen3.5-0.8B we train with the DroneCATS recipe and the released DroneCATS-2B, and train each further with our recipe (\cref{app:drone}). Our agent asks three questions per frame (visible? reached? where?) and moves by a fixed rule (\cref{app:protocol}). The simulator runs in real time, so a slow decision acts on an old frame.

\looseness=-1 \paragraph{Drone results.} On the L40S GPU the 0.8B decision model and the same backbone writing JSON reach the target about equally often, so our further training and the agent's rule give no advantage of their own. The models differ in the time each decision takes: the episodes reached hold up to about \vSetReachLim\,s per decision and fall beyond it (\cref{fig:dronelat}a). On the Pi the 0.8B model needs \vPiMeanQJ\,s to write an action and \vPiMeanQD\,s, near that limit, to read one. With each decision held back to its Pi time, the writer falls from \vNtQJGpu{} to \vNsQJPiReach{} episodes and ours does not. The Pi's smaller frame then costs every model, and ours keeps the most, \vDrReachQDPiS\,$\pm$\,\vSdQDPiS{} episodes against \vNtQJPiS\,$\pm$\,\vSdQJPiS{} for the writer (\cref{tab:dronebody}, \cref{fig:dronelat}b,c). At 2B even the read takes over \vSetReachLim\,s, and ours reaches fewer than the writer.

\begin{table}[!t]
\centering\footnotesize
\setlength{\tabcolsep}{1.2pt}
\providecommand{\stwo}[2]{\begin{tabular}[b]{@{}c@{}}#1\\#2\end{tabular}}
\begin{tabular}{@{}lcccc@{}}
\toprule
& \multicolumn{3}{c}{\stwo{Mind2Web}{(offline)}} & \stwo{MiniWoB++}{(live)} \\ \cmidrule(lr){2-4}
& \stwo{click}{acc.\,$\uparrow$} & \stwo{wrong}{@70\%\,$\downarrow$} & \stwo{safe}{tasks\,$\uparrow$} & \stwo{success}{(\%)\,$\uparrow$} \\
\midrule
Qwen3.5-0.8B, writes & \vWbMwPtsZeroEightAcc & \vWbMwPtsZeroEightWrong & \vWtSafePtsZeroEight & \vNsSFTMw \\
\oursrow Qwen3.5-0.8B, ours & \textbf{\vWbMwOursZeroEightAcc} & \textbf{\vWbMwOursZeroEightWrong} & \textbf{\vWtSafeOursZeroEight} & \textbf{\vNsFinMw} \\
\addlinespace[2pt]
GUI-Owl-1.5-2B$^{\ast}$~\cite{xu2026mobileagent35} & \textbf{\vGoAcc} & \vGoWrong & \vGoSafe & \textbf{\vGoMw} \\
GUI-Decisions-2B$^{\ast}$~\cite{guidecisions} & \vWbRelGuidAcc & \vWbRelGuidWrong & \vWtSafeGuid & \vNguidMw \\
Qwen3.5-2B, writes & \vWbMwPtsTwoAcc & \vWbMwPtsTwoWrong & \vWtSafePtsTwo & \vNmSFTMw \\
\oursrow Qwen3.5-2B, ours & \vWbMwOursTwoAcc & \textbf{\vWbMwOursTwoWrong} & \textbf{\vWtSafeOursTwo} & \vNmFinMw \\
\addlinespace[2pt]
GroundingPI-4.8B~\cite{yu2026groundingpi} & \vWbRelGpiAcc & \vWbRelGpiWrong & \vWtSafeGpi & \vNgpiMw \\
MJev-Spatial-4B$^{\ast}$~\cite{jevspatial} & \vWbRelMjevAcc & \vWbRelMjevWrong & \vWtSafeMjev & \textbf{\vNmjevMw} \\
Qwen3.5-4B, writes & \vWbMwPtsFourAcc & \vWbMwPtsFourWrong & \vWtSafePtsFour & \vNlSFTMw \\
\oursrow Qwen3.5-4B, ours & \textbf{\vWbMwOursFourAcc} & \textbf{\vWbMwOursFourWrong} & \textbf{\vWtSafeOursFour} & \vNlFinMw \\
\bottomrule
\end{tabular}

\caption{\textbf{Web agents, one step offline and on live pages.} Mind2Web scores one click per step (\%) on pages of 0.6 megapixels, and a task is safe when none of the model's most confident 70\% of steps is wrong. MiniWoB++ scores the live page after one click (\vSetMwTasks{} tasks, 50 seeds each). ``Writes'' is the point-writing model, and bold is the best of each size. Ours is ranked by its self-check, the other pointing models by their top $10{\times}10$ cell mass, and MJev-Spatial by the product of its options' probabilities.  $^{\ast}$Built for GUI agents; GUI-Owl is in its own tool-call format.}
\label{tab:webbody}
\vspace{-1.4em}
\end{table}

\arxivwidenext{1}\begin{figure}[!t]
\centering
\includegraphics[width=\linewidth]{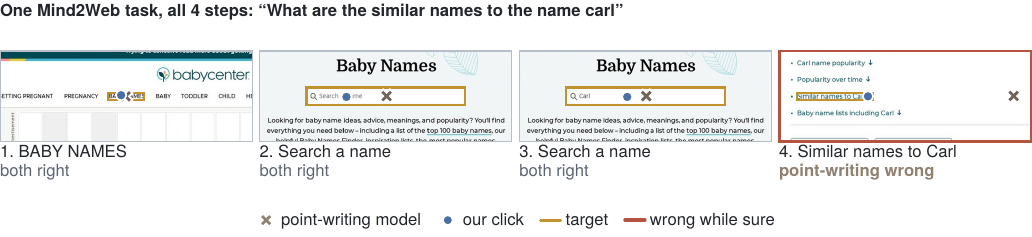}
\vspace{-1.6em}
\caption{\textbf{One web task, step by step.} Every step of one Mind2Web task for the point-writing Qwen3.5-2B and ours, cropped around the element to click. At the last step the point-writing model clicks the blank space beside the link although the step is among its most confident (red frame), and ours clicks the link.}
\label{fig:webstep}
\vspace{-1.0em}
\end{figure}

\looseness=-1 \paragraph{Web agents.} On Multimodal-Mind2Web~\cite{deng2023mind2web,zheng2024seeact}, ours clicks correctly about as often as the point-writing model and at 2B \vGoOverOurs{} pp less often than GUI-Owl-1.5-2B. Its gain is in confident errors: on its most confident 70\% of steps, ours makes fewer wrong clicks and so keeps more tasks free of such a click than any other model of its size (\cref{tab:webbody}, \cref{fig:webstep}; larger images in \cref{app:ownpix}). On live MiniWoB++~\cite{miniwob} our Qwen3.5 models succeed more often than the point-writing ones at every size (\cref{tab:miniwob}), and less often than GUI-Owl-1.5-2B and MJev-Spatial-4B, built for GUIs.

\section{Analysis}
\label{sec:analysis}

\begin{figure}[!t]
\centering
\includegraphics[width=\linewidth]{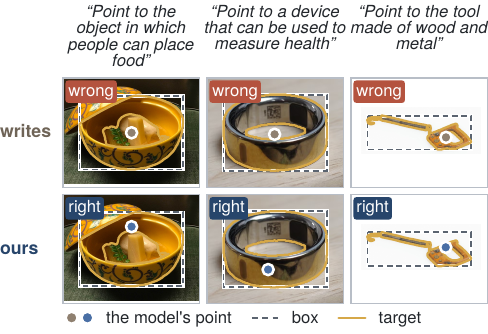}
\vspace{-1.6em}
\caption{\textbf{Qualitative comparison of the point-writing Qwen3.5-2B (top) and ours (bottom).} The point-writing model tends toward the centre of the box, here the food in the bowl, the hole of the ring and the opening of the saw's handle, and ours points at the named object itself.}
\label{fig:qual}
\vspace{-1.0em}
\end{figure}

\subsection{Where the gain comes from}
\label{sec:sets}
\label{sec:ablate}
\label{sec:box}

The point-writing model is biased toward the centre of the box, while ours points at the named object itself (\cref{fig:qual}). On the \vQlPoolN{} test images whose box centre is not on the object (\cref{app:centres}), the point-writing 2B points at the object on \vQlPoolPts\% and ours on \vQlPoolOurs\% (\vQlPoolDiff{} pp, 95\% CI \vQlPoolLo{} to \vQlPoolHi), as \cref{prop:sets} predicts.

\looseness=-1 On those images the point-writing model points at the box centre \vPnbPtCen\% of the time and ours \vPnbOuCen\%. A point-writing model trained without box records does not have this bias, but it ends \nosign{\vPnbDOuNbAcc} pp below ours overall (\cref{app:pointsnobox}), so set labels let box data be used without the bias. The gain is not limited to such images: ScreenSpot-v2, whose box centres always lie on the element, gains for every model (\cref{tab:sets}).

\subsection{What each part of the recipe adds}
\label{sec:ablation}

\begin{table}[!t]
\centering\footnotesize
\setlength{\tabcolsep}{0.6pt}
\begin{tabular}{@{}l@{\hspace{4pt}}c@{\hspace{6pt}}c@{\hspace{6pt}}c@{}}
\toprule
Qwen3.5-2B & \begin{tabular}[b]{@{}c@{}}point\\acc.\,(\%)\end{tabular} & \begin{tabular}[b]{@{}c@{}}wrong\\@70\%\end{tabular} & AUROC \\
\midrule
writes its point & \vAsPtsPtMean{\scriptsize\,$\pm$\vAsPtsPtSd} & \vAsPtsWrongMean{\scriptsize\,$\pm$\vAsPtsWrongSd} & \vAsPtsAuMean{\scriptsize\,$\pm$\vAsPtsAuSd} \\
\ + each axis its own question & \vAsParPtMean{\scriptsize\,$\pm$\vAsParPtSd} & \vAsParWrongMean{\scriptsize\,$\pm$\vAsParWrongSd} & \vAsParAuMean{\scriptsize\,$\pm$\vAsParAuSd} \\
\ + box as a label set & \vAsSetPtMean{\scriptsize\,$\pm$\vAsSetPtSd} & \vAsSetWrongMean{\scriptsize\,$\pm$\vAsSetWrongSd} & \vAsSetAuMean{\scriptsize\,$\pm$\vAsSetAuSd} \\
\quad + box-edge questions & \vAsSetBPtMean{\scriptsize\,$\pm$\vAsSetBPtSd} & \vAsSetBWrongMean{\scriptsize\,$\pm$\vAsSetBWrongSd} & \vAsSetBAuMean{\scriptsize\,$\pm$\vAsSetBAuSd} \\
\quad + self-check question & \vAsSetQPtMean{\scriptsize\,$\pm$\vAsSetQPtSd} & \vAsSetQWrongMean{\scriptsize\,$\pm$\vAsSetQWrongSd} & \vAsSetQAuMean{\scriptsize\,$\pm$\vAsSetQAuSd} \\
\oursrow \quad + edges and self-check (ours) & \vAsOursPtMean{\scriptsize\,$\pm$\vAsOursPtSd} & \vAsOursWrongMean{\scriptsize\,$\pm$\vAsOursWrongSd} & \vAsOursAuMean{\scriptsize\,$\pm$\vAsOursAuSd} \\
\bottomrule
\end{tabular}

\caption{\textbf{What each part of the Qwen3.5-2B recipe adds.} All rows use the same data and steps, and cells are means and standard deviations over three seeds. Wrong@70\% and AUROC rank by the self-check where a row has one and by the top-cell mass otherwise. Other design choices are in \cref{tab:ablationfull}.}
\label{tab:ablation}
\end{table}

\looseness=-1 Training the box as a label set raises accuracy from \vAsParPtMean{} to \vAsSetPtMean\%, well beyond the seed spread (\cref{tab:ablation}). Asking each axis as its own question adds less, from \vAsPtsPtMean{} to \vAsParPtMean\% (\cref{tab:ablation}; one seed in \cref{tab:layout}), and the box-edge questions let the model also return a box. The self-check cuts wrong@70\% from \vAsSetWrongMean\% for the set-trained model to \vAsSetQWrongMean\%, and to \vAsOursWrongMean\% with the box-edge questions added. Without the self-check, that model ranks its errors no better than the point-writing model.

\subsection{A new head falls behind on fine grids}
\label{sec:head}

\looseness=-1 We also compare the digits with a new output head that scores bins. We train a linear head with one value per cell, 10, 100 or 1000 per axis, on the same model, data and steps (\cref{tab:head}). With 10 values the head is about as accurate as the digits, and the finer the head, the further it falls behind, by \nosign{\vHtDPt} pp with 1000 values. A head gives each of the 1000 values its own output, while the digits factor the choice into three ten-way choices and start from what the model knows about numbers.

\begin{table}[!t]
\centering\footnotesize
\setlength{\tabcolsep}{2pt}
\begin{tabular}{@{}l@{\hspace{8pt}}ccc@{}}
\toprule
& \multicolumn{3}{c}{accuracy (\%) on a grid of} \\
\cmidrule(l){2-4}
Qwen3.5-2B, box as a set & $10{\times}10$ & $100{\times}100$ & $1000{\times}1000$ \\
\midrule
\oursrow reads the digits & \vHzDigitsCell & \vHzDigitsHund & \vHzDigitsPt \\
new head, one value per cell & \vHzTenCell & \vHzHundHund & \vHtHeadPt \\
\midrule
head minus digits (pp) & \textcolor{gray}{\vHzTenD} & \textbf{\vHzHundD} & \textbf{\vHtDPt} \\
\bottomrule
\end{tabular}

\caption{\textbf{A new head falls behind as it gets finer.} All rows train only the box as a set, with no box-edge or self-check question, on the same data and steps, so the digits row is the set row of \cref{tab:ablation} for one seed (\cref{tab:ablationfull}). Each head is trained and read at its own grid. A bold gap has a 95\% interval that excludes zero, and a grey one includes it.}
\label{tab:head}
\vspace{-1.4em}
\end{table}

\phantomsection\label{tab:freeconfbody}
\phantomsection\label{fig:act}

\section{Limitations and Conclusion}
\label{sec:limits}
\label{sec:conclusion}

\looseness=-1 This work extends decision models to grounding. A decision model reads its answer instead of decoding it, with many questions sharing one prefix, and pointing fits this form: every digit prefix of a coordinate is known in advance, so a VLM that writes coordinates can be read in one pass with no new head. Training each label as a set that contains the right answer makes the read more accurate, and its self-check tells when the model is likely wrong and where a closer look helps. The benefit holds in closed loops, for a drone agent at Raspberry Pi~5 speed and for web agents. Our read returns one point per question, which caps the counting category of Point-Bench, where several points are asked for (\cref{app:published}). We leave three things to future work: larger models, video, and more edge devices, since we tested one, a Raspberry Pi~5, and flew the drone only in a simulator.

\FloatBarrier
\section*{Acknowledgments}
We used Claude (Anthropic) throughout this work, under the author's direction and review: to write code and run experiments, to analyse results, to prepare data and draw figures and tables, to draft and edit the text, including the related work, and to give reviewer-style feedback on drafts. We also obtained reviewer-style feedback on a draft from Deep Review (Google). The author proposed the ideas, designed the study, checked the experiments, numbers, references and claims, and takes full responsibility for the content.

{
\small
\bibliographystyle{ieeenat_fullname}
\bibliography{refs}
}
\clearpage
\appendix
\setcounter{section}{0}\setcounter{table}{0}\setcounter{figure}{0}\setcounter{equation}{0}
\renewcommand{\thesection}{S\arabic{section}}
\renewcommand{\thetable}{S\arabic{table}}
\renewcommand{\thefigure}{S\arabic{figure}}
\renewcommand{\theequation}{S\arabic{equation}}
\setcounter{proposition}{0}\setcounter{lemma}{0}\renewcommand{\theproposition}{S\arabic{proposition}}
\renewcommand{\thetheorem}{S\arabic{theorem}}
\renewcommand{\thelemma}{S\arabic{lemma}}
\renewcommand{\thedefinition}{S\arabic{definition}}
\renewcommand{\theassumption}{S\arabic{assumption}}
\renewcommand{\theremark}{S\arabic{remark}}
\renewcommand{\topfraction}{0.95}\renewcommand{\bottomfraction}{0.8}\renewcommand{\textfraction}{0.05}\renewcommand{\floatpagefraction}{0.5}\renewcommand{\dbltopfraction}{0.95}\renewcommand{\dblfloatpagefraction}{0.5}\setcounter{topnumber}{4}\setcounter{totalnumber}{6}\setcounter{dbltopnumber}{3}  %

{\centering{\huge\optimisticfont\bfseries Appendix\par}\vspace{1.0em}\includegraphics[width=\textwidth]{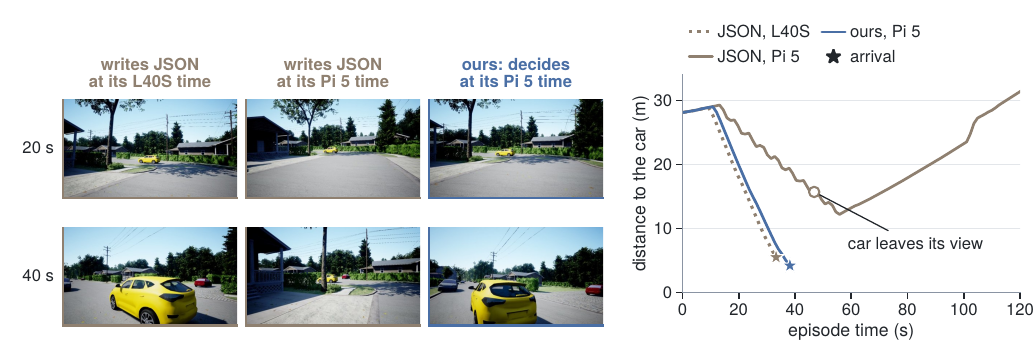}\par\vspace{-0.2em}\captionof{figure}{\textbf{Time per decision decides a tracking episode.} The same JSON writer runs at its L40S time and at its Pi time, and ours at its Pi time, on one tracking episode (not that of \cref{fig:dronelat}c), all on the simulator's frames. Left: the camera frame each model is acting on at 20\,s and 40\,s. Right: the distance to the car. The writer flies toward the car in both settings and differs only in its median time per decision, \vVidGpuSec\,s on the L40S and \vVidPiSec\,s on the Pi. At its Pi time it falls behind until the car leaves its view (circle), whereas ours, at a median of \vVidOursSec\,s per decision, reaches the car. By 40\,s the L40S writer and ours have arrived, so their panels show their last frames. The project page shows the whole episode.}\label{fig:video}%
\par}\vspace{1.0em}

\noindent This appendix gives the details behind the main paper: how we read the decision, how we train the model, the full evaluation protocols, and further results for each comparison.

\noindent\emph{Video.} The project page shows the whole episode of \cref{fig:video}, in which a car keeps driving while each model decides. \Cref{app:video} describes the episode and explains why the time per decision on the device determines its outcome.

\noindent\emph{Overview.} We order the sections from how the decision is read to how it is used.
\begin{itemize}[leftmargin=*,itemsep=0pt,topsep=1pt,parsep=0pt]
\item How the decision is read in one pass, on a GPU and in llama.cpp (\cref{app:readout,app:serving}).
\item What each label estimates and how the model is trained (\cref{app:sets,app:data,app:centres,app:selfteacher}).
\item When to act, and when to read the image again at 1\,MP (\cref{app:falsepos}).
\item Closed loops: the drone and the web (\cref{app:closed}).
\item The protocols, the 7--9B models and the published comparisons (\cref{app:protocol,app:large,app:standard}).
\item Further detail on the comparisons of the main text: how the released option models (decision models that choose among options, as in \cref{tab:dec}) were timed and prompted (\cref{app:opttime}), confident errors within each benchmark (\cref{app:perbench}), the drone agent at 2B and the video (\cref{app:drone2b}), and decision models served through an API (\cref{app:apidec}).
\end{itemize}

\noindent\emph{Conventions.} These hold unless a section says otherwise.
\begin{itemize}[leftmargin=*,itemsep=0pt,topsep=1pt,parsep=0pt]
\item Ours is the decision model. The point-writing model (\emph{writes} in the tables) is the same VLM fine-tuned on the same records for the same number of steps to write the annotated point (the box centre for box records). Our rows and columns are shaded, and $\Delta$ is ours minus the point-writing model.
\item Pointing numbers are on the \vBcNAll{} test images at $384^2$. A read point is the most probable value; a written point is the greedy answer.
\item A 95\% CI is a paired bootstrap interval over images with \vSetBoot{} resamples, and differences are computed before rounding.
\item wrong@70\% is the wrong share among a model's 70\% most confident images, with ours ranked by $P(\text{yes})$, released decision models by their own confidence (\cref{tab:dec}) and the others by the mass of their most probable $10{\times}10$ cell, or $100{\times}100$ where \cref{app:perbench} says so. AUROC measures how well the same confidence separates right from wrong answers.
\end{itemize}

\noindent\emph{Pre-specified choices.} The questions used to obtain the $10{\times}10$ decision (\cref{app:protocol}) were fixed before the models were run, the adoption rule for each recipe variant in \cref{tab:pilots} was written before training, and the data control of \cref{app:seedsna} is reported regardless of its outcome.

\newcommand{\stwo}[2]{\begin{tabular}[b]{@{}c@{}}#1\\#2\end{tabular}}
\renewcommand{\floatpagefraction}{0.75}\renewcommand{\dblfloatpagefraction}{0.75}
\section{Reading the Decision}
\label{app:readout}

\paragraph{Questions.} The prefill holds the image and ``Point to $\ell$.'' in one user turn of the model's chat template. Each question is appended to that turn as its own branch after the shared prefix (the cached image and instruction) and is followed by the assistant header (for Qwen3.5 with an empty thinking block) and ``Answer:~''. The $x$ question reads ``What is the x coordinate of its center? Answer with a three-digit number from 000 (left edge) to 999 (right edge) of the image.'' The $y$ question is the same with ``y coordinate'' and ``(top edge) to 999 (bottom edge)''. The four box-edge questions replace ``its center'' with ``its left edge'', ``its top edge'', ``its right edge'' and ``its bottom edge''. The self-check question is ``Can you point to it precisely at this image resolution? Answer yes or no.'', and $P(\text{yes})$ is the softmax over the \texttt{Yes} and \texttt{No} tokens after ``Answer:~'', the two tokens it is trained on.

\paragraph{Reading the digits.} The next-token distribution after ``Answer:~'' gives the first digit. The ten one-digit and hundred two-digit prefixes that can follow are known in advance, so they enter the same forward pass as the question tails, as nodes of a tree attention mask. Each node attends to the image prefix, its own branch and its own earlier digits at the positions it would have if written, and in the recurrent layers of Qwen3.5 it continues from its parent's state. Hi-Token\newref{hitoken}{Zhu et al., arXiv:2608.03471} also writes a coordinate as hundreds, tens and units, with the value's probability the product over its digits; it adds digit tokens for each axis and decodes them, whereas we read the digits the VLM already writes, for all prefixes at once.

\paragraph{Exactness.} This pass computes the same conditionals as feeding the digits one step at a time, up to floating-point rounding. In fp32 the two reads differ by at most \vOsFpMaxDp{} per value on the $x$ and $y$ distributions of \vAfFpImages{} Point-Bench images (InternVL3.5-2B). In bf16, the precision of the reported evaluation, they differ by at most \vAfBfMaxDp{} per value on \vAfBfImages{} images. The two reads pick the same band of ten values on \vAfBfCellSame{} of the \vAfBfAxes{} axes. The single most probable value agrees on \vAfBfValueSame{} axes, and on the others the two top values are nearly tied, since bf16 logits leave exact ties among the 1,000 values on a third to two fifths of the axes (\cref{app:latency}) and an exact tie goes to the smallest value. Gemma-4-E4B attends within a sliding window on most layers, where this pass is not exact, so we read it with one pass per digit.

In the zero-padded format each digit is renormalized over the ten digit tokens, so the 1,000 values sum to one. The value 347 then has probability $q(3)\,q(4\mid3)\,q(7\mid34)$, the column 300--399 has mass $q(3)$ and the column 340--349 mass $q(3)\,q(4\mid3)$, and a cell has the product of the masses of its column and its row (\cref{app:sets}).

\paragraph{Asking for $x$ and $y$ apart without training.} Reading both axes in one pass needs our training, since an untrained model answers a question about one axis alone less accurately than it writes $y$ after $x$. An off-the-shelf Qwen3.5 writes $y$ after its own $x$, so our read of its answer takes two passes. The second pass conditions $y$ on the one $x$ the model would write, $x^{*}$, so the map is $p(x)\,p(y\mid x^{*})$ and not the joint distribution over both axes. Read in one pass in its own format, with $y$ asked after a fixed $x$ of 500 so that it need not wait for $x$, the untrained model is right on \vUtAxsAcc, \vUtAxmAcc{} and \vUtAxlAcc\% at 0.8B, 2B and 4B, against \vUtAxsTwo, \vUtAxmTwo{} and \vUtAxlTwo\% with the two-pass read ($\vUtAxsDiff$ [$\vUtAxsLo$, $\vUtAxsHi$], $\vUtAxmDiff$ [$\vUtAxmLo$, $\vUtAxmHi$] and $\vUtAxlDiff$ [$\vUtAxlLo$, $\vUtAxlHi$] pp, with 95\% CIs in brackets).

\paragraph{Training a model that writes location tokens.} Our recipe also works for a model that writes each coordinate as one location token. Trained as a decision model, Rex-Omni-3B (Qwen2.5-VL-3B) is right on \vNdOursPt\% against \vNdWrPt\% for the same model trained to write its point, and makes fewer confident errors: on the most confident 70\%, ours is wrong on \vNdOursWrong\% against \vNdWrWrong\%, with AUROC \vNdOursAuc{} against \vNdWrAuc. Here ours is ranked by its self-check and the point-writing model by the mass of its most probable $100{\times}100$ cell. Most of the gain is on RefCOCO. We keep Rex-Omni's 1,000 location tokens, so each axis, box edge and self-check of ours is one read over them. Both models use the data, steps, settings and question wording of our other rows (a three-digit number), so the training data stay the same. Its authors train Rex-Omni for pointing and detection, and off the shelf its written point is right on \vRwRexWrite\% (\cref{tab:readwrite}). Ours matches the released model and adds the self-check. Ranked by the mass of its top $100{\times}100$ cell instead, ours is wrong on \vNdOursPeakWrong\% (AUROC \vNdOursPeakAuc), so it is the self-check that ranks its errors.

\paragraph{Time with location tokens.} With one location token per coordinate, our read costs more than the bare point and adds the map, the box and the self-check. In the HF engine our full read takes \vLHrexFull\,ms. The point-writing model takes \vLHrexGenFmt\,ms for the eight tokens of a complete answer, which gives the \vLHrexRatioFFmt$\times$ of \cref{tab:decb}, and \vLHrexGen\,ms for what it actually writes, since it rarely closes its answer and continues to \vLHrexTok{} tokens on average. Its bare point is only three tokens (\vLHrexGenMin\,ms).

\begin{table}[!htbp]
\centering\footnotesize
\setlength{\tabcolsep}{3pt}
\begin{tabular}{@{}l@{\hspace{10pt}}l@{\hspace{12pt}}c@{\hspace{8pt}}c@{\hspace{14pt}}c@{}}
\toprule
& & \multicolumn{2}{c}{point acc.\ (\%)\,$\uparrow$} & \\
\cmidrule(lr){3-4}
Model & format & written & read & AUROC\,$\uparrow$ \\
\midrule
Qwen3.5-0.8B & digits & \vRwQSWrite & \vRwQSRead & \vRzNatReadSPtAu \\
Qwen3.5-2B & digits & \vRwQMWrite & \vRwQMRead & \vRzNatReadMPtAu \\
Qwen3.5-4B & digits & \vRwQLWrite & \vRwQLRead & \vRzNatReadLPtAu \\
Qwen3-VL-2B & digits & \vRwQvlTwoWrite & \vRwQvlTwoRead & \vQnTwoAu \\
Qwen3-VL-4B & digits & \vRwQvlFourWrite & \vRwQvlFourRead & \vQnFourAu \\
Molmo-7B-D & percent & \vRwMolmoWrite & \vRwMolmoRead & \vRgMolmoAu \\
GroundingPI & loc. & \vRwGpiWrite & \vRwGpiRead & \vRgGpiAu \\
Rex-Omni & loc. & \vRwRexWrite & \vRwRexRead & \vRgRexAu \\
\midrule
\multicolumn{5}{@{}l}{\textit{The base VLMs of our other trained rows, off the shelf}} \\
InternVL3.5-1B$^\dagger$ & box & \vRxIvlAWrite & \vRxIvlARead & \vRxIvlAAu \\
InternVL3.5-2B$^\dagger$ & box & \vRxIvlWrite & \vRxIvlRead & \vRxIvlAu \\
InternVL3.5-4B$^\dagger$ & box & \vRxIvlDWrite & \vRxIvlDRead & \vRxIvlDAu \\
Gemma-4-E4B$^\ast$ & digits & \vRxGemWrite & \vRxGemRead & \vRxGemAu \\
LFM2.5-VL-3B$^\ddagger$ & 1 tok. & \vRxLfmWrite & \vRxLfmRead & \vRxLfmAu \\
LFM3-VL-2B$^\ddagger$ & 1 tok. & \vRxLfvWrite & \vRxLfvRead & \vRxLfvAu \\
\bottomrule
\end{tabular}

\caption{\textbf{The read works across coordinate formats, with no training.} Off-the-shelf VLMs that write coordinates as digits, percents, one token per value (1 tok.) or location tokens (loc.). The most probable point of our read is as accurate as the point each model of the upper block writes with the same prompt, and the read adds a confidence, the mass of its most probable $10{\times}10$ cell ($100{\times}100$ for Qwen3-VL, the map peak for GroundingPI, Rex-Omni and Molmo), whose AUROC for separating right from wrong points is in the last column (chance is 0.5). The upper block holds the eight VLMs read in the main text, each writing a point in its documented format. $^\ast$Gemma-4 writes a point as $[y, x]$. $^\dagger$InternVL3.5 follows a pointing instruction poorly (\vRxIvlAPlain{} to \vRxIvlDPlain\% right), so it writes a box in its own format and its point is the centre of four edge reads. This centre is \vRxBoxGapMin{} to \vRxBoxGapMax{} pp less accurate than the written one, as an uncertain edge's most probable value is often the border, 0 or 1000. $^\ddagger$The LFM models document no coordinate format, so they are given a plain pointing instruction. Off-the-shelf times are in \cref{app:engine}.}
\label{tab:readwrite}
\end{table}

\paragraph{Coordinate formats.} \Cref{tab:families} reads nine off-the-shelf models in their own formats. The one-pass read applies unchanged whenever the first coordinate token picks a group of values, as for zero-padded digits, decimals of three fixed digits (RoboPoint), one token per value (LFM2.5-VL) or location tokens (GroundingPI, Rex-Omni).

Integers without leading zeros (Qwen3-VL, Qwen3.5) and percentages (Molmo) need one more step, since a first digit $t$ can begin $t$, $10t$ to $10t{+}9$ or $100t$ to $100t{+}99$. For these formats we read every prefix of the digit tree in one batched call and renormalize each node over the digits and the end that can follow it, so the value 73 has probability $q(7)\,q(3\mid7)\,q(\text{end}\mid73)$ and the read stays exact. The end is the summed probability of every token that begins with the character following the number in the model's own format, a comma after $x$ and a closing bracket after $y$ for Qwen. The tree holds only the spellings the format allows: a first digit 0 is the value 0, leading zeros are not read, and a fourth digit is read only after 100, where the end gives 100 and a further 0 gives 1000. The 1,001 values of an integer axis then sum to one with no final rescaling.

The accuracies of off-the-shelf Qwen3.5 come from this stepped read. Only its times in \cref{tab:dec} come from a packed read, in which the whole tree of $x$ is packed into one pass and that of $y$, after the greedy $x$, into a second (\vUtFastsThou, \vUtFastmThou{} and \vUtFastlThou\,ms at 0.8B, 2B and 4B). The packed read computes the same products, and on the same logits the two agree to within \vAfArithDp{} per value. It computes its logits in fp32, while the stepped read takes bf16 logits for each prefix, and since $y$ is conditioned on the greedy $x$, a near-tied last digit of $x$ can change the $y$ map. The two maps therefore differ by up to \vUtFastlGrDp{} per value.

\begin{table}[!htbp]
\centering\footnotesize
\setlength{\tabcolsep}{1.6pt}
\begin{tabular}{@{}llc>{\columncolor{oursbg}}cc@{}}
\toprule
 & & & \multicolumn{2}{c}{\stwo{90\% region}{(\% of image) $\downarrow$}} \\ \cmidrule(l){4-5}
Model & \stwo{passes to}{write a value} & \stwo{written point}{acc. (\%)\,$\uparrow$} & \stwo{read}{map} & \stwo{disk around}{written point} \\
\midrule
\multicolumn{5}{@{}l}{\textit{Integer without leading zeros: $x=52$ is written \texttt{5}\,\texttt{2}}} \\
\quad Qwen3-VL-4B~\cite{qwen3vl} & one per digit & 78.7 & 0.9 & 7.1 \\
\quad Qwen3-VL-2B~\cite{qwen3vl} & one per digit & 72.7 & 2.7 & 11.0 \\
\quad Qwen3.5-2B~\cite{qwen35} & one per digit & 54.9 & 5.3 & 15.8 \\
\quad Qwen3.5-4B~\cite{qwen35} & one per digit & 70.8 & 2.3 & 9.2 \\
\addlinespace[2pt]
\multicolumn{5}{@{}l}{\textit{One token per value: \texttt{52} is a single token}} \\
\quad LFM2.5-VL-3B~\cite{lfm25vl} & one & 73.4 & 0.7 & 7.2 \\   %
\addlinespace[2pt]
\multicolumn{5}{@{}l}{\textit{Location token: \texttt{<52>}}} \\
\quad Rex-Omni~\cite{rexomni} & one & 77.1 & 2.8 & 10.6 \\
\quad GroundingPI~\cite{yu2026groundingpi} & one & 78.6 & 0.8 & 10.1 \\
\addlinespace[2pt]
\multicolumn{5}{@{}l}{\textit{Percent: \texttt{5.2}}} \\
\quad Molmo-7B-D~\cite{molmo} & one per digit & 75.2 & 1.9 & 8.1 \\
\addlinespace[2pt]
\multicolumn{5}{@{}l}{\textit{Decimal fraction: \texttt{0.052}}} \\
\quad RoboPoint-13B~\cite{robopoint} & one per digit & 46.4 & 18.2 & 28.7 \\
\bottomrule
\end{tabular}

\caption{\textbf{Off-the-shelf VLMs read in their own coordinate formats.} Each group heading shows how the format writes a coordinate 5.2\% of the way across the image. Our read batches every prefix of the format in one call. Accuracy is that of the point each model writes, on \vfNrecs{} test records; RoboPoint-13B was read only on these, so it is not in \cref{tab:readwrite}, which uses all \vBcNAll{} test images. The last two columns give the area of the 90\% region of the read map and of a disk around the written point, each calibrated to meet the target on 90\% of images (\cref{app:conformal}).}
\label{tab:families}
\end{table}
\section{Theory: What Each Label Estimates}
\label{app:sets}

\paragraph{Setting.} A question has a finite set of answers $\Ys$: the 1,000 three-digit values of an axis for a pointing question, or the option codes of a multiple-choice field. A model gives a probability vector $\mu(\cdot\mid x)$ on $\Ys$. A record has a view $\xp$ (the image with the question), and the model is shown $x=\rho(\xp)$, a fixed function of it such as the image at the deployment resolution. We write $\mu(A\mid x)=\sum_{y\in A}\mu(y\mid x)$, $\mu(\cdot\mid A)$ for $\mu$ restricted to $A$ and rescaled, $\CE(t,\mu)=-\sum_yt(y)\log\mu(y)$, $\KL(t\|\mu)=\sum_yt(y)\log\frac{t(y)}{\mu(y)}$ and $H(t)=-\sum_yt(y)\log t(y)$. Expectations average over records and over the label they carry.

\paragraph{Labels.} The right answer is $Y\in\Ys$ with distribution $\ps(\cdot\mid\xp)$, and the label is a set $A\ni Y$ drawn from a rule $\kappa(A\mid\xp,Y)$. Such a label is coarse data~\cite{gill1997car}.
\refstepcounter{assumption}\label{th:car}\emph{Assumption~\theassumption{} (coarsening at random): $Y\in A$ always, and $\kappa(A\mid\xp,y)=\kappa(A\mid\xp,y')=:\kappa_A(\xp)$ for all $y,y'\in A$.} A gold option ($A=\{g\}$) satisfies it, and a box drawn for an object approximately does, whichever of its points is intended.

\begin{lemma}[A fixed target gives a fixed optimum]\label{lem:th:lin}
If $T$ is a random target distribution on $\Ys$ and $\mu$ may use only the information in $\mathcal H$ ($\mu$ is $\mathcal H$-measurable), then $\E[\CE(T,\mu)\mid\mathcal H]=\CE(\E[T\mid\mathcal H],\mu)$, and $\E\,\CE(T,\mu)$ is minimized only by $\mu=\E[T\mid\mathcal H]$.
\end{lemma}
\begin{proof}
$\CE(t,\mu)$ is linear in $t$, and given $\mathcal H$ the factor $-\log\mu$ is fixed. With $\bar t=\E[T\mid\mathcal H]$, $\CE(\bar t,\mu)=H(\bar t)+\KL(\bar t\|\mu)$, which by Gibbs' inequality is smallest only at $\mu=\bar t$.
\end{proof}

\begin{lemma}[Label term plus shape term]\label{lem:th:dec}
If $q(A)=1$ and $\mu(A)>0$, then $\CE(q,\mu)=-\log\mu(A)+\KL(q\|\mu(\cdot\mid A))+H(q)$.
\end{lemma}
\begin{proof}
For $y\in A$, $\log\mu(y)=\log\mu(A)+\log\mu(y\mid A)$. Multiplying by $-q(y)$ and summing gives $-\log\mu(A)+\CE(q,\mu(\cdot\mid A))$, and $\CE(q,\nu)=H(q)+\KL(q\|\nu)$.
\end{proof}

\par\vskip 0pt plus 8\baselineskip\penalty-100\vskip 0pt plus -8\baselineskip
{\renewcommand{\theproposition}{1 (restated)}\addtocounter{proposition}{-1}\begin{proposition}[What each label estimates]\label{th:sets}
Under \cref{th:car}, the label set has the law $r(A\mid\xp)=\sum_{y\in A}\ps(y\mid\xp)\kappa(A\mid\xp,y)=\kappa_A(\xp)\,\ps(A\mid\xp)$~\cite{heitjan1991ignorability}.
\begin{enumerate}[label=(\alph*),leftmargin=*,itemsep=1pt]
\item \emph{Sets.} Over maps $\xp\mapsto\mu(\cdot\mid\xp)$, the risk $\E[-\log\mu(A\mid\xp)]$ is minimized by $\mu=\ps(\cdot\mid\xp)$, and every minimizer satisfies $\mu(A\mid\xp)=\ps(A\mid\xp)$ for every set $A$ with $r(A\mid\xp)>0$.
\item \emph{Training on one point.} If the trainer replaces the set by one point $g(A)\in A$ chosen by a fixed rule, such as the centre of a box, the unique minimizer of $\E[-\log\mu(g(A)\mid\xp)]$ is the law $\nu_g(y\mid\xp)=\sum_{A:\,g(A)=y}r(A\mid\xp)$ of that point. It differs from $\ps(\cdot\mid\xp)$ whenever $g(A)$ can fall where $\ps$ is $0$, and then its most probable answer can be one that is never right, whatever the capacity of the model.
\end{enumerate}
Part (b) holds with the model's input $x=\rho(\xp)$ in place of $\xp$, by \cref{lem:th:lin} with $\mathcal H=\sigma(x)$. Part (a) does too when $\kappa$ also satisfies \cref{th:car} given $x$, for example when $\kappa_A(\xp)$ depends on $\xp$ only through $x$.
\end{proposition}}
\begin{proof}
(a) Let $\mathcal O$ be the sets with $r(A\mid\xp)>0$. Under \cref{th:car}, $\kappa(\cdot\mid\xp,y)$ is a distribution over the sets that contain $y$, with terms $\kappa_A(\xp)$, so $\sum_{A\ni y}\kappa_A(\xp)=1$ for every $y$. Let $q_\mu(A):=\kappa_A(\xp)\mu(A\mid\xp)$ be the probability of the label $A$ if the answer were drawn from $\mu$. Regrouping by $y$ gives $\sum_{A\in\mathcal O}q_\mu(A)=\sum_y\mu(y\mid\xp)\sum_{A\in\mathcal O,\,A\ni y}\kappa_A(\xp)\le1$. Since $-\log\mu(A\mid\xp)=-\log q_\mu(A)+\log\kappa_A(\xp)$ and the second term does not depend on $\mu$, the risk equals the cross-entropy from the label law $r$ to $q_\mu$ plus a constant. Gibbs' inequality also holds when $q\ge0$ and $\sum q\le1$: $-\sum r\log q\ge-\sum r\log r$, with equality only at $q=r$. The risk is therefore smallest exactly when $q_\mu=r(\cdot\mid\xp)$, that is when $\mu(A\mid\xp)=\ps(A\mid\xp)$ on every set in $\mathcal O$, and $\mu=\ps$ satisfies this.

(b) Averaging the one-hot target $\delta_{g(A)}$ over $A\sim r(\cdot\mid\xp)$ gives $\nu_g$, so \cref{lem:th:lin} gives the unique minimizer. For the second claim, let the right answer be uniform on a ring-shaped object whose box is the only label set: then $\nu_g=\delta_{g(A)}$ sits on the box centre, where $\ps$ puts no mass.
\end{proof}

\paragraph{Scope.} Proposition~1 covers boxes and gold options, not annotated points. It fixes the mass of every set the labels report and leaves open how mass is spread inside a set that no other label splits. The label set of an annotated point is a window centred on that point, so the window reveals where the answer lies inside it and \cref{th:car} fails. We treat the window as a tolerance and also keep the point's own cross-entropy, which by \cref{lem:th:dec} equals the window's set loss plus $\KL(\delta_{y_0}\|\mu(\cdot\mid A))$, so it pulls the mass inside the window toward the annotated point $y_0$.

\paragraph{Products of the two axes.} The read-out gives $\mu=p_x\otimes p_y$ from two questions on the same prefix. For a rectangle $A=A_x\times A_y$, $-\log\mu(A)=-\log p_x(A_x)-\log p_y(A_y)$, so the risk splits into one set loss per axis, and (a) applies to each axis when each axis label satisfies \cref{th:car}. The mass on $A$ is $p_x(A_x)\,p_y(A_y)$, which equals $\ps(A\mid x)$ only when the two coordinates are independent given the input. The self-check label is computed from this product. Against a joint target $\mu_T$ with marginals $\mu^x_T,\mu^y_T$, the product loses exactly the mutual information $I_T$ of the two coordinates under $\mu_T$:
\begin{equation}
\KL(\mu_T\|p_x\otimes p_y)=\KL(\mu^x_T\|p_x)+\KL(\mu^y_T\|p_y)+I_T.
\label{eq:product}
\end{equation}
The product puts mass on crossed cells, such as $(200,800)$ for two cups at $(200,200)$ and $(800,800)$. We count a missed point as lying on a crossed cell when it is off the target mask while the mask has pixels in the point's column and in its row. Among the misses of the 2B and 4B models, the point lies on such a crossed cell in \vRvCrossOursQm\% and \vRvCrossOursQl\% for ours and \vRvCrossPtsQm\% and \vRvCrossPtsQl\% for the point-writing model, which writes the two coordinates together. Over all images these are \vRvCrossAllOursQm\% and \vRvCrossAllOursQl\% for ours against \vRvCrossAllPtsQm\% and \vRvCrossAllPtsQl\%, so reading the axes apart does not put the point on a crossed cell more often. This count also includes misses inside one concave object, so it bounds the crossings between objects from above.
\section{Data, Licenses and Decontamination}
\label{app:data}

\paragraph{Label sets.} The label set of a record is the set of points its label accepts. For a box record it is the box itself. For an annotated point it is a window of a hundred values per axis, the ten bands of ten values from five bands below the point's band to four above it, clipped to the image; a record with several annotated points $(x_k,y_k)$ takes the union of their two-dimensional windows, $A=\bigcup_k W(x_k)\times W(y_k)$, whose mass is $\mu(A)=\sum_{(i,j)\in A}p_x(i)\,p_y(j)$ and not the product of two per-axis unions. The cross-entropy of the annotated point, $-\log p_x(x_1)-\log p_y(y_1)$, is added with weight one, for the first point when a record has several. The label set of a box edge is the interval of $\pm$\vSetEdgeTol\% of the box side around the annotated edge, and at least $\pm$\vSetEdgeMin{} values.

\paragraph{Training data.} The pointing pool has \vfMixSize{} records from \vfMixSources{} public sources: points (\vfMixPointFrac), boxes (\vfMixBoxFrac) and GUI elements (\vfMixGuiFrac). The points come from PixMo-Points and PixMo-Count~\cite{molmo} and from the free-space and object-reference parts of RoboPoint~\cite{robopoint}, the boxes from the RefCOCO, RefCOCO+ and RefCOCOg train splits and Visual Genome regions, and the GUI elements from the web and desktop parts of ShowUI, RICO widget captions and Wave-UI. The multiple-choice pool holds public multiple-choice and yes/no questions and questions generated by rules from COCO, RefCOCO and RICO. Of its \vfTypedMixN{} questions, \vfAgentN{} are agent-style questions on COCO images whose wording mentions no drone, robot or benchmark: whether the target is in the image, whether the camera has reached it (by the size of its RefCOCO box) and which way to turn. Neither pool has drone or navigation images, and their only robot images are those of the two RoboPoint parts. Every model trains on the same subset with the settings of \cref{tab:hparams}. Training takes \vTrHoursS, \vTrHoursM{} and \vTrHoursL{} hours on \vTrGpus{} A100 GPUs at 0.8B, 2B and 4B, for \vTrExamples{} examples and one LoRA adapter.

\paragraph{Budgets of the released models.} The released models we compare with were trained with very different budgets: GUI-Decisions-2B trains a LoRA on \vOtGuidEx{} examples in about \vOtGuidMin{} minutes on one GPU~\cite{guidecisions}, Intern-Decision fine-tunes the language backbone on \vOtInternEx{} samples~\cite{interndecision}, MJev-Spatial trains the full language backbone on \vOtMjevQa{} question-answer pairs~\cite{jevspatial}, and the pointing model MolmoPoint-8B trains for \vOtMolmoSteps{} steps of \vOtMolmoBatch{} examples~\cite{clark2026molmopoint}.

\begin{table}[!htbp]
\centering\footnotesize
\setlength{\tabcolsep}{3pt}
\begin{tabular}{@{}lc@{\hspace{8pt}}lc@{}}
\toprule
\multicolumn{4}{@{}l}{\emph{Shared by both models}} \\
Training steps & 6,000 & Pointing records & 288,000 \\
Learning rate & 0.0001 & Multiple-choice share & 0.5 \\
Warm-up steps & 100 & Max.\ pixels (train) & 1,003,520 \\
LoRA rank & 64 & Min.\ pixels (train) & 65,536 \\
LoRA $\alpha$ & 128 & Data seed & 0 \\
Examples per step & 96 & & \\
\midrule
Loss & \cellcolor{oursbg}ours & \multicolumn{2}{c@{}}{writes} \\
\midrule
Point target & \cellcolor{oursbg}set, $-\log\mu(A)$ & \multicolumn{2}{c@{}}{annotated point, CE} \\
Point CE on points & \cellcolor{oursbg}weight 1 & \multicolumn{2}{c@{}}{only loss} \\
Self-check weight & \cellcolor{oursbg}1 & \multicolumn{2}{c@{}}{0} \\
Box-edge weight & \cellcolor{oursbg}1 & \multicolumn{2}{c@{}}{0} \\
\bottomrule
\end{tabular}

\caption{\textbf{Training settings.} The upper block is shared by both models; the lower block gives the losses in which the decision model (ours) and the point-writing model (writes) differ. The \vAfPointRecords{} pointing records are a fixed subset of the \vfMixSize{} pool, drawn with the data seed, and each is seen once, since half of the \vAfStepExamples{} examples of each of the \vAfTrainSteps{} steps are pointing records. One LoRA adapter is trained on every projection of the language model in one stage from the base model, with the vision encoder frozen. Training images are resized to a pixel count drawn log-uniformly between the two limits, and the learning rate follows a cosine schedule.}
\label{tab:hparams}
\end{table}

\paragraph{Layout.} The instruction that introduces multiple-choice fields sits in a system message before the image, so a device computes it once and caches it across frames. The last option of a multiple-choice field is the bare word ``unknown''. Every pointing, self-check and agent question has a long wording (\cref{app:readout}) and a short one for the edge device, each used for half of the records. The short questions are ``x?'', ``y?'', ``left?'', ``top?'', ``right?'', ``bottom?'' and ``precise?''.

\paragraph{Licenses.} Images keep their upstream terms and are referenced by path, never redistributed. Six multiple-choice sources are non-commercial or research-only (DVQA, TQA, ScienceQA, DocVQA, InfographicVQA, MAmmoTH-VL). We will release code and models as far as the licenses allow, and also a model trained without the restricted sources, together with its results.

\paragraph{Decontamination.} Training images were matched against all evaluation images by file hash, perceptual hash and COCO identifier, which removed \vfLeakImages{} images (\vfLeakPB{} of them Point-Bench images found in PixMo). RefCOCO/+/g training records whose image appears in any of their validation or test splits are dropped.

\paragraph{The drone models.}\label{app:drone} No 0.8B DroneCATS model is released, so we train Qwen3.5-0.8B with the DroneCATS recipe~\cite[App.~C]{dronecats}, a full fine-tune for \vSetDcSteps{} steps of \vSetDcBatch{} examples on the single-image part of MAmmoTH-VL~\cite{mammothvl} and the benchmark's DroneCATS-Pilot-54K (CC BY-NC 4.0). We mix the two in equal parts, which gives this baseline a larger share of drone frames than the \vSetDcPaperMammoth\% MAmmoTH-VL mix of the DroneCATS paper. At 2B we use the released DroneCATS-2B. Each of our drone decision models starts from the DroneCATS model of its size, which writes its action as JSON, and is trained with our recipe and the settings of \cref{tab:hparams} on the training data above plus Pilot-54K. Pilot-54K is rewritten as answers to the visibility and arrival questions of our agent: the action ``finished'' is labelled reached, ``go'' visible but not reached, and a rotation neither visible nor reached, since the benchmark rotates only when the target is out of view.

\paragraph{Leakage check.} No DroneCATS benchmark frame or instruction appears in Pilot-54K. The \vsDcPilotRows{} Pilot rows (\vsDcPilotImgs{} images) come from NYU v2, SUN RGB-D and three TartanAir environments, and the benchmark runs in two other simulator maps. The smallest pHash distance between \vsDcStates{} benchmark frames and all Pilot images is \vsDcPhashMin{} of 64 bits, and none of the \vsDcCmds{} benchmark instructions matches a Pilot instruction or target description.
\section{Evaluation Protocols}
\label{app:protocol}

\paragraph{Models and families.}\label{app:families} A family is a release line with one architecture, language-model layers and vision encoder together, so Qwen3.5 and Qwen3-VL are separate families. Gemma-4-E4B has 4B effective parameters, so we group it with the 4B models.

\paragraph{Pointing.} We use Point-Bench (\vNnPB{} images), ScreenSpot-v2 (\vNnSS), Where2Place (\vNnWP), RoboRefIt (\vNnRR) and RefCOCO val (UNC; \vNnRC{} of its \vSetRcValN{} expressions, drawn with a fixed seed so that RefCOCO does not outweigh the others), each resized to the pixel area of \vfEdgeRes{} with its aspect ratio kept. The most probable point is right when it falls inside the element box (ScreenSpot-v2), the target box (RoboRefIt) or the mask (the others), and a $10{\times}10$ decision is right when its cell meets that region. A random $10{\times}10$ cell is right on \vWtRand\% of images.

\paragraph{Boxes and looking closer.} Boxes are tested on the eight RefCOCO/+/g splits at up to \vSetOneMP{} pixels, taking the most probable value of each box-edge branch and scoring Acc@0.5, the share of boxes with IoU${\ge}0.5$. To look closer, an image is read again at up to \vSetOneMP{} pixels where the model says no ($P(\text{yes})<0.5$), and the second answer replaces the first.

\paragraph{Multiple choice.} Multiple-choice accuracy is chance-corrected, read from the option codes at \vfEdgeRes{} and averaged over four cyclic orders of the options, with items that overlap the training data dropped. We report the unweighted mean over the \vMmNPair{} benchmarks of our suite on which both models of a pair have a result, so every benchmark counts equally. The suite is ImajevBench, JevBench, MMStar, CharXiv, InfographicVQA, Mind2Web, KIE (CORD and FUNSD), Hateful Memes, HallusionBench, MME, SEED-Bench, MMBench, AI2D, ScienceQA, POPE, CV-Bench, BLINK, RealWorldQA, R-Bench-M and MMMU-Pro. ImajevBench is the image decision benchmark released with imajev~\cite{imajev}, of which we use the items with public answers, and JevBench is a text-only decision benchmark asked in the same option layout; both are scored with the option codes as above.

\paragraph{Released decision models.} Models without a pointing interface get the $10{\times}10$ decision as two multiple-choice questions in their own format: ``The image is divided into 10 equal columns, numbered 1 (leftmost) to 10 (rightmost). Which column contains \{target\}?'' with the options 1 to 10, and the same for rows. Each model answers through its own endpoint, or else through its language-model head. GUI-Decisions-2B and MJev-Spatial-4B point, so we score the cell of their point and rank it by its mass (for MJev-Spatial, the product of its three $3{\times}3$ cell probabilities). GUI-Decisions-2B is read as its release reads it: one forward pass over its answer template, whose $x$ and $y$ slots each give probabilities over 256 single-token labels, one per bin of the axis (its \texttt{slotread\_config.json}).

\paragraph{Latency.} Every time starts from the original image and is taken end to end at the client, as the median over 200 Point-Bench images after five warm-up images, on one A100 (80\,GB) in bf16 at batch one. For rows on our engine the clock starts with the decoded image on the host and stops when the answer is on the host, so it covers the copy to the GPU, the processor's resize (run on the GPU and bit-identical to the processor's own CPU resize), rescaling, normalization, tokenization, the vision tower and every decoder pass. For vLLM rows the clock is the whole HTTP request carrying the original image, which vLLM decodes, resizes and preprocesses on its own CPU workers. For Qwen3.5, the point-writing model writes greedily with its own training prompt, up to \vSetMaxTok{} tokens, in vLLM \vfVllmVer{}, and we read with our read-out (\cref{app:latency}). Qwen3-VL reads in our CUDA-graph one-pass engine (bit-exact to the eager read) and writes in vLLM. GroundingPI also writes in vLLM, through its Transformers backend with the model's own serving code, and the other families write and read in the plain HF engine. The released decision models run on CUDA graphs, which leave every choice they make unchanged.

\paragraph{Generating rows of \cref{tab:dec}.} Generators stop at their first point and are scored on the point they write and on the cell that holds it. A model that must write its whole answer before any grid can be read has the same time in all three grid columns, while the padded digits of Qwen3.5 can stop once the cell is fixed. In structured decoding the format of the answer is given and only its six digits are written; these rows time the same model on our CUDA-graph engine and keep the accuracy of its generated point. Trained rows of \cref{tab:dec} are single runs.

\paragraph{MiniWoB++.} On MiniWoB++ we use the 15 tasks that a single click solves, with 50 seeds each and one click per episode. The model gets the target phrase of the instruction (``Click on the `No' button.'' becomes ``the `No' button''), because, when given the instruction as printed, the larger models point at the printed text itself. With the printed instruction, the 9B point-writing model clicks correctly in \vNxSFTMwNat\% of the episodes instead of \vNxSFTMw\%. A re-read crops the page to the bounding box of the $10{\times}10$ cells in the model's 90\% region (\cref{app:conformal}), widened by one cell on every side and to at least \vAfCropMinSide\% of each side of the page, or around the most probable cell when the region is empty. The model reads the crop once at its own pixels, and the click is the most probable point of that read, mapped back to the page. The threshold of the region is the validation threshold of \cref{app:conformal} for ours, and for the other models it was set on ScreenSpot-v2.

\paragraph{Multimodal-Mind2Web.} Ours keeps the most tasks safe at every size when it acts on its most confident 50\% of the steps, as it does at the 70\% cut of \cref{tab:webbody}. A task is safe when none of the steps acted on is wrong, and at 2B ours keeps \vWcOursMLo\% of tasks safe at the 50\% cut, against at most \vWcOthMLo\% for the other models. On Multimodal-Mind2Web~\cite{deng2023mind2web,zheng2024seeact} we follow the SeeAct-V protocol that UGround~\cite{gou2025uground} uses for small grounding models. The model gets the $1280{\times}1000$ page block holding the element and a description from the ground-truth step.

\paragraph{DroneCATS.} We run the \vDrEpis{} episodes of~\cite{dronecats} in real time (300\,s per episode, frames at $640{\times}352$), with all rows run in the same four-GPU L40S job. The models that write JSON act in the benchmark's own agent; in the main text they run without its THINK action, and \cref{tab:droneruns} also gives their results with it. For each frame, our agent sends the benchmark's system prompt, the frame and three questions, ``Is the \{target\} in the image?'', ``Has the camera reached the \{target\}?'' and the point, and one rule, the same for every decision model, acts on the answers:
\begin{enumerate}[label=(\arabic*),leftmargin=*,itemsep=0pt,topsep=2pt]
\item If $P(\text{visible})<0.5$, the drone turns $90^\circ$ toward the half of the image where the $x$ answer puts more mass; after four turns in a row it flies forward instead.
\item Otherwise, if $P(\text{reached})\ge0.5$, it declares arrival, or flies toward the target point if the previous step was already a declaration.
\item Otherwise it flies toward the most probable $10{\times}10$ cell with the benchmark's farthest move (``5\,m or more'', which the simulator caps at 3\,m outdoors and 1.5\,m indoors).
\end{enumerate}
A declaration counts as reached, and ends the episode, only when the drone is inside the success radius and the target is visible. A move continues until the next decision arrives, for at most 2.5\,s, and a turn runs at 60$^\circ$/s. The Pi columns delay each answer by the model's Pi time (\cref{app:edge}).
\section{Speed of the Decision}
\label{app:serving}

\begin{table}[!htbp]
\centering\footnotesize
\setlength{\tabcolsep}{2.1pt}
\begin{tabular}{@{}l>{\columncolor{oursbg}}c>{\columncolor{oursbg}}cccc@{}}
\toprule
& \multicolumn{2}{c}{ours reads (ms)\,$\downarrow$} & \multicolumn{2}{c}{writes (ms)\,$\downarrow$} & \\
\cmidrule(lr){2-3}\cmidrule(lr){4-5}
Model & $10{\times}10$ & $1000{\times}1000$ & \stwo{whole}{answer} & \stwo{structured}{decoding} & faster\,$\uparrow$ \\
\midrule
\multicolumn{6}{@{}l}{\emph{Qwen3.5 at $384^2$, generated in vLLM}} \\
0.8B & \vWfsReadC & \vOssThou & \vAuStopS & \vWfsTpl & \vSpStopsRatioThou$\times$ \\
2B & \vWfmReadC & \vOsmThou & \vAuStopM & \vWfmTpl & \vSpStopmRatioThou$\times$ \\
4B & \vWflReadC & \vOslThou & \vAuStopL & \vWflTpl & \vSpStoplRatioThou$\times$ \\
9B & \vWfxReadC & \vOsxThou & \vWfxVl & \vWfxTpl & \vOsxVlRatioThou$\times$ \\
\multicolumn{6}{@{}l}{\emph{Qwen3-VL at $384^2$, our CUDA-graph read, vLLM generation}} \\
2B & \vVgQvlTwoCell & \vVgQvlTwoThou & \vUtQvlTwoPtsVl & \vFdQvlTwoTplThou & \vVgQvlTwoFasterVl$\times$ \\
4B & \vVgQvlFourCell & \vVgQvlFourThou & \vUtQvlFourPtsVl & \vFdQvlFourTplThou & \vVgQvlFourFasterVl$\times$ \\
\multicolumn{6}{@{}l}{\emph{Other families at $384^2$, HF engine}} \\
InternVL3.5-1B & \vOsHivlACell & \vOsHivlAThou & \vLHivlAGen & -- & \vOsHivlARatioThou$\times$ \\
InternVL3.5-2B & \vOsHivlCell & \vOsHivlThou & \vLHivlGen & -- & \vOsHivlRatioThou$\times$ \\
InternVL3.5-4B & \vOsHivlDCell & \vOsHivlDThou & \vLHivlDGen & -- & \vOsHivlDRatioThou$\times$ \\
Gemma-4-E4B & \vLHgemCell & \vLHgemFull$^\ast$ & \vLHgemGen & -- & \vLHgemRatioF$\times$ \\
LFM2.5-VL-3B & \vLHlfmCell & \vLHlfmFull & \vLHlfmGen & -- & \vLHlfmRatioF$\times$ \\
LFM3-VL-2B & \vLHlfvCell & \vLHlfvFull & \vLHlfvGen & -- & \vLHlfvRatioF$\times$ \\
\multicolumn{6}{@{}l}{\emph{Qwen3.5 at 1\,MP, generated in vLLM}} \\
0.8B & \vWfsHiReadC & \vOssHiThou & \vWfsHiVl & \vWfsHiTpl & \vOssHiVlRatioThou$\times$ \\
2B & \vWfmHiReadC & \vOsmHiThou & \vWfmHiVl & \vWfmHiTpl & \vOsmHiVlRatioThou$\times$ \\
4B & \vWflHiReadC & \vOslHiThou & \vWflHiVl & \vWflHiTpl & \vOslHiVlRatioThou$\times$ \\
9B & \vWfxHiReadC & \vOsxHiThou & \vWfxHiVl & \vWfxHiTpl & \vOsxHiVlRatioThou$\times$ \\
\bottomrule
\end{tabular}

\caption{\textbf{Time per decision of every model, read by the decision model (ours) and written by the point-writing model (writes).} Cells are median ms per decision on one A100 at batch one (\cref{app:protocol}). Qwen3.5 and Qwen3-VL write in vLLM \vfVllmVer{}, stopping at the closing brace up to 4B at $384^2$ and writing the whole answer otherwise, and the other families write in HF after a given \texttt{\{"x":}. On our engine the point-writing model can also use structured decoding, in which the format is given and only its six digits are written. \emph{Faster} is the whole answer over our full map, as in \cref{tab:decb}; it divides times from one session, so it can differ in the last digit from the quotient of the printed medians. For the other families except Gemma-4-E4B the $10{\times}10$ read and the full map are the same read, timed twice in one process, and structured decoding exists only on our engine (--). For InternVL3.5 the two printed times differ by less than 1\,ms, within the spread of two timing runs of the same read at 1B and 2B. Gemma-4-E4B ($^\ast$) is read digit by digit, since its one-pass read does not match the stepped read exactly.}
\label{tab:speed}
\end{table}

\paragraph{Writing on our engine.}\label{app:engine} On our own engine the read stays \vNwFreeOverMapMin{} to \vNwFreeOverMapMax{} times faster than writing. With our engine and timing protocol, the point-writing model writes its whole answer in \vWfsFree, \vWfmFree{} and \vWflFree\,ms at 0.8B, 2B and 4B, against \vOssThou, \vOsmThou{} and \vOslThou\,ms for our full map. Off-the-shelf Qwen3.5 writes its native answer on our engine in \vNwGraphs, \vNwGraphm{} and \vNwGraphl\,ms, against \vUtVls, \vUtVlm{} and \vUtVll\,ms in vLLM and \vUtFastsThou, \vUtFastmThou{} and \vUtFastlThou\,ms for our two-pass read. Against the faster of the two engines at each size, the read is \vNwBestOverReadMin{} to \vNwBestOverReadMax{} times faster. The generating rows up to 4B at $384^2$ stop at the closing brace of the point, and the 9B and 1\,MP rows write their whole answer, which takes about as long (\vWflVl{} against \vAuStopL\,ms at 4B).

\begin{table}[!htbp]
\centering\footnotesize
\setlength{\tabcolsep}{4pt}
\begin{tabular}{@{}llccc@{\hspace{8pt}}cc@{}}
\toprule
& & \multicolumn{3}{c}{ours reads (ms)\,$\downarrow$} & \multicolumn{2}{c@{}}{writes (ms)\,$\downarrow$} \\
\cmidrule(lr){3-5}\cmidrule(l){6-7}
GPU & size & $10^2$ & $100^2$ & $1000^2$ & structured & whole answer \\
\midrule
A100 & 2B & \vDvAMCell & \vDvAMHund & \vDvAMFull & \vDvAMTpl & \vDvAMFree \\
 & 0.8B & \vDvASCell & \vDvASHund & \vDvASFull & \vDvASTpl & \vDvASFree \\
\addlinespace[2pt]
L40S & 2B & \vDvLsMCell & \vDvLsMHund & \vDvLsMFull & \vDvLsMTpl & \vDvLsMFree \\
 & 0.8B & \vDvLsSCell & \vDvLsSHund & \vDvLsSFull & \vDvLsSTpl & \vDvLsSFree \\
\addlinespace[2pt]
L4 & 2B & \vDvLfMCell & \vDvLfMHund & \vDvLfMFull & \vDvLfMTpl & \vDvLfMFree \\
 & 0.8B & \vDvLfSCell & \vDvLfSHund & \vDvLfSFull & \vDvLfSTpl & \vDvLfSFree \\
\bottomrule
\end{tabular}

\caption{\textbf{Read and write times of Qwen3.5 on three GPUs.} Median ms per decision of our read and of the point-writing model (writes), with structured decoding (six digits) or its whole answer, all on our engine and so not the vLLM times of \cref{tab:speed}.}
\label{tab:devices}
\end{table}

\paragraph{Other GPUs.}\label{app:devices} \Cref{tab:devices} repeats the timings on an L40S and an L4. Up to $100{\times}100$ the read is \vDvTplOverHundMin{} to \vDvTplOverHundMax{} times faster than writing the six structured digits on every GPU and at both sizes, and the $100{\times}100$ read costs only a few percent more than the $10{\times}10$ read. The full map is \vDvFreeOverFullMin{} to \vDvFreeOverFullMax{} times faster than writing the whole answer and \vDvTplOverFullMin{} to \vDvTplOverFullMax{} times faster than structured decoding, on every GPU and at both sizes. On a small device the read stops at the grid the task needs, as on the Pi below, where the full map takes \vPqThouS\,s and the $100{\times}100$ grid \vPqHundS\,s.

\subsection{On a Raspberry Pi 5}
\label{app:edge}

\emph{Click decision.} We merge the LoRA of our 0.8B model and convert it to Q8\_0 weights in llama.cpp, which has no tree attention mask, so each branch of the click decision gets its own sequence. The click decision (the two axis branches and the self-check, read at their first answer token) takes a median of \vPqClickS\,s for a new image of \vPcImgTok{} image tokens and \vPqClickWarmS\,s for another question on the same image, over \vPcNClick{} RefCOCO and ScreenSpot-v2 images without thermal throttling, in the session that also timed the finer maps below. On \vPcNAll{} test records (RefCOCO, ScreenSpot-v2 and Point-Bench, \vSetPcPerBench{} each), the Q8\_0 and full-precision weights both reach \vPcAccQeAll\% cell accuracy and pick the same cell on \vPcSameAll\%. For the finer maps each one- and two-digit prefix is one more token that reads the recurrent state of its parent in place, with no sequence and no state of its own, so one \texttt{llama\_decode} call computes every prefix of the full map within \vPfPeakGb\,GB of memory. With the image already processed, the read takes \vPfHundWarmNew\,s for $100{\times}100$ and \vPfThouWarmNew\,s for the full map, and \vPqHundS{} and \vPqThouS\,s on a new image. With a copied sequence per prefix the 8\,GB Pi holds \vSetPiSeqs{} sequences, the full map takes two calls, and the same reads take \vPfHundWarmRef{} and \vPfThouWarmRef\,s; decoded in the same calls, the two ways of reading give identical logits.

\emph{Drone models.} We time the four models of \cref{tab:dronebody} on the same Pi (8\,GB, four CPU threads, Q8\_0 weights with a 16-bit vision projector) over \vSetPiFramesN{} agent frames (\cref{tab:piinputs}). We ported DroneCATS-2B's architecture to llama.cpp and checked that its greedy JSON actions match those of HF transformers on these frames. At the Pi setting each model was timed in three runs on different days, with no thermal throttling; the medians differ by at most \vPiSpreadMax\% between runs, and the parse rate and the agreement with the server CPU's actions are the same in every run.

\emph{Pi times in the simulator.} The Pi does not run the simulator. In the Pi columns of \cref{tab:dronebody}, each answer is held back on the L40S until the model's median Pi time at $256{\times}128$ in the first run has passed since the request, and the model sees the Pi's $256{\times}128$-pixel frame.

\begin{table}[!htbp]
\centering\footnotesize
\setlength{\tabcolsep}{1.5pt}
\begin{tabular}{@{}lcc@{}}
\toprule
 & \stwo{Pi setting,}{short questions} & \stwo{frame and long questions}{of the L40S runs} \\
frame (pixels) & $256{\times}128$ & $640{\times}352$ \\
\midrule
\multicolumn{3}{@{}l}{\emph{Qwen3.5-0.8B, trained as in DroneCATS~\cite{dronecats}: s per decision\,$\downarrow$}} \\
\quad writes JSON & \vPiMeanQJ\,$\pm$\,\vPiSdQJ & \vPiInQJNat \\
\oursrow\quad + our recipe, decides & \vPiMeanQD\,$\pm$\,\vPiSdQD & \vPiInQDNat \\
\multicolumn{3}{@{}l}{\emph{DroneCATS-2B, released: s per decision\,$\downarrow$}} \\
\quad writes JSON & \vPiMeanCJ\,$\pm$\,\vPiSdCJ & \vPiInCJNat \\
\oursrow\quad + our recipe, decides & \vPiMeanCD\,$\pm$\,\vPiSdCD & \vPiInCDNat \\
\multicolumn{3}{@{}l}{\emph{Image tokens, 0.8B / DroneCATS-2B}} \\
\quad per frame & \vPiImgQBase\,/\,\vPiImgCBase & \vPiImgQNat\,/\,\vPiImgCNat \\
\bottomrule
\end{tabular}

\caption{\textbf{Decision time of the four drone models on the Raspberry Pi 5.} Warm median seconds per agent step over \vSetPiFramesN{} DroneCATS frames, at the Pi setting of \cref{tab:dronebody} (mean $\pm$ standard deviation over three runs on different days) and at the L40S runs' frame and question wording (one run), with the image tokens per frame in the last row.}
\label{tab:piinputs}
\end{table}

\subsection{Serving with prefix-state reuse}
\label{app:latency}

\paragraph{Why our own engine.} For Qwen3.5, vLLM \vfVllmVer{} reuses a cached prefix only in blocks of \vSetVllmBlock{} tokens, longer than our shared prefix of image and instruction, so every question would recompute the image. Our read-out computes the prefix once and reads all question tails and digit prefixes in one more forward pass under the tree mask of \cref{app:readout}, replayed as a CUDA graph.

\paragraph{Agreement with the reported evaluation.} The timed engine differs from the reported evaluation only in floating-point arithmetic. In the full-map read a second-digit prefix takes its recurrent state from its parent without a copy, the same step in a different floating-point order. The engine also accumulates the output logits in fp32, while the reported evaluation computes them in bf16, which leaves exact ties among the 1,000 values on a third to two fifths of the axes for the point-writing model and for ours alike. On the \vBcNAll{} test images at $384^2$ this engine changes no $10{\times}10$ or $100{\times}100$ cell up to 4B and moves point accuracy by at most \vFzMaxShift{} pp against the engine with the copy. At 9B and at 1\,MP it changes at most one $100{\times}100$ cell and moves point accuracy by at most \vFzHiMaxShift{} pp. Its point accuracy is \vFzAccS, \vFzAccM{} and \vFzAccL\% at 0.8B, 2B and 4B, against the reported \vFzRepS, \vFzRepM{} and \vFzRepL\%.

\paragraph{Other engines and families.} On \vDNGraphNineN{} stored drone frames, our engine and vLLM, reading the same Qwen3.5 decision model of ours, take the same decision on \vDNGraphTwoSame{} frames with the 2B model and on \vDNGraphNineSame{} with the 9B model, about as often as two vLLM runs agree with each other. Qwen3-VL has no recurrent layers, so the same read runs as two CUDA graphs, the prefill with its vision tower and one branch forward. It matches the eager one-pass read exactly, and the stored reads to at most \vVgQvlTwoStMaxDp{} per value at 2B and \vVgQvlFourStMaxDp{} at 4B, the level of bf16 noise.
\section{Where the Gain Comes From}
\label{app:centres}

\begin{table*}[!tp]
\centering\footnotesize
\setlength{\tabcolsep}{2.5pt}
\begin{tabular}{@{}l*{5}{c>{\columncolor{oursbg}}c}@{\hspace{7pt}}cc>{\columncolor{oursbg}}cc@{\hspace{7pt}}c>{\columncolor{oursbg}}c@{}}
\toprule
& \multicolumn{10}{c}{point acc. (\%)\,$\uparrow$ per benchmark} & \multicolumn{4}{c}{point acc. (\%)\,$\uparrow$, all 5,354 images} & \multicolumn{2}{c}{\stwo{multiple choice}{(\%)\,$\uparrow$}} \\
& \multicolumn{2}{c}{Point-Bench} & \multicolumn{2}{c}{ScreenSpot-v2} & \multicolumn{2}{c}{Where2Place} & \multicolumn{2}{c}{RoboRefIt} & \multicolumn{2}{c}{RefCOCO} & \multicolumn{4}{c}{} & \multicolumn{2}{c}{} \\
\cmidrule(lr){2-3}\cmidrule(lr){4-5}\cmidrule(lr){6-7}\cmidrule(lr){8-9}\cmidrule(lr){10-11}
Model & writes & ours & writes & ours & writes & ours & writes & ours & writes & ours & writes & \stwo{sets}{only} & ours & \stwo{$\Delta$ ours $-$}{writes (95\% CI)} & writes & ours \\
\midrule
\multicolumn{17}{@{}l}{\emph{Qwen3.5}} \\
\quad Qwen3.5-0.8B & \vNsSFTPB & \vNsFinPB & \vNsSFTSS & \vNsFinSS & \vNsSFTWP & \vNsFinWP & \vNsSFTRR & \vNsFinRR & \vNsSFTRC & \vNsFinRC & \vNsSFTPt & \textbf{\vNsSetsPt} & \vNsFinPt & \vNCisFinSFT\ [\vNCisFinSFTLo, \vNCisFinSFTHi] & \vMsSFTPair & \vMsFinPair \\
\quad Qwen3.5-2B & \vNmSFTPB & \vNmFinPB & \vNmSFTSS & \vNmFinSS & \vNmSFTWP & \vNmFinWP & \vNmSFTRR & \vNmFinRR & \vNmSFTRC & \vNmFinRC & \vNmSFTPt & \vNmSetsPt & \textbf{\vNmFinPt} & \vNCimFinSFT\ [\vNCimFinSFTLo, \vNCimFinSFTHi] & \vMmSFTPair & \vMmFinPair \\
\quad Qwen3.5-4B & \vNlSFTPB & \vNlFinPB & \vNlSFTSS & \vNlFinSS & \vNlSFTWP & \vNlFinWP & \vNlSFTRR & \vNlFinRR & \vNlSFTRC & \vNlFinRC & \vNlSFTPt & \vNlSetsPt & \textbf{\vNlFinPt} & \vNCilFinSFT\ [\vNCilFinSFTLo, \vNCilFinSFTHi] & \vMlSFTPair & \vMlFinPair \\
\quad Qwen3.5-9B & \vNxSFTPB & \vNxFinPB & \vNxSFTSS & \vNxFinSS & \vNxSFTWP & \vNxFinWP & \vNxSFTRR & \vNxFinRR & \vNxSFTRC & \vNxFinRC & \vNxSFTPt & \vNxSetsPt & \textbf{\vNxFinPt} & \vNCixFinSFT\ [\vNCixFinSFTLo, \vNCixFinSFTHi] & \vMxSFTPair & \vMxFinPair \\
\multicolumn{17}{@{}l}{\emph{InternVL3.5}} \\
\quad InternVL3.5-1B & \vNivlASFTPB & \vNivlAFinPB & \vNivlASFTSS & \vNivlAFinSS & \vNivlASFTWP & \vNivlAFinWP & \vNivlASFTRR & \vNivlAFinRR & \vNivlASFTRC & \vNivlAFinRC & \vNivlASFTPt & \textbf{\vSoSetsIvA} & \vNivlAFinPt & \vNCiivlAFinSFT\ [\vNCiivlAFinSFTLo, \vNCiivlAFinSFTHi] & \vMivlASFTPair & \vMivlAFinPair \\
\quad InternVL3.5-2B & \vNivlSFTPB & \vNivlFinPB & \vNivlSFTSS & \vNivlFinSS & \vNivlSFTWP & \vNivlFinWP & \vNivlSFTRR & \vNivlFinRR & \vNivlSFTRC & \vNivlFinRC & \vNivlSFTPt & \textbf{\vIsSetsPt} & \vNivlFinPt & \vNCiivlFinSFT\ [\vNCiivlFinSFTLo, \vNCiivlFinSFTHi] & \vMivlSFTPair & \vMivlFinPair \\
\quad InternVL3.5-4B & \vNivlDSFTPB & \vNivlDFinPB & \vNivlDSFTSS & \vNivlDFinSS & \vNivlDSFTWP & \vNivlDFinWP & \vNivlDSFTRR & \vNivlDFinRR & \vNivlDSFTRC & \vNivlDFinRC & \vNivlDSFTPt & \vSoSetsIvD & \textbf{\vNivlDFinPt} & \vNCiivlDFinSFT\ [\vNCiivlDFinSFTLo, \vNCiivlDFinSFTHi] & \vMivlDSFTPair & \vMivlDFinPair \\
\quad InternVL3.5-8B & \vNivlHSFTPB & \vNivlHFinPB & \vNivlHSFTSS & \vNivlHFinSS & \vNivlHSFTWP & \vNivlHFinWP & \vNivlHSFTRR & \vNivlHFinRR & \vNivlHSFTRC & \vNivlHFinRC & \vNivlHSFTPt & \textbf{\vSoSetsIvH} & \textbf{\vNivlHFinPt} & \vNCiivlHFinSFT\ [\vNCiivlHFinSFTLo, \vNCiivlHFinSFTHi] & \vMivlHSFTPair & \vMivlHFinPair \\
\multicolumn{17}{@{}l}{\emph{Qwen3-VL}} \\
\quad Qwen3-VL-2B & \vAfQvlTwoWrPB & \vAfQvlTwoOursPB & \vAfQvlTwoWrSS & \vAfQvlTwoOursSS & \vAfQvlTwoWrWP & \vAfQvlTwoOursWP & \vAfQvlTwoWrRR & \vAfQvlTwoOursRR & \vAfQvlTwoWrRC & \vAfQvlTwoOursRC & \vQvTwoPtsPt & \textbf{\vSoSetsQvTwo} & \vQvTwoOursPt & \vQvTwoD\ [\vQvTwoDLo, \vQvTwoDHi] & \vAfQvlTwoMcWr & \vAfQvlTwoMcOurs \\
\quad Qwen3-VL-4B & \vAfQvlFourWrPB & \vAfQvlFourOursPB & \vAfQvlFourWrSS & \vAfQvlFourOursSS & \vAfQvlFourWrWP & \vAfQvlFourOursWP & \vAfQvlFourWrRR & \vAfQvlFourOursRR & \vAfQvlFourWrRC & \vAfQvlFourOursRC & \vQvFourPtsPt & \textbf{\vSoSetsQvFour} & \vQvFourOursPt & \vQvFourD\ [\vQvFourDLo, \vQvFourDHi] & \vAfQvlFourMcWr & \vAfQvlFourMcOurs \\
\multicolumn{17}{@{}l}{\emph{Other families}} \\
\quad Gemma-4-E4B & \vNgemSFTPB & \vNgemFinPB & \vNgemSFTSS & \vNgemFinSS & \vNgemSFTWP & \vNgemFinWP & \vNgemSFTRR & \vNgemFinRR & \vNgemSFTRC & \vNgemFinRC & \vNgemSFTPt & \textbf{\vSoSetsGem} & \textbf{\vNgemFinPt} & \vNCigemFinSFT\ [\vNCigemFinSFTLo, \vNCigemFinSFTHi] & \vMgemSFTPair & \vMgemFinPair \\
\quad LFM2.5-VL-3B & \vNlfmSFTPB & \vNlfmFinPB & \vNlfmSFTSS & \vNlfmFinSS & \vNlfmSFTWP & \vNlfmFinWP & \vNlfmSFTRR & \vNlfmFinRR & \vNlfmSFTRC & \vNlfmFinRC & \vNlfmSFTPt & \vSoSetsLfm & \textbf{\vNlfmFinPt} & \vNCilfmFinSFT\ [\vNCilfmFinSFTLo, \vNCilfmFinSFTHi] & \vMlfmSFTPair & \vMlfmFinPair \\
\quad LFM3-VL-2B & \vNlfvSFTPB & \vNlfvFinPB & \vNlfvSFTSS & \vNlfvFinSS & \vNlfvSFTWP & \vNlfvFinWP & \vNlfvSFTRR & \vNlfvFinRR & \vNlfvSFTRC & \vNlfvFinRC & \vNlfvSFTPt & \vSoSetsLfv & \textbf{\vNlfvFinPt} & \vNCilfvFinSFT\ [\vNCilfvFinSFTLo, \vNCilfvFinSFTHi] & \vMlfvSFTPair & \vMlfvFinPair \\
\quad LLaVA-1.5-7B & \vNllvSFTPB & \vNllvFinPB & \vNllvSFTSS & \vNllvFinSS & \vNllvSFTWP & \vNllvFinWP & \vNllvSFTRR & \vNllvFinRR & \vNllvSFTRC & \vNllvFinRC & \vNllvSFTPt & \vSoSetsLlv & \textbf{\vNllvFinPt} & \vNCillvFinSFT\ [\vNCillvFinSFTLo, \vNCillvFinSFTHi] & \vMllvSFTPair & \vMllvFinPair \\
\quad Rex-Omni-3B & \vAfRexWrPB & \vAfRexOursPB & \vAfRexWrSS & \vAfRexOursSS & \vAfRexWrWP & \vAfRexOursWP & \vAfRexWrRR & \vAfRexOursRR & \vAfRexWrRC & \vAfRexOursRC & \vNdWrPt & \textbf{\vSoSetsRex} & \vNdOursPt & \vNdDfPt\ [\vNdDfPtLo, \vNdDfPtHi] & \vAfRexMcWr & \vAfRexMcOurs \\
\bottomrule
\end{tabular}

\caption{\textbf{Point accuracy per benchmark of the point-writing model (writes) and the decision model (ours), for every base model.} The left block gives each benchmark, the middle block all \vBcNAll{} images with $\Delta$ (ours minus writes, 95\% CI), and the right block chance-corrected multiple-choice accuracy on \vMmNPair{} benchmarks. The Qwen3.5 rows up to 4B are the first seed (\cref{tab:allrows}), so their $\Delta$ differs slightly from the seed means in Tab.~2. Sets only trains the box as a set, with no box-edge or self-check question, and bold is the best of the three. Ours is within \vSoFvSIvAcell{} to \vNCilFinSets{}~pp of sets only, so the edges and self-check cost no accuracy. Rex-Omni-3B is discussed in \cref{app:readout}.}
\label{tab:sets}
\end{table*}

\paragraph{Off-centre images.} On the \vBcNPool{} off-centre images of \cref{sec:box}, only our 2B model is right on \vQlOnlyOurs{} images and only the point-writing 2B on \vQlOnlyCentre{} (examples in \cref{fig:qual}). These are the RefCOCO, Point-Bench and Where2Place images (\vPoolRef, \vPoolPb{} and \vPoolWp) whose target mask does not contain the centre of the mask's bounding box. ScreenSpot-v2 and RoboRefIt are scored on boxes, whose centre always lies inside. A best guess counts as lying at the box centre when it is less than \vAtCentreTol\% of the image from that centre, with the image width and height each scaled to one.

\paragraph{Multiple choice.} The right block of \cref{tab:sets} checks that the point labels leave the model's other answers in place. Both models of a pair are trained on the same multiple-choice data, and the chance-corrected accuracy of ours stays within one point of the point-writing model for \vMcDiffWithinN{} of the \vMcDiffRows{} models and within \nosign{\vMcDiffLo}~pp for all of them.

\paragraph{Acting at the centre of a predicted box.}\label{app:boxcentre} Our point read is more accurate than the centre of our own predicted box, most of all where the object does not contain its box centre (\cref{tab:boxcentre}). At 2B the point read is right on \vBcTwoAllPt\% of all images against \vBcTwoAllBx\% for the box centre (\vBcTwoAllD{} pp, 95\% CI \vBcTwoAllDLo{} to \vBcTwoAllDHi), and on the \vBcNPool{} off-centre images on \vBcTwoPoolPt\% against \vBcTwoPoolBx\% (\vBcTwoPoolD{} pp, 95\% CI \vBcTwoPoolDLo{} to \vBcTwoPoolDHi).

\begin{table}[!htbp]
\centering\footnotesize
\setlength{\tabcolsep}{4pt}
\begin{tabular}{@{}lcccc@{}}
\toprule
& \multicolumn{2}{c}{all images} & \multicolumn{2}{c@{}}{\vBcNPool{} off-centre images} \\
\cmidrule(lr){2-3}\cmidrule(l){4-5}
Ours & box centre & point read & box centre & point read \\
\midrule
2B & \vBcTwoAllBx & \vBcTwoAllPt & \vBcTwoPoolBx & \vBcTwoPoolPt \\
\quad $\Delta$ (95\% CI) & \multicolumn{2}{c}{\vBcTwoAllD{} [\vBcTwoAllDLo, \vBcTwoAllDHi]} & \multicolumn{2}{c@{}}{\vBcTwoPoolD{} [\vBcTwoPoolDLo, \vBcTwoPoolDHi]} \\
4B & \vBcFourAllBx & \vBcFourAllPt & \vBcFourPoolBx & \vBcFourPoolPt \\
\bottomrule
\end{tabular}
\caption{\textbf{Our point is more accurate than the centre of our own predicted box, most of all on off-centre images.} Point accuracy (\%) of our Qwen3.5 models when the action is the centre of the most probable box edges or the most probable point; off-centre images are those whose annotated box centre misses the target, and $\Delta$ (2B) is point read minus box centre.}
\label{tab:boxcentre}
\end{table}

\paragraph{Point writing without box-derived labels.}\label{app:pointsnobox} Ours stays more accurate over all images even than a point-writing model trained without box-centre labels, which is no longer pulled toward the box centre and points at the object on off-centre images slightly more often than ours. We retrained the 2B point-writing model without every record whose point label is a box centre (\vPnbDroppedN{} records from RefCOCO/+/g, Visual Genome regions and GUI element boxes), which leaves \vPnbKeptN{} human-annotated points and the same multiple-choice pool and \vSetTrainSteps{} steps. On the \vPnbPoolN{} off-centre images the retrained model points at the object on \vPnbNbPool\%, against \vPnbPtPool\% for the point-writing model and \vPnbOuPool\% for ours. Its best guess lies at the box centre on \vPnbNbCen\% of these images, against \vPnbPtCen\% for the point-writing model and \vPnbOuCen\% for ours, the shares of \cref{sec:box}. Ours lies there \nosign{\vPnbDOuPtCen} pp less often than the point-writing model (95\% CI \nosign{\vPnbDOuPtCenHi} to \nosign{\vPnbDOuPtCenLo}) and about as often as the retrained model (\vPnbDOuNbCen{} pp, 95\% CI \vPnbDOuNbCenLo{} to \vPnbDOuNbCenHi). Over all \vBcNAll{} images it reaches \vPnbNbAcc\%, and ours leads it by \nosign{\vPnbDOuNbAcc}~pp (95\% CI \vPnbDOuNbAccLo{} to \vPnbDOuNbAccHi). Dropping these records also removes the GUI data, and on the \vRnNgN{} images of the four benchmarks without screens ours still leads by \nosign{\vRnNgD}~pp (95\% CI \vRnNgDLo{} to \vRnNgDHi).

\paragraph{The answer layout and the label.}\label{app:layout} The label sets account for most of the gain (\cref{tab:layout}): they add \nosign{\vLyLabPtM} to \nosign{\vLyLabPtS}~pp at every size, with every interval above zero, and the answer layout adds \nosign{\vLyLayPtL} to \nosign{\vLyLayPtS}~pp, within noise at 4B. A model trained on points with the two axes asked apart, on the same data and steps, separates the two effects: the layout is this model minus the point-writing one, and the label is the sets-only model minus this model.

\begin{table}[!htbp]
\centering\footnotesize
\setlength{\tabcolsep}{5pt}
\begin{tabular}{@{}lcc@{}}
\toprule
Qwen3.5 & layout gain, pp (95\% CI) & label-set gain, pp (95\% CI) \\
\midrule
0.8B & \vLyLayPtS{} [\vLyLayPtSLo, \vLyLayPtSHi] & \vLyLabPtS{} [\vLyLabPtSLo, \vLyLabPtSHi] \\
2B & \vLyLayPtM{} [\vLyLayPtMLo, \vLyLayPtMHi] & \vLyLabPtM{} [\vLyLabPtMLo, \vLyLabPtMHi] \\
4B & \vLyLayPtL{} [\vLyLayPtLLo, \vLyLayPtLHi] & \vLyLabPtL{} [\vLyLabPtLLo, \vLyLabPtLHi] \\
\bottomrule
\end{tabular}
\caption{\textbf{Separating the answer layout from the label.} Gain in point accuracy from asking each axis as its own question (layout) and from training on label sets instead of points (label), each measured between two runs that differ only in that choice. Each run is the first seed; over \vAsSetNWord{} seeds at 2B, the point-writing model, the layout alone and the label sets reach \vAsPtsPtMean, \vAsParPtMean{} and \vAsSetPtMean\% (Tab.~6).}
\label{tab:layout}
\end{table}

\paragraph{Qwen3-VL.}\label{app:qwen3vl} As decision models, Qwen3-VL-2B and -4B beat both the point-writing model and the off-the-shelf model (\cref{tab:qwen3vl}). We time them in \cref{tab:dec}. Fine-tuned to write its point, Qwen3-VL is pulled to the box centre, as part (b) of Proposition~1 predicts (\cref{app:sets}): it puts its best guess there on \vQvTwoPtsAtc\% (2B) and \vQvFourPtsAtc\% (4B) of the images whose box centre is not on the object, and so falls below the off-the-shelf model by \nosign{\vQvTwoDPtsBase} pp at 2B (95\% CI \vQvTwoDPtsBaseLo{} to \vQvTwoDPtsBaseHi) and by \nosign{\vQvFourDPtsBase} pp at 4B, where the interval includes zero (95\% CI \vQvFourDPtsBaseLo{} to \vQvFourDPtsBaseHi).

\begin{table}[!htbp]
\centering\footnotesize
\setlength{\tabcolsep}{0.8pt}
\begin{tabular}{@{}lcccccc@{}}
\toprule
& \stwo{point}{acc.\,$\uparrow$} & \stwo{cell}{acc.\,$\uparrow$} & \stwo{off-}{centre\,$\uparrow$} & \stwo{at box}{centre\,$\downarrow$} & \stwo{wrong}{@70\%\,$\downarrow$} & \stwo{}{AUROC\,$\uparrow$} \\
\midrule
\multicolumn{7}{@{}l}{\emph{Qwen3-VL-2B}} \\
o.t.s. & \vQvTwoBasePt & \vGcQvlTwoCell & -- & -- & \vQvTwoBaseWrong & \vQvTwoBaseAu \\
writes & \vQvTwoPtsPt & \vQvTwoPtsCell & \vQvTwoPtsPool & \vQvTwoPtsAtc & \vQvTwoPtsWrong & \vQvTwoPtsAu \\
\oursrow ours & \textbf{\vQvTwoOursPt} & \textbf{\vQvTwoOursCell} & \textbf{\vQvTwoOursPool} & \textbf{\vQvTwoOursAtc} & \textbf{\vQvTwoOursWrong} & \textbf{\vQvTwoOursAu} \\
\quad\scriptsize $\Delta$ & \scriptsize\vQvTwoD & \scriptsize\vQvTwoDCell & \scriptsize\vQvTwoDPool & -- & -- & -- \\
\quad\scriptsize 95\% CI & \scriptsize[\vQvTwoDLo, \vQvTwoDHi] & \scriptsize[\vQvTwoDCellLo, \vQvTwoDCellHi] & \scriptsize[\vQvTwoDPoolLo, \vQvTwoDPoolHi] & & & \\
\multicolumn{7}{@{}l}{\emph{Qwen3-VL-4B}} \\
o.t.s. & \vQvFourBasePt & \vGcQvlFourCell & -- & -- & \vQvFourBaseWrong & \vQvFourBaseAu \\
writes & \vQvFourPtsPt & \vQvFourPtsCell & \vQvFourPtsPool & \vQvFourPtsAtc & \vQvFourPtsWrong & \vQvFourPtsAu \\
\oursrow ours & \textbf{\vQvFourOursPt} & \textbf{\vQvFourOursCell} & \textbf{\vQvFourOursPool} & \textbf{\vQvFourOursAtc} & \textbf{\vQvFourOursWrong} & \textbf{\vQvFourOursAu} \\
\quad\scriptsize $\Delta$ & \scriptsize\vQvFourD & \scriptsize\vQvFourDCell & \scriptsize\vQvFourDPool & -- & -- & -- \\
\quad\scriptsize 95\% CI & \scriptsize[\vQvFourDLo, \vQvFourDHi] & \scriptsize[\vQvFourDCellLo, \vQvFourDCellHi] & \scriptsize[\vQvFourDPoolLo, \vQvFourDPoolHi] & & & \\
\bottomrule
\end{tabular}

\caption{\textbf{Qwen3-VL off-the-shelf (o.t.s.), fine-tuned to write its point (writes) and as the decision model (ours).} Off-centre is point accuracy on the \vQlPoolN{} images whose box centre misses the target, and at box centre is the share of those images where the model's best guess lies at the centre. Cell accuracy is that of the top $10{\times}10$ cell of the map, and for o.t.s.\ of the written point. The small rows give ours minus writes with 95\% CIs.}
\label{tab:qwen3vl}
\end{table}
\section{What in the Training Recipe Matters}
\label{app:recipe}
\label{app:selfteacher}
\label{sec:selfteacher}

\begin{table*}[!tp]
\centering\footnotesize
\setlength{\tabcolsep}{3pt}
\begin{tabular}{@{}l@{\hspace{7pt}}cccc@{\hspace{7pt}}c@{\hspace{7pt}}cc@{\hspace{7pt}}cc@{}}
\toprule
& \multicolumn{5}{c}{point acc. (\%)\,$\uparrow$} & \multicolumn{2}{c}{wrong@70\% (\%)\,$\downarrow$} & \multicolumn{2}{c}{AUROC\,$\uparrow$} \\
& \stwo{main}{run} & seed 1 & seed 2 & \stwo{no agent}{questions} & \stwo{$\Delta$ no agent $-$}{main (95\% CI)} & \stwo{main}{run} & \stwo{no agent}{questions} & \stwo{main}{run} & \stwo{no agent}{questions} \\
\midrule
\multicolumn{10}{@{}l}{\emph{Qwen3.5-0.8B}} \\
\quad writes & \vSnAccSPtsOrig & \vSnAccSPtsSeedA & -- & \vSnAccSPtsNa & \vSnNaSPtsAcc\ [\vSnNaSPtsAccLo, \vSnNaSPtsAccHi] & \vSnWrSPtsOrig & \vSnWrSPtsNa & \vSnAuSPtsOrig & \vSnAuSPtsNa \\
\oursrow\quad ours & \vSnAccSOursOrig & \vSnAccSOursSeedA & -- & \vSnAccSOursNa & \vSnNaSOursAcc\ [\vSnNaSOursAccLo, \vSnNaSOursAccHi] & \vSnWrSOursOrig & \vSnWrSOursNa & \vSnAuSOursOrig & \vSnAuSOursNa \\
\multicolumn{10}{@{}l}{\emph{Qwen3.5-2B}} \\
\quad writes & \vSnAccMPtsOrig & \vSnAccMPtsSeedA & \vSnAccMPtsSeedB & \vSnAccMPtsNa & \vSnNaMPtsAcc\ [\vSnNaMPtsAccLo, \vSnNaMPtsAccHi] & \vSnWrMPtsOrig & \vSnWrMPtsNa & \vSnAuMPtsOrig & \vSnAuMPtsNa \\
\oursrow\quad ours & \vSnAccMOursOrig & \vSnAccMOursSeedA & \vSnAccMOursSeedB & \vSnAccMOursNa & \vSnNaMOursAcc\ [\vSnNaMOursAccLo, \vSnNaMOursAccHi] & \vSnWrMOursOrig & \vSnWrMOursNa & \vSnAuMOursOrig & \vSnAuMOursNa \\
\multicolumn{10}{@{}l}{\emph{Qwen3.5-4B}} \\
\quad writes & \vSnAccLPtsOrig & \vSnAccLPtsSeedA & -- & \vSnAccLPtsNa & \vSnNaLPtsAcc\ [\vSnNaLPtsAccLo, \vSnNaLPtsAccHi] & \vSnWrLPtsOrig & \vSnWrLPtsNa & \vSnAuLPtsOrig & \vSnAuLPtsNa \\
\oursrow\quad ours & \vSnAccLOursOrig & \vSnAccLOursSeedA & -- & \vSnAccLOursNa & \vSnNaLOursAcc\ [\vSnNaLOursAccLo, \vSnNaLOursAccHi] & \vSnWrLOursOrig & \vSnWrLOursNa & \vSnAuLOursOrig & \vSnAuLOursNa \\
\bottomrule
\end{tabular}

\caption{\textbf{Seeds and the data control for Qwen3.5.} The main run is that of \cref{tab:dec,tab:sets}, and \cref{tab:decb} gives the mean over these seeds. The seed columns retrain it with another seed (--: not run), and the no-agent column retrains it without the agent-style questions of the training mix; $\Delta$ is that run minus the main run.}
\label{tab:seedsna}\label{tab:allrows}
\end{table*}

\begin{table}[!htbp]
\centering\footnotesize
\setlength{\tabcolsep}{2.5pt}\renewcommand{\arraystretch}{0.85}
\begin{tabular}{@{}lccc@{}}
\toprule
Qwen3.5-2B & \begin{tabular}[b]{@{}c@{}}point\\acc. (\%)\end{tabular} & $\Delta$ & 95\% CI \\
\midrule
fine-tuned to write its point & 68.7 & -- & -- \\
+ each axis asked apart & 70.4 & +1.6 & [+0.7, +2.5] \\
+ box as a set & 72.3 & +1.9 & [+0.9, +3.0] \\
+ box edges (vs.\ box as a set) & \textcolor{gray}{72.9} & \textcolor{gray}{+0.6} & \textcolor{gray}{[$-$0.3, +1.5]} \\
+ self-check (vs.\ box as a set) & \textcolor{gray}{72.6} & \textcolor{gray}{+0.3} & \textcolor{gray}{[$-$0.5, +1.2]} \\
\oursrow + both = ours (vs.\ box as a set) & \textbf{73.3} & +1.0 & [+0.1, +1.9] \\
\midrule
\multicolumn{4}{@{}l}{\emph{Changing the box-as-a-set run (72.3)}} \\
\quad mass spread evenly over the box & 68.9 & $-$3.4 & [$-$4.4, $-$2.4] \\
\quad point window 4\% (ours: 10\%) & \textcolor{gray}{72.5} & \textcolor{gray}{+0.1} & \textcolor{gray}{[$-$0.7, +1.0]} \\
\quad point window 20\% & \textcolor{gray}{72.2} & \textcolor{gray}{$-$0.1} & \textcolor{gray}{[$-$1.0, +0.7]} \\
\quad a linear output head & 70.1 & $-$2.2 & [$-$3.2, $-$1.2] \\
\quad linear head, 3$\times$ learning rate & 70.9 & $-$1.4 & [$-$2.4, $-$0.4] \\
\midrule
\multicolumn{4}{@{}l}{\emph{Changing ours (73.3)}} \\
\quad no cross-entropy on points & 70.6 & $-$2.7 & [$-$3.6, $-$1.8] \\
\quad self-check yes if point in box & \textcolor{gray}{72.5} & \textcolor{gray}{$-$0.8} & \textcolor{gray}{[$-$1.7, +0.1]} \\
\quad exact box edges, no intervals & \textcolor{gray}{73.0} & \textcolor{gray}{$-$0.3} & \textcolor{gray}{[$-$1.2, +0.5]} \\
\bottomrule
\end{tabular}

\caption{\textbf{The ablation of the main text with intervals.} Point accuracy of the main training run of each Qwen3.5-2B row (seed 0), its difference from the run it modifies and the paired bootstrap 95\% CI of that difference over images (grey: it includes zero). \Cref{tab:ablation} gives the means over seeds.}
\label{tab:ablationfull}
\end{table}

\paragraph{One choice replaced.} If the training target spreads its mass evenly over the box, accuracy falls back to that of the point-writing model ($\vAfSpreadD$ pp, 95\% CI $\vAfSpreadLo$ to $\vAfSpreadHi$), because an even spread again puts mass on the parts of the box that are not the object. The extra loss on annotated points also matters: without it our accuracy falls by \nosign{\vAfNoPtCeD}~pp (95\% CI $\vAfNoPtCeLo$ to $\vAfNoPtCeHi$). The size of the point window, the rule that labels the self-check and exact box edges change accuracy within noise (\cref{tab:ablationfull}). Labelling the self-check yes when the model's most probable point lies in the label set ranks errors as well as ours (AUROC \vAxCtopAuc{} against \vAxCtopAucFin, difference \vAxCtopAucDiffV, 95\% CI \vAxCtopAucLoV{} to \vAxCtopAucHiV), but it says no on \vAxCtopNo\% of images against \vNmFinNoShare\% and is wrong on \vAfCtopWrong\% of its most confident 70\% against \vSvmOurs\%.

\begingroup\renewcommand{\stale}[1]{#1}
\begin{table}[!htbp]
\centering\footnotesize
\setlength{\tabcolsep}{2pt}
\begin{tabular}{@{}>{\raggedright\arraybackslash}p{0.31\columnwidth}cccc@{}}
\toprule
One change to the final recipe & \stwo{point}{acc. (\%)\,$\uparrow$} & \stwo{$\Delta$ point}{(95\% CI)} & \stwo{box}{acc. (\%)\,$\uparrow$} & \stwo{$\Delta$}{box} \\
\midrule
\oursrow Final recipe (ours) & \vPlFinPt & -- & \vPlFinBx & -- \\
Box edges trained on $P(\mathrm{IoU}\ge0.5)$ & \vPlEiPt & \vPlEiPtD\ [\vPlEiPtLo, \vPlEiPtHi] & \vPlEiBx & \vPlEiBxD \\
Base-model 1\,MP teacher, gate 0.5 & \vPlTfPt & \vPlTfPtD\ [\vPlTfPtLo, \vPlTfPtHi] & \vPlTfBx & \vPlTfBxD \\
Base-model 1\,MP teacher, gate 0.7 & \vPlTsPt & \vPlTsPtD\ [\vPlTsPtLo, \vPlTsPtHi] & \vPlTsBx & \vPlTsBxD \\
Point cross-entropy weight 2 & \vPlCtPt & \vPlCtPtD\ [\vPlCtPtLo, \vPlCtPtHi] & \vPlCtBx & \vPlCtBxD \\
Cross-entropy on annotated edges & \vPlEcPt & \vPlEcPtD\ [\vPlEcPtLo, \vPlEcPtHi] & \vPlEcBx & \vPlEcBxD \\
\bottomrule
\end{tabular}

\caption{\textbf{Single changes to the final Qwen3.5-2B recipe.} Each row changes one choice of the final recipe (first row) and gives point accuracy and box accuracy, each with $\Delta$ against the final recipe. Box acc.\ is Acc@0.5 averaged over the eight RefCOCO/+/g splits at 1\,MP. In the teacher rows, the mass inside the set is fitted to the base model's read at 1\,MP, on the records where that read puts at least the gate's share of its mass on the set.}
\label{tab:pilots}\label{tab:controls}\label{tab:ablate}
\end{table}
\endgroup

\paragraph{Variants of the final recipe.} None of the single changes of \cref{tab:pilots} raises point or box accuracy over the final recipe, so none met the adoption rule we fixed before training. Under that rule, a variant replaces the final recipe only if the 95\% CI of its point-accuracy gain lies above zero or, for a variant aimed at boxes, if it raises box accuracy by at least one point while the upper end of its point interval is not below zero. Training a box on the probability that its IoU reaches 0.5 is worse on boxes than training each edge on its interval, and a cross-entropy on the exact annotated edges lowers box accuracy by \nosign{\vPlEcBxD} pp (\cref{tab:pilots}). The linear output head of \cref{tab:head} also gives a larger 90\% region than the read-out (\vHdReg{} against \vHdRoReg\% of the image).

\begingroup\renewcommand{\stale}[1]{#1}
\paragraph{The head and rare values.} Even with a three times larger learning rate, the 1,000-value head stays \nosign{\vHdXPtDelta}~pp behind reading the digits in point accuracy (95\% CI \vHdXPtLo{} to \vHdXPtHi). Rare values do not explain the head's gap in \cref{tab:head}: the 100 most frequent values of an axis hold \vHtTopHundShareX\% ($x$) and \vHtTopHundShareY\% ($y$) of the training label sets (uniform: 10\%), and inside each benchmark the gap on rare and on common target values differs within noise (\vHtStrRareMinusCommonLo{} to \vHtStrRareMinusCommonHi{} pp).

\paragraph{A self-teacher.} Distilling a 1\,MP read, clipped to the label set, into the \vfEdgeRes{} read~\cite{vapnik2009lupi,lv2020proden,rpopsd} did not help: the model's own 1\,MP read as teacher gives \vEmJDPt\% against \vEmJRPt\% with labels alone, within the spread over seeds (\cref{tab:seedsna}), in runs with an earlier training layout, and the base-model teacher of \cref{tab:pilots} lowers point accuracy, so the final recipe uses no teacher.
\endgroup

\subsection{Seeds and the data control}
\label{app:seedsna}

Over three seeds, the point accuracy of our 2B model spans \vSnSeedSpreadM{} pp (\cref{tab:seedsna}). Removing the \vfAgentN{} agent-style questions moves point accuracy by \vSnNaAccMin{} to \vSnNaAccMax{} pp with every interval including zero.

\section{Knowing When to Act}
\label{app:falsepos}

\begin{table*}[!tp]
\centering\footnotesize
\setlength{\tabcolsep}{3.2pt}\renewcommand{\arraystretch}{0.86}
\begin{tabular}{@{}lccccc@{\hspace{7pt}}>{\columncolor{oursbg}}c>{\columncolor{oursbg}}c>{\columncolor{oursbg}}c>{\columncolor{oursbg}}c>{\columncolor{oursbg}}c>{\columncolor{oursbg}}c@{\hspace{7pt}}c>{\columncolor{oursbg}}c@{}}
\toprule
 & \multicolumn{5}{c}{writes: AUROC\,$\uparrow$ of each training-free score} & \multicolumn{6}{c}{ours: AUROC\,$\uparrow$ of each score} & \multicolumn{2}{c}{wrong@70\% (\%)\,$\downarrow$} \\
Model & peak & \stwo{digit}{entropy} & \stwo{map}{entropy} & window & region & peak & \stwo{digit}{entropy} & \stwo{map}{entropy} & window & region & \stwo{self-}{check} & \stwo{writes,}{best score} & \stwo{ours,}{self-check} \\
\midrule
Qwen3.5-0.8B & 0.73 & 0.77 & 0.75 & 0.78 & 0.69 & 0.71 & 0.77 & 0.76 & 0.76 & 0.58 & \textbf{0.82} & 25.6 & \textbf{19.2} \\
Qwen3.5-2B & 0.71 & 0.76 & 0.74 & 0.79 & 0.68 & 0.68 & 0.77 & 0.74 & 0.76 & 0.56 & \textbf{0.81} & 18.8 & \textbf{14.2} \\
Qwen3.5-4B & 0.70 & 0.76 & 0.73 & 0.78 & 0.63 & 0.63 & 0.76 & 0.71 & 0.76 & 0.47 & \textbf{0.82} & 13.5 & \textbf{9.3} \\
InternVL3.5-1B & 0.78 & 0.78 & 0.77 & 0.80 & 0.75 & 0.69 & 0.74 & 0.74 & 0.73 & 0.62 & \textbf{0.81} & 28.3 & \textbf{23.0} \\
InternVL3.5-2B & 0.74 & 0.76 & 0.73 & 0.79 & 0.71 & 0.64 & 0.76 & 0.72 & 0.76 & 0.53 & \textbf{0.80} & 19.2 & \textbf{15.0} \\
InternVL3.5-4B & 0.72 & 0.75 & 0.73 & 0.77 & 0.67 & 0.65 & 0.76 & 0.72 & 0.76 & 0.53 & \textbf{0.81} & 16.3 & \textbf{\vSvivlDOurs} \\
LFM2.5-VL-3B & 0.73 & 0.76 & 0.75 & 0.78 & 0.64 & 0.66 & 0.74 & 0.72 & 0.74 & 0.51 & \textbf{0.80} & 15.6 & \textbf{10.8} \\
LFM3-VL-2B & 0.75 & 0.77 & 0.76 & 0.80 & 0.72 & 0.74 & 0.75 & 0.70 & 0.76 & 0.63 & \textbf{0.81} & 16.5 & \textbf{13.5} \\
Gemma-4-E4B & 0.61 & 0.64 & 0.57 & 0.72 & 0.71 & 0.55 & 0.58 & 0.56 & 0.66 & 0.59 & \textbf{0.79} & 17.0 & \textbf{12.6} \\
Qwen3-VL-2B & \vRcQvlTwoPPeakAu & \vRcQvlTwoPPaceAu & \vRcQvlTwoPEntAu & \vRcQvlTwoPWinAu & \vRcQvlTwoPRegionAu & \vRcQvlTwoOursPeakAu & \vRcQvlTwoOursPaceAu & \vRcQvlTwoOursEntAu & \vRcQvlTwoOursWinAu & \vRcQvlTwoOursRegionAu & \textbf{\vRcQvlTwoOursYesAu} & \vRcQvlTwoPBestW & \textbf{\vRcQvlTwoOursYesW} \\
Qwen3-VL-4B & \vRcQvlFourPPeakAu & \vRcQvlFourPPaceAu & \vRcQvlFourPEntAu & \vRcQvlFourPWinAu & \vRcQvlFourPRegionAu & \vRcQvlFourOursPeakAu & \vRcQvlFourOursPaceAu & \vRcQvlFourOursEntAu & \vRcQvlFourOursWinAu & \vRcQvlFourOursRegionAu & \textbf{\vRcQvlFourOursYesAu} & \vRcQvlFourPBestW & \textbf{\vRcQvlFourOursYesW} \\
Rex-Omni-3B & \vRcRexPPeakAu & \vRcRexPPaceAu & \vRcRexPEntAu & \vRcRexPWinAu & \vRcRexPRegionAu & \vRcRexOursPeakAu & \vRcRexOursPaceAu & \vRcRexOursEntAu & \vRcRexOursWinAu & \vRcRexOursRegionAu & \textbf{\vRcRexOursYesAu} & \vRcRexPBestW & \textbf{\vRcRexOursYesW} \\
\bottomrule
\end{tabular}

\caption{\textbf{AUROC of each confidence score, per model.} The left block ranks the points of the point-writing model (writes) by each training-free score of its map, the middle block those of the decision model (ours) by the same scores and by its self-check, and the right block gives the wrong share among the most confident 70\%, with the point-writing model ranked by its best training-free score. The region score of Qwen3-VL and Rex-Omni was not computed (--). Rex-Omni writes one location token per value, so its digit entropy is taken over the three digits of that value. Digit entropy is the place-aware entropy of~\cite{pace}, map entropy that of the two axis distributions, window the mass of the $10\%{\times}10\%$ window around the mode, region the size of the 90\% region and peak the mass of the most probable cell of the $100{\times}100$ grid. One run per model.}
\label{tab:freeconf}
\end{table*}

Acting only where $P(\text{yes})\ge0.5$, ours at 2B acts on \vFtQmAct\% of the \vBcNAll{} images and is wrong on \vFtQmOurs\% of the images it acts on, while the point-writing model acts on every image and is wrong on \vFtQmPts\%. $P(\text{yes})$ estimates whether half of the map's mass lies on the label set, which a large box covers more easily than a point's small window, so we use it to rank decisions and to choose between acting and holding, and not as the probability that the point is right.

\paragraph{The self-check against training-free scores of the same map.}\label{app:freeconf} For every model, the self-check adds \vRvAuYesGainMin{} to \vRvAuYesGainMax{} AUROC over the peak of our own map (\cref{tab:freeconf}), and on the most confident 70\% ours is wrong \vRcElRelBestMin--\vRcElRelBestMax\% less often, in relative terms, than the point-writing model ranked by its best training-free score.

\paragraph{Points with the same self-check.}\label{app:ptsq} The label sets matter beyond the self-check. A Qwen3.5-2B trained on points with the same self-check, labelled as for ours, is right on \vPcAccPtsQ\% against \vPcAccOurs\% for ours, and on its most confident 70\% it is wrong on \vPcWrYesPtsQ\% against \vPcWrYesOurs\% (ours minus this model \vPcDWrOursQ{} pp, 95\% CI \vPcDWrOursQLo{} to \vPcDWrOursQHi).

\subsection{Calibrated regions}
\label{app:conformal}

The 90\% region is the set of most probable cells of the $100{\times}100$ grid, $R_\tau=\{c:\mu(c)\ge \tau\}$~\cite{izbicki2022hpd}, with $\tau$ set by split conformal prediction at $\alpha=0.1$~\cite{angelopoulos2023conformal,sadinle2019lac}. One threshold per model is set on \vNValN{} held-out records of the training mixture at $384^2$, none shared with training or test by identifier, perceptual hash, pixel hash or file hash, and used unchanged on every benchmark. The score $s_k=1-\max_{c\,:\,c\cap A_k\neq\emptyset}\mu_k(c)$ is taken against the record's label set $A_k$, so the 90\% is the share of images on which the region touches the label set. For the off-the-shelf models of \cref{tab:families}, the threshold comes from one random half of each benchmark and is tested on the other half (20 splits).

The region of our 2B model covers \vNmFinReg\% of the image on average and holds the target on \vNmFinCov\% of all images: on \vNmFinSSCov\% of ScreenSpot-v2, \vNmFinRRCov\% of RoboRefIt and \vNmFinRCCov\% of RefCOCO, and on \vNmFinPBCov\% of Point-Bench and \vNmFinWPCov\% of Where2Place, which are furthest from the training data. With the threshold set instead on ten labelled records of the benchmark itself, drawn at random many times, the region holds the target on \vNmFsPBTenCov\% of Point-Bench and on \vNmFsWPTenCov\% of Where2Place on average, and in nine of ten draws on at least \vNmFsPBTenLow\% and \vNmFsWPTenLow\%.

\subsection{One use of a no: looking closer}
\label{app:closerrows}

\begin{figure}[!htbp]
\centering
\includegraphics[width=\linewidth]{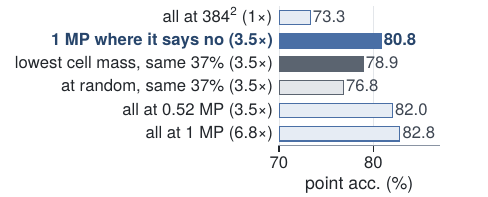}
\caption{\textbf{Reading again where the self-check says no.} Accuracy of our Qwen3.5-2B for six ways to spend pixels (in brackets, one \vfEdgeRes{} image is 1). The dark bar re-reads at 1\,MP where the self-check says no, the grey bars re-read the same share chosen by the lowest top-cell mass or at random, and the light bars read every image at one resolution. The pixels of a re-read count both reads. Reading every image at 0.52\,MP spends the same pixels and is slightly more accurate, but every decision then waits for the larger read, while re-reading where it says no keeps the images it says yes on at the cost of one $384^2$ read.}
\label{fig:reread}
\end{figure}

\Cref{tab:closer} gives the re-read of \cref{sec:closer} for each Qwen3.5 model, and \cref{fig:reread} shows the 2B row as bars. At every size, re-reading where the self-check says no is more accurate than re-reading the same share of images chosen by the weakest map (least peak mass) or at random, and it recovers \vNxFinNoGain{} to \vNsFinNoGain\% of the gain of re-reading every image at 1\,MP while reading again only \vNxFinNoShare{} to \vNsFinNoShare\% of them.

\begin{table}[!htbp]
\centering\footnotesize
\setlength{\tabcolsep}{3pt}
\begin{tabular}{@{}lcccc@{}}
\toprule
Our Qwen3.5 model & 0.8B & 2B & 4B & 9B \\
\midrule
\multicolumn{5}{@{}l}{\emph{Point acc.\ (\%)\,$\uparrow$ when re-reading at 1\,MP}} \\
\quad none ($384^2$ only) & \vNsFinLo & \vNmFinLo & \vNlFinLo & \vNxFinLo \\
\oursrow\quad the images it says no on & \textbf{\vNsFinNoAcc} & \textbf{\vNmFinNoAcc} & \textbf{\vNlFinNoAcc} & \textbf{\vNxFinNoAcc} \\
\quad the same share, lowest top-cell mass & \vNsFinNoTm & \vNmFinNoTm & \vNlFinNoTm & \vNxFinNoTm \\
\quad the same share, at random & \vNsFinNoRn & \vNmFinNoRn & \vNlFinNoRn & \vNxFinNoRn \\
\quad every image & \vNsFinHi & \vNmFinHi & \vNlFinHi & \vNxFinHi \\
\multicolumn{5}{@{}l}{\emph{Cost and gain of re-reading where it says no}} \\
\quad says no on (\% of images) & \vNsFinNoShare & \vNmFinNoShare & \vNlFinNoShare & \vNxFinNoShare \\
\quad pixels read ($384^2$ reads)\,$\downarrow$ & \vNsFinNoPix & \vNmFinNoPix & \vNlFinNoPix & \vNxFinNoPix \\
\quad gain of every-image read recovered (\%)\,$\uparrow$ & \vNsFinNoGain & \vNmFinNoGain & \vNlFinNoGain & \vNxFinNoGain \\
\bottomrule
\end{tabular}

\caption{\textbf{Re-reading at 1\,MP, per Qwen3.5 model.} Every image is first read at \vfEdgeRes; the rows choose which images are read again at 1\,MP, whose answer is kept. The lower block gives the share where the model says no, the pixels read per image over both reads in units of one \vfEdgeRes{} read, which is one for the first read plus \vNPixRatio{} for each image read again at 1\,MP (at 2B, $1+\vNmFinNoShare\%\times\vNPixRatio=\vNmFinNoPix$), and the share of the gain of re-reading every image that is recovered (from unrounded accuracies).}
\label{tab:closer}
\end{table}

\section{Decisions in a Closed Loop}
\label{app:closed}

\begin{table*}[!tp]
\centering\footnotesize
\setlength{\tabcolsep}{2.6pt}
\begin{tabular}{@{}lccccccc@{}}
\toprule
& \multicolumn{5}{c}{episodes that reach the target\,$\uparrow$} & & \\
\cmidrule(lr){2-6}
& \stwo{in view}{(of 20)} & \stwo{search}{(of 20)} & \stwo{track}{(of 20)} & \stwo{search +}{track (of 20)} & \stwo{all of 80}{(in radius)} & \stwo{time per}{decision\,$\downarrow$} & \stwo{all,}{each run} \\
\midrule
\multicolumn{8}{@{}l}{\emph{L40S time and simulator frame (first of three runs, ms per decision)}} \\
\quad Qwen3.5-0.8B trained as in DroneCATS, writes JSON & \vNsQJGpuApp & \vNsQJGpuSea & \vNsQJGpuTra & \vNsQJGpuSet & \vNsQJGpuReach\,(\vNsQJGpuNear) & \vNsQJGpuStep & \vLrQJGpuRuns \\
\oursrow\qquad + our recipe, decides & 6 & 5 & 5 & 4 & \vDrReachQDGpu\,(\vDrNearQDGpu) & \vDrMsQDGpu & \vLrQDGpuRuns \\
\quad DroneCATS-2B, released, writes JSON & \vNsCJGpuApp & \vNsCJGpuSea & \vNsCJGpuTra & \vNsCJGpuSet & \vNsCJGpuReach\,(\vNsCJGpuNear) & \vNsCJGpuStep & \vLrCJGpuRuns \\
\oursrow\qquad + our recipe, decides & 7 & 8 & 10 & 4 & \vDrReachCDGpu\,(\vDrNearCDGpu) & \vDrMsCDGpu & \vLrCDGpuRuns \\
\multicolumn{8}{@{}l}{\emph{Pi 5 time, simulator frame (one run, s per decision)}} \\
\quad Qwen3.5-0.8B trained as in DroneCATS, writes JSON & \vNsQJPiApp & \vNsQJPiSea & \vNsQJPiTra & \vNsQJPiSet & \vNsQJPiReach\,(\vNsQJPiNear) & \vPiMeanQJ & \vNsQJPiRuns \\
\oursrow\qquad + our recipe, decides & 8 & 6 & 4 & 4 & \vDrReachQDPi\,(\vDrNearQDPi) & \vPiMeanQD & -- \\
\quad DroneCATS-2B, released, writes JSON & \vNsCJPiApp & \vNsCJPiSea & \vNsCJPiTra & \vNsCJPiSet & \vNsCJPiReach\,(\vNsCJPiNear) & \vPiMeanCJ & \vNsCJPiRuns \\
\oursrow\qquad + our recipe, decides & 5 & 6 & 0 & 0 & \vDrReachCDPi\,(\vDrNearCDPi) & \vPiMeanCD & -- \\
\multicolumn{8}{@{}l}{\emph{Pi 5 time and frame (mean over runs, s per decision)}} \\
\quad Qwen3.5-0.8B trained as in DroneCATS, writes JSON & \vNsQJPiSApp & \vNsQJPiSSea & \vNsQJPiSTra & \vNsQJPiSSet & \vNsQJPiSReach\,(\vNsQJPiSNear) & \vPiMeanQJ & \vNsQJPiSRuns \\
\oursrow\qquad + our recipe, decides & 2.7 & 2.3 & 4.0 & 2.0 & \vDrReachQDPiS\,(\vDrNearQDPiS) & \vPiMeanQD & 11, 9, 13 \\
\quad DroneCATS-2B, released, writes JSON & \vNsCJPiSApp & \vNsCJPiSSea & \vNsCJPiSTra & \vNsCJPiSSet & \vNsCJPiSReach\,(\vNsCJPiSNear) & \vPiMeanCJ & \vNsCJPiSRuns \\
\oursrow\qquad + our recipe, decides & 2.3 & 0.7 & 0.0 & 0.0 & \vDrReachCDPiS\,(\vDrNearCDPiS) & \vPiMeanCD & 3, 2, 4 \\
\bottomrule
\end{tabular}

\caption{\textbf{Drone episodes per scenario, timing setting and run.} Episodes that reach the target in each scenario (20 each), with those that come within the success radius at some point in brackets, and the median time per decision. The three blocks are the L40S time on the simulator's frames, the Pi time on the simulator's frames, and the Pi time with the Pi's 256-pixel frame, where cells are means over runs. In both Pi blocks each answer is delayed by the Pi 5's measured time for that model at the 256-pixel frame (\cref{tab:piinputs}), so the middle block separates the delay from the frame size. The last column lists each run's total (--: one run), also for the three L40S runs whose means are in \cref{tab:dronebody}. Every model acts with the same three actions, so the JSON writers run without the benchmark's THINK action. With it, the 0.8B writer reaches the target in \vDrReachQJGpu, \vDrReachQJPi{} and \vDrReachQJPiS{} episodes in the three blocks and DroneCATS-2B in \vDrReachCJGpu, \vDrReachCJPi{} and \vDrReachCJPiS.}
\label{tab:droneruns}\label{tab:dronefull}
\end{table*}
\phantomsection\label{app:dronelat}
\begin{table}[!htbp]
\centering\footnotesize
\setlength{\tabcolsep}{2.5pt}
\begin{tabular}{@{}l@{\hspace{7pt}}cc@{\hspace{7pt}}>{\columncolor{oursbg}}c>{\columncolor{oursbg}}c>{\columncolor{oursbg}}c@{}}
\toprule
& \multicolumn{2}{c}{\stwo{writes or released:}{success (\%)\,$\uparrow$}} & \multicolumn{3}{c}{\stwo{ours:}{success (\%)\,$\uparrow$}} \\
Model & click & re-read$^r$ & click & re-read & acts \\
\midrule
\multicolumn{6}{@{}l}{\emph{Released models}} \\
\quad GUI-Owl-1.5-2B$^\dagger$ & \vNowlMw & \vNowlMwG & \multicolumn{1}{c}{--} & \multicolumn{1}{c}{--} & \multicolumn{1}{c@{}}{--} \\
\quad GUI-Decisions-2B~\cite{guidecisions} & \vNguidMw & \vNguidMwG & \multicolumn{1}{c}{--} & \multicolumn{1}{c}{--} & \multicolumn{1}{c@{}}{--} \\
\quad GroundingPI-4.8B & \vNgpiMw & \vNgpiMwG & \multicolumn{1}{c}{--} & \multicolumn{1}{c}{--} & \multicolumn{1}{c@{}}{--} \\
\quad MJev-Spatial-4B & \vNmjevMw & -- & \multicolumn{1}{c}{--} & \multicolumn{1}{c}{--} & \multicolumn{1}{c@{}}{--} \\
\multicolumn{6}{@{}l}{\emph{Qwen3.5, trained by us}} \\
\quad Qwen3.5-0.8B & \vNsSFTMw & \vNsSFTMwG & \vNsFinMw & \vNsFinMwY & \vNsFinMwYAct \\
\quad Qwen3.5-2B & \vNmSFTMw & \vNmSFTMwG & \vNmFinMw & \vNmFinMwY & \vNmFinMwYAct \\
\quad Qwen3.5-4B & \vNlSFTMw & \vNlSFTMwG & \vNlFinMw & \vNlFinMwY & \vNlFinMwYAct \\
\quad Qwen3.5-9B & \vNxSFTMw & \vNxSFTMwG & \vNxFinMw & \vNxFinMwY & \vNxFinMwYAct \\
\multicolumn{6}{@{}l}{\emph{InternVL3.5, trained by us}} \\
\quad InternVL3.5-1B & \vNivlASFTMw & -- & \vNivlAFinMw & \vNivlAFinMwY & \vNivlAFinMwYAct \\
\quad InternVL3.5-2B & \vNivlSFTMw & \vNivlSFTMwG & \vNivlFinMw & \vNivlFinMwY & \vNivlFinMwYAct \\
\quad InternVL3.5-4B & \vNivlDSFTMw & -- & \vNivlDFinMw & \vNivlDFinMwY & \vNivlDFinMwYAct \\
\quad InternVL3.5-8B & \vNivlHSFTMw & -- & \vNivlHFinMw & \vNivlHFinMwY & \vNivlHFinMwYAct \\
\multicolumn{6}{@{}l}{\emph{Other families, trained by us}} \\
\quad Gemma-4-E4B & \vNgemSFTMw & -- & \vNgemFinMw & \vNgemFinMwY & \vNgemFinMwYAct \\
\quad LFM2.5-VL-3B & \vNlfmSFTMw & \vNlfmSFTMwG & \vNlfmFinMw & \vNlfmFinMwY & \vNlfmFinMwYAct \\
\quad LFM3-VL-2B & \vNlfvSFTMw & \vNlfvSFTMwG & \vNlfvFinMw & \vNlfvFinMwY & \vNlfvFinMwYAct \\
\quad LLaVA-1.5-7B & \vNllvSFTMw & \vNllvSFTMwG & \vNllvFinMw & -- & -- \\
\bottomrule
\end{tabular}

\caption{\textbf{Success on the MiniWoB++ single-click tasks} (\%, 15 tasks $\times$ 50 seeds, target phrase given). Click is the most probable point. A re-read clicks the new most probable point in a crop around the region, for ours where it answers no and for the others where their top cell holds less than half of the mass ($^r$). Acts is the share of episodes in which ours answers yes and clicks without re-reading. $^\dagger$GUI-Owl-1.5-2B is given the pointing instruction of the other rows here, and in its own tool-call format it succeeds on \vGoMw\% with one click, the number of \cref{tab:webbody}. --: not run.}
\label{tab:miniwob}\label{tab:closed}\label{tab:miniwobnative}
\end{table}
\begin{table}[!htbp]
\centering\footnotesize
\setlength{\tabcolsep}{2pt}
\begin{tabular}{@{}lccc@{}}
\toprule
Model & \stwo{click acc.}{(\%)\,$\uparrow$} & \stwo{wrong@70\%}{(\%)\,$\downarrow$} & AUROC\,$\uparrow$ \\
\midrule
\multicolumn{4}{@{}l}{\emph{Qwen3.5-0.8B}} \\
\quad off-the-shelf & \vWbMwBaseZeroEightAcc & -- & -- \\
\quad writes & \vWbMwPtsZeroEightAcc & \vWbMwPtsZeroEightWrong & \vWbMwPtsZeroEightAuc \\
\oursrow\quad ours & \vWbMwOursZeroEightAcc & \vWbMwOursZeroEightWrong & \vWbMwOursZeroEightAuc \\
\quad\scriptsize ours $-$ writes & \scriptsize \vWbMwDiffZeroEight\,[\vWbMwDiffZeroEightLo, \vWbMwDiffZeroEightHi] & \scriptsize \vWbMwWrongDiffZeroEight\,[\vWbMwWrongDiffZeroEightLo, \vWbMwWrongDiffZeroEightHi] & \scriptsize \vWbMwAucDiffZeroEight\,[\vWbMwAucDiffZeroEightLo, \vWbMwAucDiffZeroEightHi] \\
\multicolumn{4}{@{}l}{\emph{Qwen3.5-2B}} \\
\quad off-the-shelf & \vWbMwBaseTwoAcc & -- & -- \\
\quad writes & \vWbMwPtsTwoAcc & \vWbMwPtsTwoWrong & \vWbMwPtsTwoAuc \\
\oursrow\quad ours & \vWbMwOursTwoAcc & \vWbMwOursTwoWrong & \vWbMwOursTwoAuc \\
\quad\scriptsize ours $-$ writes & \scriptsize \vWbMwDiffTwo\,[\vWbMwDiffTwoLo, \vWbMwDiffTwoHi] & \scriptsize \vWbMwWrongDiffTwo\,[\vWbMwWrongDiffTwoLo, \vWbMwWrongDiffTwoHi] & \scriptsize \vWbMwAucDiffTwo\,[\vWbMwAucDiffTwoLo, \vWbMwAucDiffTwoHi] \\
\multicolumn{4}{@{}l}{\emph{Qwen3.5-4B}} \\
\quad off-the-shelf & \vWbMwBaseFourAcc & -- & -- \\
\quad writes & \vWbMwPtsFourAcc & \vWbMwPtsFourWrong & \vWbMwPtsFourAuc \\
\oursrow\quad ours & \vWbMwOursFourAcc & \vWbMwOursFourWrong & \vWbMwOursFourAuc \\
\quad\scriptsize ours $-$ writes & \scriptsize \vWbMwDiffFour\,[\vWbMwDiffFourLo, \vWbMwDiffFourHi] & \scriptsize \vWbMwWrongDiffFour\,[\vWbMwWrongDiffFourLo, \vWbMwWrongDiffFourHi] & \scriptsize \vWbMwAucDiffFour\,[\vWbMwAucDiffFourLo, \vWbMwAucDiffFourHi] \\
\bottomrule
\end{tabular}

\caption{\textbf{Multimodal-Mind2Web, three test splits pooled} (\vWbMwTaskN, \vWbMwSiteN{} and \vWbMwDomN{} elements for cross-task, cross-website and cross-domain). A click is correct if it falls inside the element's box. Wrong@70\% and AUROC rank ours by its self-check and the point-writing model by its top $10{\times}10$ cell mass. The small rows give ours minus the point-writing model (writes), with paired bootstrap 95\% CIs over elements.}
\label{tab:webmw}
\end{table}

\begin{table*}[!tp]
\centering\footnotesize
\setlength{\tabcolsep}{3.2pt}
\begin{tabular}{@{}l@{\hspace{7pt}}c>{\columncolor{oursbg}}c>{\columncolor{oursbg}}cc@{\hspace{7pt}}c>{\columncolor{oursbg}}cc@{\hspace{7pt}}>{\columncolor{oursbg}}c@{\hspace{7pt}}c>{\columncolor{oursbg}}c@{\hspace{7pt}}c>{\columncolor{oursbg}}c@{}}
\toprule
& \multicolumn{4}{c}{ms per decision\,$\downarrow$} & \multicolumn{3}{c}{point acc.\ (\%)\,$\uparrow$} & \multicolumn{1}{c}{\stwo{box}{acc.\ (\%)\,$\uparrow$}} & \multicolumn{2}{c}{\stwo{MiniWoB++}{success (\%)\,$\uparrow$}} & \multicolumn{2}{c}{\stwo{wrong@70\%}{(\%)\,$\downarrow$}} \\
Model & writes & \stwo{ours,}{$10^2$} & \stwo{ours,}{$1000^2$} & \stwo{faster}{$10^2$ / $1000^2$} & writes & ours & \stwo{$\Delta$ ours $-$}{writes (95\% CI)} & ours & writes & ours & writes & ours \\
\midrule
Qwen3.5-9B & \vWfxVl & \vWfxReadC & \vOsxThou & \vWfxVlRatioC/\vOsxVlRatioThou$\times$ & \vNxSFTPt & \textbf{\vNxFinPt} & \vNCixFinSFT\ [\vNCixFinSFTLo, \vNCixFinSFTHi] & \vNxFinBxAvg & \vNxSFTMw & \textbf{\vNxFinMw} & \vRvWsPtsQx & \textbf{\vRvWsPyesQx} \\
InternVL3.5-8B & \vLHivlHGen$^\ddagger$ & \vLHivlHCell & \vOsHivlHThou & \vLHivlHRatioC/\vOsHivlHRatioThou$\times$ & \vNivlHSFTPt & \textbf{\vNivlHFinPt} & \vNCiivlHFinSFT\ [\vNCiivlHFinSFTLo, \vNCiivlHFinSFTHi] & \vNivlHFinBxAvg & \vNivlHSFTMw & \textbf{\vNivlHFinMw} & \vRvWsPtsIvd & \textbf{\vRvWsPyesIvd} \\
LLaVA-1.5-7B & \vLHllvGen$^\ddagger$ & \vLHllvCell & \vOsHllvThou & \vLHllvRatioC/\vOsHllvRatioThou$\times$ & \vNllvSFTPt & \textbf{\vNllvFinPt} & \vNCillvFinSFT\ [\vNCillvFinSFTLo, \vNCillvFinSFTHi] & \vNllvFinBxAvg & \vNllvSFTMw & \textbf{\vNllvFinMw} & \vRvWsPtsLlv & \textbf{\vRvWsPyesLlv} \\
\bottomrule
\end{tabular}
\caption{\textbf{The 7--9B models, read as in \cref{tab:dec}.} Speed of the point-writing model (writes) and the decision model (ours), point accuracy with $\Delta$ (95\% CI), box Acc@0.5 over the eight RefCOCO/+/g splits, MiniWoB++ success, and the wrong share among the 70\% most confident images, with ours ranked by its self-check and the point-writing model by its top $10{\times}10$ cell, as in \cref{tab:decb}. The point-writing Qwen3.5-9B writes in vLLM, and $^\ddagger$ marks rows written and read in the HF engine.}
\label{tab:large}
\end{table*}

\paragraph{The drone table per scenario and run.}\label{app:dronebeh} \Cref{tab:droneruns} gives the counts per scenario behind \cref{tab:dronebody}, with the L40S block from the first of its three runs. Repeated runs use the same episodes and differ only through the real-time simulator and serving. Across these runs the totals of our 0.8B model on the Pi range over \vDrRangeQDPiS{} episodes, against \vDrRangeQJPiS{} for the 0.8B writer, so the 0.8B gap lies outside the run-to-run spread. Differences of one or two episodes, as between the two 2B models, lie within it.

\paragraph{MiniWoB++.}\label{app:mwrows}\label{app:miniwob}\label{app:mwseeds} On a second set of seeds, ours again succeeds more often than the point-writing model with one click: in \vMwSecNsFinClick, \vMwSecNmFinClick{} and \vMwSecNlFinClick\% of the episodes at 0.8B, 2B and 4B, against \vMwSecNsSFTClick, \vMwSecNmSFTClick{} and \vMwSecNlSFTClick\%. At 4B the re-read lowers our success, because the model often answers no after a correct click: over both sets of seeds, a re-read turns a correct click into a wrong one in \vMwLostLAll{} episodes and a wrong click into a correct one in \vMwWonLAll.

In \cref{tab:miniwob} ours acts without re-reading on \vNsFinMwYAct\% (Qwen3.5-0.8B) to \vNlfmFinMwYAct\% (LFM2.5-VL-3B) of the episodes and reads the page again in the others. The second read raises success for Qwen3.5 at 0.8B, 2B and 9B, for InternVL3.5-1B and for both LFM models, most at 0.8B (\vNsFinMw{} to \vNsFinMwY\%). It lowers success for InternVL3.5 from 2B up and for Gemma-4-E4B, which, like Qwen3.5-4B above, answer no after a correct click. InternVL3.5-4B is the one trained row in which the point-writing model clicks correctly more often than ours with one click (\vNivlDSFTMw{} against \vNivlDFinMw\%).

\paragraph{Multimodal-Mind2Web.}\label{app:mind2web} Scored per element, ours clicks correctly as often as the point-writing model or slightly more often, and at every size it is wrong less often on its most confident 70\% of elements, with every interval of that difference below zero (\cref{tab:webmw}). At 2B, for example, ours is wrong on \vWbMwOursTwoWrong\% against \vWbMwPtsTwoWrong\%. The protocol is that of \cref{app:protocol}.

\paragraph{A larger image.}\label{app:ownpix} A larger image helps the GUI model more than ours in accuracy, and ours still makes the fewest confident errors and keeps the most tasks safe. Every row of \cref{tab:webbody} gets the page under the same budget of 0.6 megapixels (\vOpTokShared{} vision tokens on average). GUI-Owl-1.5 is built for larger inputs, so we repeat the 2B rows at its own budget, which leaves the pages at full size (\vOpTokOwn{} tokens). GUI-Owl-1.5-2B then clicks correctly on \vOpOwlAcc\% of the steps, \vOpOwlGain{} pp more than at the shared budget, with \vOpOwlWrong\% wrong@70\% and \vOpOwlSafe\% of tasks safe. Ours clicks correctly on \vOpOursAcc\%, a gain of \vOpOursGain{} pp that leaves it \vOpOwlOverOurs{} pp below GUI-Owl, with \vOpOursWrong\% wrong@70\% and \vOpOursSafe\% of tasks safe. The point-writing model reaches \vOpPtsAcc\%, \vOpPtsWrong\% and \vOpPtsSafe\% on the same three measures.

\section{Models of 7B to 9B}
\label{app:large}

Our recipe carries over to 7B to 9B models. Qwen3.5-9B, InternVL3.5-8B and LLaVA-1.5-7B gain \nosign{\vNCixFinSFT} to \nosign{\vNCillvFinSFT}~pp of point accuracy over the point-writing model, are wrong less often on their most confident 70\% of images (for Qwen3.5-9B \vRvWsPyesQx\% against \vRvWsPtsQx\%) and click correctly more often on MiniWoB++ (\cref{tab:large}). Their decisions are also faster than the point-writing model's written answers, by \vWfxVlRatioC{} to \vLHllvRatioC{} times on the $10{\times}10$ grid and \vOsxVlRatioThou{} to \vOsHllvRatioThou{} times on the $1000{\times}1000$ grid. LLaVA-1.5 resizes every image to $336^2$, so its self-check label comes from that one resolution.

\section{Comparison with Released and Published Models}
\label{app:standard}
\label{app:pointers}%

\subsection{Released decision and pointing models}
\label{app:h2h}

\begin{table}[!htbp]
\centering\footnotesize
\setlength{\tabcolsep}{4pt}
\begin{tabular}{@{}lccc@{}}
\toprule
cell accuracy (\%) & $2{\times}2$ & $3{\times}3$ & $10{\times}10$ \\
\midrule
d3-nano (2B) & \vCgDnTwo & \vCgDnThree & \vCgDnTen \\
pplx-decider (4B) & \vCgPpTwo & \vCgPpThree & \vCgPpTen \\
\oursrow decision model (ours, Qwen3.5-2B) & \textbf{\vCgOursTwo} & \textbf{\vCgOursThree} & \textbf{\vCgOursTen} \\
\midrule
chance & \vCgChanceTwo & \vCgChanceThree & \vCgChanceTen \\
best fixed cell & \vCgConstTwo & \vCgConstThree & \vCgConstTen \\
\bottomrule
\end{tabular}
\caption{\textbf{Option models choose well among a few large cells and fall far behind ours as the cells shrink.} The same column and row questions as for the $10{\times}10$ decision, with 2 or 3 options each (ten for $10{\times}10$), on the \vBcNAll{} test images. A decision is right when the chosen cell meets the target. Chance is the mean share of cells that meet the target, and the best fixed cell always answers the cell that is right most often. Ours sums its read into the coarser cells.}
\label{tab:coarsegrid}
\end{table}

\begin{table}[!tp]
\centering\footnotesize
\setlength{\tabcolsep}{2.5pt}
\begin{tabular}{@{}l@{\hspace{4pt}}ccc@{\hspace{5pt}}c@{\hspace{5pt}}cc@{}}
\toprule
& \multicolumn{3}{c}{accuracy (\%)\,$\uparrow$} & & \multicolumn{2}{c@{}}{at the point} \\
\cmidrule(lr){2-4}\cmidrule(l){6-7}
Model & $10^2$ & $100^2$ & $1000^2$ & ms\,$\downarrow$ & \stwo{wrong}{@70\%\,$\downarrow$} & AUROC\,$\uparrow$ \\
\midrule
\multicolumn{7}{@{}l}{\emph{0.8B}} \\
d3-lite~\cite{d3nano} & \vRzDlCellAcc & \vRzDlHundAcc & \vRzDlPtAcc & \vFdDlThou & \vRzDlPtFP & \vRzDlPtAu \\
Intern-Decision~\cite{interndecision} & \vRzIntSCellAcc & \vRzIntSHundAcc & \vRzIntSPtAcc & \vFdIntSThou & \vRzIntSPtFP & \vRzIntSPtAu \\
OneJev$^\S$~\cite{onejev} & \vRzOjCellAcc & \vRzOjHundAcc & \vRzOjPtAcc & \vRzOjPtMs & \vRzOjPtFP & \vRzOjPtAu \\
OneJev1.5 & \vRzOjfCellAcc & \vRzOjfHundAcc & \vRzOjfPtAcc & \vFdOjfThou & \vRzOjfPtFP & \vRzOjfPtAu \\
JPT$^\S$~\cite{jpt} & \vRzJpCellAcc & \vRzJpHundAcc & \vRzJpPtAcc & \vRzJpPtMs & \vRzJpPtFP & \vRzJpPtAu \\
Qwen3.5 o.t.s., read & \vRzNatSCellAcc & \vRzNatSHundAcc & \vRwQSRead & \vUtFastsThou & \vRzNatReadSPtFP & \vRzNatReadSPtAu \\
Qwen3.5, writes & \vGcPtsSCell & \vGcPtsSHund & \vRzPtsSPtAcc & \vAuStopS & \vSvsPts & \vAusPts \\
\oursrow\textbf{ours} & \textbf{\vRzOursSCellAcc} & \textbf{\vRzOursSHundAcc} & \textbf{\vRzOursSPtAcc} & \textbf{\vOssThou} & \textbf{\vSvsOurs} & \textbf{\vAusOurs} \\
\midrule
\multicolumn{7}{@{}l}{\emph{2B}} \\
d3-nano & \vRzDnCellAcc & \vRzDnHundAcc & \vRzDnPtAcc & \vFdDnThou & \vRzDnPtFP & \vRzDnPtAu \\
Intern-Decision & \vRzIntMCellAcc & \vRzIntMHundAcc & \vRzIntMPtAcc & \vFdIntMThou & \vRzIntMPtFP & \vRzIntMPtAu \\
Tacit$^\S$~\cite{tacit} & \vRzTaCellAcc & \vRzTaHundAcc & \vRzTaPtAcc & \vRzTaPtMs & \vRzTaPtFP & \vRzTaPtAu \\
imajev$^\S$~\cite{imajev} & \vRzImCellAcc & \vRzImHundAcc & \vRzImPtAcc & \vRzImPtMs & \vRzImPtFP & \vRzImPtAu \\
GUI-Decisions$^\ast$ & \vRzGuidCellAcc & \vRzGuidHundAcc & \vRzGuidPtAcc & \vFdGuidThou & \vRzGuidPtFP & \vRzGuidPtAu \\
Qwen3-VL-2B$^\ast$ & \vGcQvlTwoCell & \vGcQvlTwoHund & \vRgQvlTwoAcc & \vFtQvlTwoVl & \vRgQvlTwoW & \vRgQvlTwoAu \\
Qwen3.5 o.t.s., read & \vRzNatMCellAcc & \vRzNatMHundAcc & \vRwQMRead & \vUtFastmThou & \vRzNatReadMPtFP & \vRzNatReadMPtAu \\
Qwen3.5, writes & \vGcPtsMCell & \vGcPtsMHund & \vRzPtsMPtAcc & \vAuStopM & \vSvmPts & \vAumPts \\
\oursrow\textbf{ours} & \textbf{\vRzOursMCellAcc} & \textbf{\vRzOursMHundAcc} & \textbf{\vRzOursMPtAcc} & \textbf{\vOsmThou} & \textbf{\vSvmOurs} & \textbf{\vAumOurs} \\
\midrule
\multicolumn{7}{@{}l}{\emph{3.7B to 4.8B}} \\
pplx-decider~\cite{pplxdecider} & \vRzPpCellAcc & \vRzPpHundAcc & \vRzPpPtAcc & \vFdPpThou & \vRzPpPtFP & \vRzPpPtAu \\
Intern-Decision & \vRzIntLCellAcc & \vRzIntLHundAcc & \vRzIntLPtAcc & \vFdIntLThou & \vRzIntLPtFP & \vRzIntLPtAu \\
Kanana-2-v-decide~\cite{kananadecide} & \vRzKkCellAcc & \vRzKkHundAcc & \vRzKkPtAcc & \vRzKkPtMs & \vRzKkPtFP & \vRzKkPtAu \\
MJev-Spatial$^\ast$ & \vRzMjevCellAcc & \vRzMjevHundAcc & \vRzMjevPtAcc & \vFdMjevThouDec & \vRzMjevPtFP & \vRzMjevPtAu \\
Qwen3-VL-4B$^\ast$ & \vGcQvlFourCell & \vGcQvlFourHund & \vRgQvlFourAcc & \vFtQvlFourVl & \vRgQvlFourW & \vRgQvlFourAu \\
GroundingPI$^\ast$ & \vGcGpiCell & \vGcGpiHund & \vRgGpiAcc & \vFtGpiVl & \vRgGpiW & \vRgGpiAu \\
Qwen3.5 o.t.s., read & \vRzNatLCellAcc & \vRzNatLHundAcc & \vRwQLRead & \vUtFastlThou & \vRzNatReadLPtFP & \vRzNatReadLPtAu \\
Qwen3.5, writes & \vGcPtsLCell & \vGcPtsLHund & \vRzPtsLPtAcc & \vAuStopL & \vSvlPts & \vAulPts \\
\oursrow\textbf{ours} & \textbf{\vRzOursLCellAcc} & \textbf{\vRzOursLHundAcc} & \textbf{\vRzOursLPtAcc} & \textbf{\vOslThou} & \textbf{\vSvlOurs} & \textbf{\vAulOurs} \\
\bottomrule
\end{tabular}

\caption{\textbf{Released decision and pointing models at every size, with our Qwen3.5 rows.} Each row gives the accuracy on three grids, the median time of the $1000{\times}1000$ answer, and for that answer the wrong share among the 70\% most confident images and the AUROC. The times use CUDA graphs and ask the column and row questions concurrently wherever this leaves the model's choices unchanged; OneJev-0.8B, JPT-0.8B, Tacit-2B, imajev-2B and Kanana keep the times of their first measurement. These are the times of the released runtimes, which encode the image again for every question; \cref{tab:optonce} gives the multi-round models with the image processed once, as in \cref{tab:dec}. ``Qwen3.5 o.t.s., read'' is the off-the-shelf (o.t.s.) Qwen3.5 read in its own format, as in \cref{tab:dec}. Option models are ranked by the product of their answers over the rounds, pointing models ($^\ast$) by their own confidence (Qwen3-VL and GroundingPI: the peak of their read map), the point-writing model (writes) by its top $10{\times}10$ cell and ours by its self-check; $^\S$ marks a model read through its language-model head in HF, not through an endpoint of its own. Ours is in bold.}
\label{tab:releasedauroc}
\end{table}

\paragraph{Released decision models.} The strongest released models are about as accurate as ours, but they are wrong more often on their most confident decisions (\cref{tab:releasedauroc}). Paired over images, our 2B leads GUI-Decisions-2B by \nosign{\vRzDOursMGuidHund} pp at $100{\times}100$ (95\% CI \vRzDOursMGuidHundLo{} to \vRzDOursMGuidHundHi) and \nosign{\vRzDOursMGuidPt} pp at the point (95\% CI \vRzDOursMGuidPtLo{} to \vRzDOursMGuidPtHi). It ties the off-the-shelf Qwen3-VL-2B at $100{\times}100$ (\vRzDOursMQvlTwoHund{} pp, 95\% CI \vRzDOursMQvlTwoHundLo{} to \vRzDOursMQvlTwoHundHi) and trails the larger MJev-Spatial-4B at the point (\vRzDOursMMjevPt{} pp, 95\% CI \vRzDOursMMjevPtLo{} to \vRzDOursMMjevPtHi). The pattern holds at 1\,MP: the released models find \vRhiGdThouAcc{} (GUI-Decisions-2B) to \vRhiQvlThouAcc\% (Qwen3-VL-4B and GroundingPI) of points and are wrong on \vRhiQvlThouWr{} to \vRhiGdThouWr\% of their most confident 70\%, against \vRhiOursMThouAcc{} and \vRhiOursLThouAcc\% of points and \vRhiOursMThouWr{} and \vRhiOursLThouWr\% wrong for our 2B and 4B. At \vfEdgeRes{} pixels the off-the-shelf Qwen3-VL-2B is wrong more often than our Qwen3.5-2B on its most confident 70\% (\vRhiQvsLoThouWr{} against \vRhiOursMLoThouWr\%), with point accuracy within one point (\vRhiQvsLoThouAcc{} against \vRhiOursMLoThouAcc\%). The option models choose well among a few large cells and fall far behind ours as the cells shrink (\cref{tab:coarsegrid}).

\paragraph{Released pointing models.} The strongest released VLMs that write a point, read off-the-shelf in their native formats (\cref{tab:families}), are about as accurate as our 4B model but rank their decisions less well and decide more slowly. Our 4B is level with Qwen3-VL-4B at the point (\vRgQvlFourDFour{} pp, 95\% CI \vRgQvlFourDFourLo{} to \vRgQvlFourDFourHi, \cref{tab:releasedauroc}). Rex-Omni and Molmo-7B-D are right on \vRwRexWrite{} and \vRwMolmoWrite\% (\cref{tab:readwrite}) and take \vRtRexHf{} and \vRtMolmoHf\,ms in the HF engine, since vLLM did not run them.

\subsection{Standard protocols}
\label{app:published}

\phantomsection\label{tab:stdpb}

Here each benchmark is scored with its own released rule, and our models read images at up to 1\,MP, as in training. On ScreenSpot-v2 our 2B reaches \vSmSS\%, above UI-TARS~\cite{qin2025uitars}, Qwen3-VL, GUI-Actor and GUI-Owl-1.5~\cite{xu2026mobileagent35} at 2B (\vPubSsLow--\vPubSsHigh\%) and below MAI-UI-2B~\cite{zhou2025maiui} and UI-Venus-1.5-2B~\cite{uivenus15}, which train on far more screen data (\cref{tab:stdss}).

On Point-Bench, as printed by their authors, Qwen3.5-2B with PointRL\footnote{PointRL, arXiv:2608.25299, Point-Bench table.} and Molmo2-4B\footnote{Molmo2 technical report, arXiv:2601.10611v4, Tab.~7; its earlier versions print higher scores that v4 attributes to an evaluation bug.} have a higher average than our 2B and 4B (\vPbPubRlAvg{} against \vSmPBAvg\%, \vPbPubMoAvg{} against \vSlPBAvg\%). The difference comes from the counting category, which a one-point read-out caps at \vSetCountCap: over the other four categories our 2B averages \vPbNcOursM\% against \vPbNcRl\% for PointRL, and our 4B \vPbNcOursL\% against \vPbNcMo\% for Molmo2-4B. Both of our models lead on spatial items (\vSmPBSpatial{} against \vPbPubRlSpatial\% at 2B, \vSlPBSpatial{} against \vPbPubMoSpatial\% at 4B).

Averaged over the eight RefCOCO/+/g splits, our 2B reaches \vNmFinBxAvg\% Acc@0.5, above every published general VLM of up to 3B that answers without reasoning (at most \vPubBestThree\%, LFM3-VL-2B), and our 4B \vNlFinBxAvg\%, above every published general VLM of up to 9B (at most \vPubBestNine\%, InternVL3.5-8B and Qwen3.5-9B). Models built for grounding report similar or higher values at small sizes: Jag-0.8B\newref{zhu2026jag}{arXiv:2610.13239} \vPubJagZeroEight\% over the five test splits and Florence-2-L (0.77B)\newref{xiao2024florence2}{Xiao et al., CVPR 2024} \vPubFlorenceTwoL\% on average over the eight. All of these models, ours included, train on the RefCOCO/+/g train splits.

On Where2Place our 4B (\vSlWP\%) is above RoboPoint-13B (\vPubWpRoboPoint\%) and below Embodied-R1-3B and RoboRefer-2B (\vPubWpEmbR{} and \vPubWpRoboRefer\%), both trained for embodied pointing; on RoboRefIt it is level with Embodied-R1-3B (\vSlRRbox{} against \vPubRrEmbR\%). On MiniWoB++ our 4B (\vSlMWft\%) is above Pix2Act (\vPubMwPixAct\%) and below SeeClick (\vPubMwSeeClick\%), both trained on MiniWoB++ episodes.

\paragraph{Faster written answers.} Other recent models make the written answer faster instead of reading it. LocateAnything\newref{locateanything}{arXiv:2605.27365} decodes a box or a point as one unit in a single step, Jag predicts the box with a head on the last input state of Qwen3.5-0.8B, LOCUS\newref{talha2026locus}{arXiv:2610.35053} drafts a box with such a head and checks it against the model's own coordinate tokens in about \vLocusPasses{} sequential passes, and PhysMoE\newref{liu2026physmoe}{arXiv:2611.13000} writes each coordinate as one token and supervises points with the signed distance field of a mask. Each returns one answer, while our read returns the probability of every point and a self-check in one pass. Reading typed answers instead of writing them is how the system one models introduced with Jev\newref{almeida2026jev}{Almeida, TypeSafe blog, 2026} decide.

\begin{table}[!htbp]
\centering\footnotesize
\setlength{\tabcolsep}{2.2pt}\renewcommand{\arraystretch}{0.88}
\begin{tabular}{@{}lccccc@{}}
\toprule
& \multicolumn{2}{c}{ScreenSpot-v2\,$\uparrow$} & Where2- & Robo- & MiniWoB \\
\cmidrule(lr){2-3}
Model & 1\,MP & \vfEdgeRes & Place\,$\uparrow$ & RefIt\,$\uparrow$ & ++\,$\uparrow$ \\
\midrule
\multicolumn{6}{@{}l}{\emph{Off-the-shelf}} \\
Qwen3.5-0.8B & \vSbsSS & \vSbsSSLo & -- & -- & -- \\
Qwen3.5-2B & \vSbmSS & \vSbmSSLo & \vSbmWP & \vSbmRRbox & \vSbmMWft \\
Qwen3.5-4B & \vSblSS & \vSblSSLo & \vSblWP & \vSblRRbox & \vSblMWft \\
\multicolumn{6}{@{}l}{\emph{Ours}} \\
\oursrow Qwen3.5-0.8B & \vSsSS & \vSsSSLo & \vSsWP & \vSsRRbox & \vSsMWft \\
\oursrow Qwen3.5-2B & \vSmSS & \vSmSSLo & \vSmWP & \vSmRRbox & \vSmMWft \\
\oursrow Qwen3.5-4B & \vSlSS & \vSlSSLo & \vSlWP & \vSlRRbox & \vSlMWft \\
\oursrow InternVL3.5-2B & \vSivlSS & \vSivlSSLo & \vSivlWP & \vSivlRRbox & \vSivlMWft \\
\multicolumn{6}{@{}l}{\emph{Published GUI models}} \\
UI-TARS-2B~\cite{qin2025uitars} & 84.7 & -- & -- & -- & -- \\
Qwen3-VL-2B (3rd)~\cite{zhou2025maiui} & 86.7 & -- & -- & -- & -- \\
GUI-Actor-2B~\cite{guiactor} & 88.6 & -- & -- & -- & -- \\
GUI-Owl-1.5-2B~\cite{xu2026mobileagent35} & 89.7 & -- & -- & -- & -- \\
MAI-UI-2B~\cite{zhou2025maiui} & 92.5 & -- & -- & -- & -- \\
UI-Venus-1.5-2B~\cite{uivenus15} & 92.8 & -- & -- & -- & -- \\
Qwen3-VL-4B (3rd)~\cite{yan2025stepgui} & 93.1 & -- & -- & -- & -- \\
GUI-Owl-1.5-4B~\cite{xu2026mobileagent35} & 93.2 & -- & -- & -- & -- \\
Step-GUI-4B~\cite{yan2025stepgui} & 93.6 & -- & -- & -- & -- \\
\multicolumn{6}{@{}l}{\emph{Published pointing and web-agent models}} \\
RoboPoint-13B~\cite{robopoint} & -- & -- & 46.8 & -- & -- \\
Embodied-R1-3B~\cite{yuan2025embodiedr1} & -- & -- & 69.5 & 85.6 & -- \\
RoboRefer-2B (RFT)~\cite{zhou2025roborefer} & -- & -- & 71.0 & -- & -- \\
Pix2Act~\cite{shaw2023pix2act}$^\dagger$ & -- & -- & -- & -- & 64.7 \\
SeeClick~\cite{screenspot}$^\dagger$ & -- & -- & -- & -- & 81.1 \\
\bottomrule
\end{tabular}
\caption{\textbf{ScreenSpot-v2, Where2Place, RoboRefIt and MiniWoB++ under their own protocols} (\%, higher is better; --: not reported, or not run for our rows). ScreenSpot-v2 (1,272 screenshots) counts a point in the official box, so its $384^2$ column differs slightly from \cref{tab:sets}, which clips the boxes. Where2Place (100 images) uses RoboPoint's soft-mask score. RoboRefIt (2,000) counts a point in the target box on the TestB subset that Embodied-R1 evaluates. MiniWoB++ is success at reward above 0.8 on SeeClick's 14 one-click tasks with 50 seeds each, while the main text uses our own 15 single-click tasks. (3rd) marks a base model evaluated by the cited third party, and $^\dagger$ a model trained on MiniWoB++ episodes, which our models never saw.}
\label{tab:stdss}\label{tab:stdrest}
\end{table}

\section{How the Released Option Models Were Timed and Prompted}
\label{app:opttime}

\paragraph{How the option models are timed.} \Cref{tab:dec} times the option models with the image processed once, as our rows are (\cref{tab:optonce}); this section first gives the times of their released runtimes, sped up only where this leaves the answers unchanged, which \cref{tab:releasedauroc} reports. These models have no coordinate output, so they are asked for a column and a row among ten options, and each finer grid repeats the two questions inside the chosen cell: one round for $10{\times}10$, two for $100{\times}100$ and three for the point. d3-nano reads its answer with the attention mode \texttt{noncausal\_full\_attention} of its released configuration, in which the attention layers attend over the whole prompt in both directions, so each question needs its own pass through the language model. The released runtimes also take one question per request, and resize and encode the image again for every request.

We run the text layers and the read-out as CUDA graphs and send the column and the row question of a round at the same time, which takes a $10{\times}10$ decision of d3-nano from \vFdDnCellOld{} to \vFdDnCell\,ms and of pplx-decider from \vFdPpCellOld{} to \vFdPpCell\,ms. Over three rounds, two axes and 205 images (the 200 timed images and the five warm-up images), the chosen option matches the released code in \vFdDnAgree{} choices for each model. Each model's own batched call for the two questions reproduces only \vFdDnBatchAgree{} and \vFdPpBatchAgree{} of these choices, so we do not use it.

\begin{table}[!t]
\centering\footnotesize
\setlength{\tabcolsep}{3pt}
\begin{tabular}{@{}l@{\hspace{6pt}}ccc@{\hspace{8pt}}ccc@{}}
\toprule
 & \multicolumn{3}{c}{released runtime} & \multicolumn{3}{c}{image processed once} \\
\cmidrule(lr){2-4}\cmidrule(l){5-7}
ms per decision\,$\downarrow$ & $10^2$ & $100^2$ & point & $10^2$ & $100^2$ & point \\
\midrule
\multicolumn{7}{@{}l}{\emph{0.8B}} \\
d3-lite & \vFdDlCell & \vFdDlHund & \vFdDlThou & \vVsDlCell & \vVsDlHund & \vVsDlThou \\
Intern-Decision-0.8B & \vFdIntSCell & \vFdIntSHund & \vFdIntSThou & \vVsIntSCell & \vVsIntSHund & \vVsIntSThou \\
OneJev1.5-0.8B & \vFdOjfCell & \vFdOjfHund & \vFdOjfThou & \vVsOjfCell & \vVsOjfHund & \vVsOjfThou \\
\oursrow ours, Qwen3.5-0.8B & -- & -- & -- & \vOssCell & \vOssHund & \vOssThou \\
\multicolumn{7}{@{}l}{\emph{2B}} \\
d3-nano & \vFdDnCell & \vFdDnHund & \vFdDnThou & \vVsDnCell & \vVsDnHund & \vVsDnThou \\
Intern-Decision-2B & \vFdIntMCell & \vFdIntMHund & \vFdIntMThou & \vVsIntMCell & \vVsIntMHund & \vVsIntMThou \\
\oursrow ours, Qwen3.5-2B & -- & -- & -- & \vWfmReadC & \vOsmHund & \vOsmThou \\
\multicolumn{7}{@{}l}{\emph{4B}} \\
pplx-decider & \vFdPpCell & \vFdPpHund & \vFdPpThou & \vVsPpCell & \vVsPpHund & \vVsPpThou \\
Intern-Decision-4B & \vFdIntLCell & \vFdIntLHund & \vFdIntLThou & \vVsIntLCell & \vVsIntLHund & \vVsIntLThou \\
MJev-Spatial-4B & -- & -- & \vFdMjevThouDec & -- & -- & \vVsMjevThouDec \\
\oursrow ours, Qwen3.5-4B & -- & -- & -- & \vWflReadC & \vOslHund & \vOslThou \\
\bottomrule
\end{tabular}

\caption{\textbf{Time per decision of the released models that need several rounds, with the released runtime and with the image processed once.} Milliseconds on one A100, median over 200 Point-Bench images. These models answer a $10{\times}10$ decision in one round, a $100{\times}100$ decision in two and the point in three. The released runtime, sped up as in \cref{tab:releasedauroc} where the choices stay the same, resizes and encodes the image for every question. Processed once, the image is resized on the GPU and encoded by a CUDA graph, as in our rows, and every choice stays the same. MJev-Spatial returns only a point and reads a new crop in every round, so each crop is still encoded; it takes an image path, and as for the other rows its times leave out decoding the file (\vFdMjevThou{} and \vVsMjevThou\,ms with it). Ours needs one pass, so only the right block applies to it (--).}
\label{tab:optonce}
\end{table}

\paragraph{With the image processed once.} Even when each image is resized and encoded only once, the rounds keep the points of the option models slower than ours (\cref{tab:optonce}). Of the \vVsDnReqMs\,ms of one d3-nano request, \vVsDnImgMs\,ms go to resizing and encoding the image. We therefore re-time the released models of \cref{tab:releasedauroc} that need several rounds, with the same CUDA graphs as there, and now resize and encode each image once. A further round then runs only the text layers. For each option model all \vVsAgree{} choices again match the released code. MJev-Spatial's \vVsMjevAgree{} choices also stay the same.

Processing the image once takes the point of d3-nano from \vFdDnThou{} to \vVsDnThou\,ms and that of pplx-decider from \vFdPpThou{} to \vVsPpThou\,ms (\cref{tab:optonce}). Each further round still adds \vVsDnRound\,ms for d3-nano and \vVsPpRound\,ms for pplx-decider, so the point of d3-nano takes \vVsDnOverOurs{} times as long as ours on the same backbone and the point of pplx-decider \vVsPpOverOurs{} times as long. One $10{\times}10$ decision of d3-nano is still as fast as the $10{\times}10$ read of ours on the same backbone (\vVsDnCell{} against \vWfmReadC\,ms). The same holds at 0.8B: one $10{\times}10$ round of d3-lite and of OneJev1.5-0.8B takes \vVsDlCell{} and \vVsOjfCell\,ms against \vOssCell\,ms for ours, and their points take \vVsDlThou{} and \vVsOjfThou\,ms against \vOssThou\,ms.

\paragraph{One round of 100 options.} Asked in one round of 100 options or in rounds of ten, d3-nano and pplx-decider find fewer than half of the $10{\times}10$ cells, against \vRzNatMCellAcc\% for the off-the-shelf Qwen3.5-2B read our way (\cref{tab:dec}). The runtimes of both models accept up to \vOhMaxOpt{} options, so the $100{\times}100$ decision can also be asked in one round, with the columns and the rows numbered 1 to 100 and the rest of the question unchanged. \Cref{tab:opthundred} compares the two. pplx-decider is more accurate in one round: it finds \vOhPpOneCell\% of $10{\times}10$ cells against \vOhPpTenCell\% (\vOhPpDCell{} pp, 95\% CI \vOhPpDCellLo{} to \vOhPpDCellHi) and \vOhPpOnePt\% of points against \vOhPpTenPt\%, but its $100{\times}100$ decision is slower, \vOhPpOneMs{} against \vOhPpTenMs\,ms. d3-nano is less accurate in one round, with \vOhDnOneCell\% of $10{\times}10$ cells against \vOhDnTenCell\% ($\vOhDnDCell$ pp, 95\% CI $\vOhDnDCellLo$ to $\vOhDnDCellHi$), and faster, \vOhDnOneMs{} against \vOhDnTenMs\,ms.

\begin{table}[!t]
\centering\footnotesize
\setlength{\tabcolsep}{3pt}
\begin{tabular}{@{}l@{\hspace{6pt}}c@{\hspace{6pt}}c@{\hspace{6pt}}c@{\hspace{8pt}}c@{}}
\toprule
 & \multicolumn{3}{c}{accuracy (\%) $\uparrow$} & ms $\downarrow$ \\
\cmidrule(lr){2-4}
Model, question & $10^2$ & $100^2$ & point & $100^2$ \\
\midrule
d3-nano, ten options per round & \vOhDnTenCell & \vOhDnTenHund & \vOhDnTenPt & \vOhDnTenMs \\
d3-nano, 100 options, one round & \vOhDnOneCell & \vOhDnOneHund & \vOhDnOnePt & \vOhDnOneMs \\
\addlinespace[3pt]
pplx-decider, ten options per round & \vOhPpTenCell & \vOhPpTenHund & \vOhPpTenPt & \vOhPpTenMs \\
pplx-decider, 100 options, one round & \vOhPpOneCell & \vOhPpOneHund & \vOhPpOnePt & \vOhPpOneMs \\
\bottomrule
\end{tabular}

\caption{\textbf{Option models asked in rounds of ten options or in one round of 100 options.} Accuracy on the \vBcNAll{} test images and the time of the $100{\times}100$ decision on one A100 with the released runtime, measured in one session, so the times with ten options differ slightly from those of \cref{tab:dec}. With 100 options, the $10{\times}10$ decision sums the probabilities of the ten options in each coarse cell, and the point is the centre of the chosen $100{\times}100$ cell. With ten options, the point takes a third round.}
\label{tab:opthundred}
\end{table}
\begin{table*}[!t]
\centering\footnotesize
\setlength{\tabcolsep}{4pt}
\begin{tabular}{@{}l*{5}{c>{\columncolor{oursbg}}c}@{}}
\toprule
& \multicolumn{10}{c}{wrong@70\% $\downarrow$ within each benchmark (number of images)} \\
& \multicolumn{2}{c}{Point-Bench (\vPbNPB)} & \multicolumn{2}{c}{ScreenSpot-v2 (\vPbNSS)} & \multicolumn{2}{c}{Where2Place (\vPbNWP)} & \multicolumn{2}{c}{RoboRefIt (\vPbNRR)} & \multicolumn{2}{c}{RefCOCO (\vPbNRC)} \\
\cmidrule(lr){2-3}\cmidrule(lr){4-5}\cmidrule(lr){6-7}\cmidrule(lr){8-9}\cmidrule(lr){10-11}
Base VLM & writes & ours & writes & ours & writes & ours & writes & ours & writes & ours \\
\midrule
Qwen3.5-0.8B & \vPbQwSPBWrWrong & \vPbQwSPBOursWrong & \vPbQwSSSWrWrong & \textbf{\vPbQwSSSOursWrong} & \vPbQwSWPWrWrong & \textbf{\vPbQwSWPOursWrong} & \vPbQwSRRWrWrong & \textbf{\vPbQwSRROursWrong} & \vPbQwSRCWrWrong & \textbf{\vPbQwSRCOursWrong} \\
Qwen3.5-2B & \textbf{\vPbQwMPBWrWrong} & \vPbQwMPBOursWrong & \vPbQwMSSWrWrong & \textbf{\vPbQwMSSOursWrong} & \vPbQwMWPWrWrong & \textbf{\vPbQwMWPOursWrong} & \vPbQwMRRWrWrong & \textbf{\vPbQwMRROursWrong} & \vPbQwMRCWrWrong & \textbf{\vPbQwMRCOursWrong} \\
Qwen3-VL-2B & \vPbQvlTwoPBWrWrong & \textbf{\vPbQvlTwoPBOursWrong} & \vPbQvlTwoSSWrWrong & \textbf{\vPbQvlTwoSSOursWrong} & \vPbQvlTwoWPWrWrong & \textbf{\vPbQvlTwoWPOursWrong} & \vPbQvlTwoRRWrWrong & \textbf{\vPbQvlTwoRROursWrong} & \vPbQvlTwoRCWrWrong & \textbf{\vPbQvlTwoRCOursWrong} \\
InternVL3.5-1B & \textbf{\vPbIvlAPBWrWrong} & \vPbIvlAPBOursWrong & \vPbIvlASSWrWrong & \textbf{\vPbIvlASSOursWrong} & \vPbIvlAWPWrWrong & \textbf{\vPbIvlAWPOursWrong} & \vPbIvlARRWrWrong & \textbf{\vPbIvlARROursWrong} & \vPbIvlARCWrWrong & \textbf{\vPbIvlARCOursWrong} \\
InternVL3.5-2B & \vPbIvlBPBWrWrong & \textbf{\vPbIvlBPBOursWrong} & \vPbIvlBSSWrWrong & \textbf{\vPbIvlBSSOursWrong} & \vPbIvlBWPWrWrong & \textbf{\vPbIvlBWPOursWrong} & \vPbIvlBRRWrWrong & \textbf{\vPbIvlBRROursWrong} & \vPbIvlBRCWrWrong & \textbf{\vPbIvlBRCOursWrong} \\
LFM3-VL-2B & \vPbLfvPBWrWrong & \textbf{\vPbLfvPBOursWrong} & \vPbLfvSSWrWrong & \textbf{\vPbLfvSSOursWrong} & \vPbLfvWPWrWrong & \textbf{\vPbLfvWPOursWrong} & \vPbLfvRRWrWrong & \textbf{\vPbLfvRROursWrong} & \vPbLfvRCWrWrong & \textbf{\vPbLfvRCOursWrong} \\
\midrule
Qwen3.5-4B & \vPbQwLPBWrWrong & \textbf{\vPbQwLPBOursWrong} & \vPbQwLSSWrWrong & \textbf{\vPbQwLSSOursWrong} & \vPbQwLWPWrWrong & \textbf{\vPbQwLWPOursWrong} & \vPbQwLRRWrWrong & \textbf{\vPbQwLRROursWrong} & \vPbQwLRCWrWrong & \textbf{\vPbQwLRCOursWrong} \\
Qwen3-VL-4B & \vPbQvlFourPBWrWrong & \textbf{\vPbQvlFourPBOursWrong} & \vPbQvlFourSSWrWrong & \textbf{\vPbQvlFourSSOursWrong} & \vPbQvlFourWPWrWrong & \textbf{\vPbQvlFourWPOursWrong} & \vPbQvlFourRRWrWrong & \textbf{\vPbQvlFourRROursWrong} & \vPbQvlFourRCWrWrong & \textbf{\vPbQvlFourRCOursWrong} \\
InternVL3.5-4B & \vPbIvlDPBWrWrong & \textbf{\vPbIvlDPBOursWrong} & \vPbIvlDSSWrWrong & \textbf{\vPbIvlDSSOursWrong} & \vPbIvlDWPWrWrong & \textbf{\vPbIvlDWPOursWrong} & \vPbIvlDRRWrWrong & \textbf{\vPbIvlDRROursWrong} & \vPbIvlDRCWrWrong & \textbf{\vPbIvlDRCOursWrong} \\
Gemma-4-E4B & \vPbGemPBWrWrong & \textbf{\vPbGemPBOursWrong} & \vPbGemSSWrWrong & \textbf{\vPbGemSSOursWrong} & \vPbGemWPWrWrong & \textbf{\vPbGemWPOursWrong} & \vPbGemRRWrWrong & \textbf{\vPbGemRROursWrong} & \vPbGemRCWrWrong & \textbf{\vPbGemRCOursWrong} \\
LFM2.5-VL-3B & \vPbLfmPBWrWrong & \textbf{\vPbLfmPBOursWrong} & \vPbLfmSSWrWrong & \textbf{\vPbLfmSSOursWrong} & \vPbLfmWPWrWrong & \textbf{\vPbLfmWPOursWrong} & \vPbLfmRRWrWrong & \textbf{\vPbLfmRROursWrong} & \vPbLfmRCWrWrong & \textbf{\vPbLfmRCOursWrong} \\
Rex-Omni-3B & \vPbRexPBWrWrong & \textbf{\vPbRexPBOursWrong} & \vPbRexSSWrWrong & \textbf{\vPbRexSSOursWrong} & \vPbRexWPWrWrong & \textbf{\vPbRexWPOursWrong} & \vPbRexRRWrWrong & \textbf{\vPbRexRROursWrong} & \vPbRexRCWrWrong & \textbf{\vPbRexRCOursWrong} \\
\bottomrule
\end{tabular}

\caption{\textbf{Wrong@70\% within each benchmark, for the point-writing model (writes) and the decision model (ours) of every base VLM.} The images of a benchmark are ranked by the model's confidence, as in \cref{tab:decb}, and a cell gives the wrong points among the 70\% the model is most confident about (\%, lower is better). The number of images is in brackets, and the lower value of each pair is in bold; when the two are equal, neither is bold. One run per model.}
\label{tab:perbench}
\end{table*}
\section{Confident Errors per Benchmark}
\label{app:perbench}

\Cref{tab:perbench} gives the wrong@70\% of \cref{tab:decb} within each benchmark (intervals from 2,000 resamples). On ScreenSpot-v2, RoboRefIt and RefCOCO, ours makes fewer confident errors than the point-writing model for all twelve models, and every 95\% CI of the difference is below zero. On Where2Place ours is also lower for all twelve, but with only 100 images the interval is below zero for \vPbSumWPSig{} of them. On Point-Bench the two models are close: ours is lower for \vPbSumPBLower{} of the twelve models, equal for \vPbSumPBTie{} (Qwen3.5-0.8B) and higher for \vPbSumPBHigher{} (Qwen3.5-2B and InternVL3.5-1B), and the interval is below zero for \vPbSumPBSig{} of them (Gemma-4-E4B and Rex-Omni-3B).

\paragraph{Which cell ranks the point-writing model.} Ours makes fewer confident errors than the point-writing model regardless of which cell is used to rank it (\cref{tab:confchoice}). \Cref{tab:dec,tab:decb,tab:perbench} rank the point-writing Qwen3-VL-2B, Qwen3-VL-4B and Rex-Omni-3B by the mass of their top $100{\times}100$ cell and the other nine models by the mass of their top $10{\times}10$ cell. \Cref{tab:confchoice} compares the two cells for six models: the $10{\times}10$ cell is the better confidence for all six and lowers the wrong@70\% of the point-writing model by \vCcGapMin{} to \vCcGapMax{} pp. Ranked by it, the point-writing Qwen3-VL-2B, Qwen3-VL-4B and Rex-Omni-3B are still wrong more often on their most confident 70\% than ours and have a lower AUROC; at 2B, for example, Qwen3-VL is wrong on \vCcQvlTwoTenWrong\% against \vCcQvlTwoOursWrong\% for ours.

\begin{table}[!t]
\centering\footnotesize
\setlength{\tabcolsep}{3pt}
\begin{tabular}{@{}l@{\hspace{6pt}}cc>{\columncolor{oursbg}}c@{\hspace{8pt}}cc>{\columncolor{oursbg}}c@{}}
\toprule
 & \multicolumn{3}{c}{wrong@70\% $\downarrow$} & \multicolumn{3}{c}{AUROC $\uparrow$} \\
\cmidrule(lr){2-4}\cmidrule(l){5-7}
 & \multicolumn{2}{c}{writes, by its top} & & \multicolumn{2}{c}{writes, by its top} & \\
Base VLM & $10^2$ cell & $100^2$ cell & ours & $10^2$ cell & $100^2$ cell & ours \\
\midrule
Qwen3.5-0.8B & \vCcQwSTenWrong$^\ast$ & \vCcQwSHundWrong & \vCcQwSOursWrong & \vCcQwSTenAu$^\ast$ & \vCcQwSHundAu & \vCcQwSOursAu \\
Qwen3.5-2B & \vCcQwMTenWrong$^\ast$ & \vCcQwMHundWrong & \vCcQwMOursWrong & \vCcQwMTenAu$^\ast$ & \vCcQwMHundAu & \vCcQwMOursAu \\
Qwen3.5-4B & \vCcQwLTenWrong$^\ast$ & \vCcQwLHundWrong & \vCcQwLOursWrong & \vCcQwLTenAu$^\ast$ & \vCcQwLHundAu & \vCcQwLOursAu \\
\addlinespace[3pt]
Qwen3-VL-2B & \vCcQvlTwoTenWrong & \vCcQvlTwoHundWrong$^\ast$ & \vCcQvlTwoOursWrong & \vCcQvlTwoTenAu & \vCcQvlTwoHundAu$^\ast$ & \vCcQvlTwoOursAu \\
Qwen3-VL-4B & \vCcQvlFourTenWrong & \vCcQvlFourHundWrong$^\ast$ & \vCcQvlFourOursWrong & \vCcQvlFourTenAu & \vCcQvlFourHundAu$^\ast$ & \vCcQvlFourOursAu \\
Rex-Omni-3B & \vCcRexTenWrong & \vCcRexHundWrong$^\ast$ & \vCcRexOursWrong & \vCcRexTenAu & \vCcRexHundAu$^\ast$ & \vCcRexOursAu \\
\bottomrule
\end{tabular}

\caption{\textbf{The point-writing model ranked by its top $10{\times}10$ cell or its top $100{\times}100$ cell.} Wrong@70\% and AUROC at the point on the \vBcNAll{} test images, with ours ranked by its self-check as in \cref{tab:decb}. $^\ast$The cell used in \cref{tab:dec,tab:decb}. One run per model; for Qwen3.5, \cref{tab:decb} gives the mean over seeds.}
\label{tab:confchoice}
\end{table}
\begin{table*}[!t]
\centering\footnotesize
\setlength{\tabcolsep}{4pt}
\begin{tabular}{@{}l@{\hspace{8pt}}ccccc@{\hspace{12pt}}ccccc@{}}
\toprule
 & \multicolumn{5}{c}{image as stored} & \multicolumn{5}{c}{\vfEdgeRes{} pixels} \\
\cmidrule(lr){2-6}\cmidrule(l){7-11}
Model & $10^2$ $\uparrow$ & $100^2$ $\uparrow$ & point $\uparrow$ & AUROC $\uparrow$ & wrong@70\% $\downarrow$ & $10^2$ $\uparrow$ & $100^2$ $\uparrow$ & point $\uparrow$ & AUROC $\uparrow$ & wrong@70\% $\downarrow$ \\
\midrule
pplx-decider-v1.2 & \vApPplxNatCell & \vApPplxNatCellH & \vApPplxNatPt & \vApPplxNatAuc & \vApPplxNatWr & \vApPplxDepCell & \vApPplxDepCellH & \vApPplxDepPt & \vApPplxDepAuc & \vApPplxDepWr \\
gpt-6-luna-decisions & \vApLunaNatCell & \vApLunaNatCellH & \vApLunaNatPt & \vApLunaNatAuc & \vApLunaNatWr & \vApLunaDepCell & \vApLunaDepCellH & \vApLunaDepPt & \vApLunaDepAuc & \vApLunaDepWr \\
Clef & \vApClefNatCell & \vApClefNatCellH & \vApClefNatPt & \vApClefNatAuc & \vApClefNatWr & \vApClefDepCell & \vApClefDepCellH & \vApClefDepPt & \vApClefDepAuc & \vApClefDepWr \\
\addlinespace[3pt]
\oursrow ours, Qwen3.5-2B & -- & -- & -- & -- & -- & \vRzOursMCellAcc & \vRzOursMHundAcc & \vRzOursMPtAcc & \vApQwMAuc & \vApQwMWr \\
\oursrow ours, Qwen3.5-4B & -- & -- & -- & -- & -- & \vRzOursLCellAcc & \vRzOursLHundAcc & \vRzOursLPtAcc & \vApQwLAuc & \vApQwLWr \\
\bottomrule
\end{tabular}

\caption{\textbf{Decision models served through an API, asked in rounds like the option models of \cref{tab:dec}.} Point accuracy and cell accuracy (\%) on the \vBcNAll{} test images, with the image as stored (long side up to 1024 pixels) and at the \vfEdgeRes{} pixels of our rows; our rows are those of \cref{tab:dec,tab:releasedauroc}, read at \vfEdgeRes{} only (--). The confidence of an API model is the probability it returns for its first-round cell, and that of ours the self-check $P(\text{yes})$. AUROC measures how well it separates right from wrong points (chance is 0.5), and wrong@70\% is the wrong share among the 70\% of images each model is most confident about.}
\label{tab:apidec}
\end{table*}
\section{The Drone Agent at 2B and the Video}
\label{app:drone2b}

At 2B the read takes \vDbDPiS\,s per decision on the Raspberry Pi~5 and the written action \vDbWPiS\,s, both far beyond about \vSetReachLim\,s per decision, up to which episodes are still reached (\cref{fig:dronelat}a). With each decision held to its Pi time and the Pi's smaller frame (long side \vSetPiFrame\,pixels), the 2B decision model reaches the target in \vDbDPisAll{} of \vDrEpis{} episodes and the 2B writer in \vDbWPisAll, on average over the runs of \cref{tab:droneruns}, and neither reaches a moving target in any episode. The writer leads when searching (\vDbWPisSearch{} against \vDbDPisSearch{} of 20 episodes) and ours with the target in view (\vDrScAppCDPiS{} against \vNtCJAppPiS), and the ranges of the totals over the runs overlap. With the Pi time on the simulator's frames, in one run, ours reaches the target in \vDrReachCDPi{} episodes and the writer in \vNsCJPiReach.

On the L40S GPU, where both models decide in well under a second (\vLrCDSecGpu\,s for ours and \vLrCJSecGpu\,s for the writer), the two reach the target in about as many episodes (\vLrCDGpuMean{} and \vLrCJGpuMean{} of \vDrEpis, \cref{tab:dronebody}). They differ in search + track, \vLrCDSetGpu{} against \vLrCJSetGpu{} of 20. The logs of the first run (\cref{tab:droneruns}) trace this to behaviour, since the delay from frame to action is the same in successful and failed episodes. The benchmark counts a declaration made within its radius with the target in view as reaching the target, which favours the writer: it keeps the target in view for \vDbWStView\% of an episode against \vDbDStView\% for ours, comes closer to it (a median of \vDbWStDist{} against \vDbDStDist\,m while it is in view) and declares arrival more often (\vDbWStDecl{} against \vDbDStDecl{} times per episode).

\paragraph{Video: the same writer at L40S and Pi time.}\label{app:video} The video on the project page and \cref{fig:video} show the effect of \cref{fig:dronelat}a on a tracking episode other than that of \cref{fig:dronelat}c, run on the simulator's frames so that only the time per decision changes. Three panels run on one episode clock: the 0.8B JSON writer at its L40S time, the same writer with each decision delayed to its Raspberry Pi~5 time, and our 0.8B decision model at its Pi time. Each panel holds a camera frame until the next decision and shows the action taken and a top view of drone and target.

The writer's first decision is the same in both settings, a flight toward the car; what changes is how often it can correct its course. At its L40S time it answers \vVidGpuDecForty{} times before it reaches the car after \vVidGpuReach\,s; each step takes longer than the decision itself, since it also includes capturing the frame and sending the command. At its Pi time it has answered \vVidPiDecForty{} times by 40\,s, against \vVidOursDecForty{} for ours. Between 10 and 30\,s the writer closes on the car at \vVidGpuClose\,m/s at its L40S time and at \vVidPiClose\,m/s at its Pi time, and ours at \vVidOursClose\,m/s. At the Pi's pace the writer keeps flying toward where the car was, the car leaves its view after \vVidPiLostT\,s, \vVidPiLostD\,m away, and it does not find the car again within the episode. Ours reaches the car after \vVidOursReach\,s.

This is why the time on the device matters for an agent: in a closed loop the world does not wait for the answer. On the Pi the written JSON action takes \vPiMeanQJ\,s, well beyond the \vSetReachLim\,s up to which episodes are reached (\cref{fig:dronelat}a), and our decision \vPiMeanQD\,s, close to it. In one run of each 0.8B model with its Pi time on the simulator's frames, our decision model alone reaches the target in \vDbVidOursOnly{} episodes, the writer alone in \vDbVidWrOnly, and both in \vDbVidBoth.

\makeatletter\setlength{\@fptop}{0pt}\setlength{\@dblfptop}{0pt}\makeatother   %
\section{Decision Models Served Through an API}
\label{app:apidec}

\paragraph{Setting.} We query three decision models served through the OpenRouter Decisions API, which returns a choice among given options with a probability for each. They are gpt-6-luna-decisions, Clef and pplx-decider-v1.2 (same family as the pplx-decider of \cref{tab:dec}). We ask them in two rounds of one 100-cell question each, with the same instruction on all \vBcNAll{} test images. The first round asks ``Which grid cell contains the point that best satisfies the instruction?'' with the 100 cells of a $10{\times}10$ grid as options, and the second asks it again, on the full image, among the 100 cells inside the chosen cell. The point is the centre of the chosen $100{\times}100$ cell, and the confidence is the returned probability of the chosen first-round cell. Ranked by the product of both rounds, as the option models are, they reach an AUROC of at most \vApLunaDepAucPP{} (ours \vApQwMAuc) and their wrong@70\% falls by under two points.

\paragraph{Images and refusals.} A refused call is retried once and then counted as wrong; gpt-6-luna-decisions declines \vApLunaNatRefuse{} of its $2{\times}$\vBcNAll{} calls with the stored image and \vApLunaDepRefuse{} at \vfEdgeRes{} pixels, and the other two decline none.

\paragraph{Time.} These times include network and service delays, so we compare accuracy and confidence only. Measured from our cluster with 24 requests in flight, a round takes a median of \vApLunaNatMs{} ms (gpt-6-luna-decisions) to \vApClefNatMs{} ms (Clef), and both rounds \vApLunaNatTwoMs{} to \vApClefNatTwoMs{} ms.

\paragraph{Results.} The API models (\cref{tab:apidec}) find at most \vApClefNatCell\% of the $10{\times}10$ cells, above d3-nano and pplx-decider in \cref{tab:dec}, and at most \vApClefNatPt\% of the points, against \vRzOursMPtAcc\% and \vRzOursLPtAcc\% for our 2B and 4B models at \vfEdgeRes{} pixels. Even against the best of them, Clef with the stored image, our 2B model finds \nosign{\vApQwMClefNatDPt} pp more points (95\% CI \vApQwMClefNatDPtLo{} to \vApQwMClefNatDPtHi; 2,000 resamples) and is wrong on \vApQwMWr\% of its most confident 70\% against \vApClefNatWr\% (95\% CI of the difference \vApQwMClefNatDWrLo{} to \vApQwMClefNatDWrHi{} pp). Their returned probability barely separates right from wrong: its AUROC is \vApLunaNatAuc{} for gpt-6-luna-decisions, \vApClefNatAuc{} (chance) for Clef and \vApPplxNatAuc{} for pplx-decider-v1.2, whose most confident answers are wrong most often, against \vApQwMAuc{} and \vApQwLAuc{} for our self-check.
\end{document}